\documentclass[10pt,twocolumn,letterpaper]{article}

\usepackage[pagenumbers]{cvpr} % To force page numbers, e.g. for an arXiv version
\usepackage{times}
\usepackage{rotating} % For vertical text
\usepackage{multirow}
\usepackage{epsfig}
\usepackage{graphicx}
\usepackage{amsmath}
\usepackage{amssymb}
\usepackage{gensymb}
\usepackage{graphicx}
\usepackage{amsmath}
\usepackage{cuted}
\usepackage{float}

\usepackage{amssymb}
\usepackage{booktabs}
\usepackage{nicefrac}
\usepackage{multirow}
\usepackage{algpseudocode}
\usepackage{algorithm}
\usepackage{comment}
\usepackage{amsthm}
\usepackage{bm}
\usepackage[dvipsnames]{xcolor}

\definecolor{cvprblue}{rgb}{0.21,0.49,0.74}
\usepackage[pagebackref,breaklinks,colorlinks,citecolor=cvprblue]{hyperref}
\usepackage{xspace}

\newcommand{\ourmethod}{\textsc{Ag}\xspace}
\newcommand{\ourlinearmethod}{\textsc{LinearAg}\xspace}
\newcommand{\cfg}{\textsc{Cfg}\xspace}

\def\paperID{***} % *** Enter the Paper ID here
\def\confName{CVPR}
\def\confYear{2024}

\newcommand\blfootnote[1]{% 
\begingroup 
\renewcommand\thefootnote{}\footnote{#1}% 
\addtocounter{footnote}{-1}% 
\endgroup 
}

\title{Adaptive Guidance: Training-free Acceleration of Conditional Diffusion Models}

\author{
Angela Castillo$^{*1}$ \quad
Jonas Kohler$^{*2}$ \quad
Juan C. Pérez$^{*2,3}$ \quad
Juan Pablo Pérez$^{1}$ \\
Albert Pumarola$^{2}$ \quad
Bernard Ghanem$^{3}$ \quad
Pablo Arbeláez$^{1}$ \quad
Ali Thabet$^{2}$ \\
$^{1}$Center for Research and Formation in Artificial Intelligence, Universidad de los Andes \\ \quad $^{2}$GenAI, Meta\\
\quad $^{3}$King Abdullah University of Science and Technology (KAUST) \\
}

\begin{document}
\maketitle
\blfootnote{* Equal contributions.}
\begin{strip}
\vspace{-2.5em}
% \vspace{-1.5cm}
\centering
\resizebox{\textwidth}{!}{
% \captionsetup{type=figure}
% \includegraphics[trim={1 1 1 1},clip,width=\linewidth]{figures/Fig1.png}
\includegraphics[width=\linewidth]{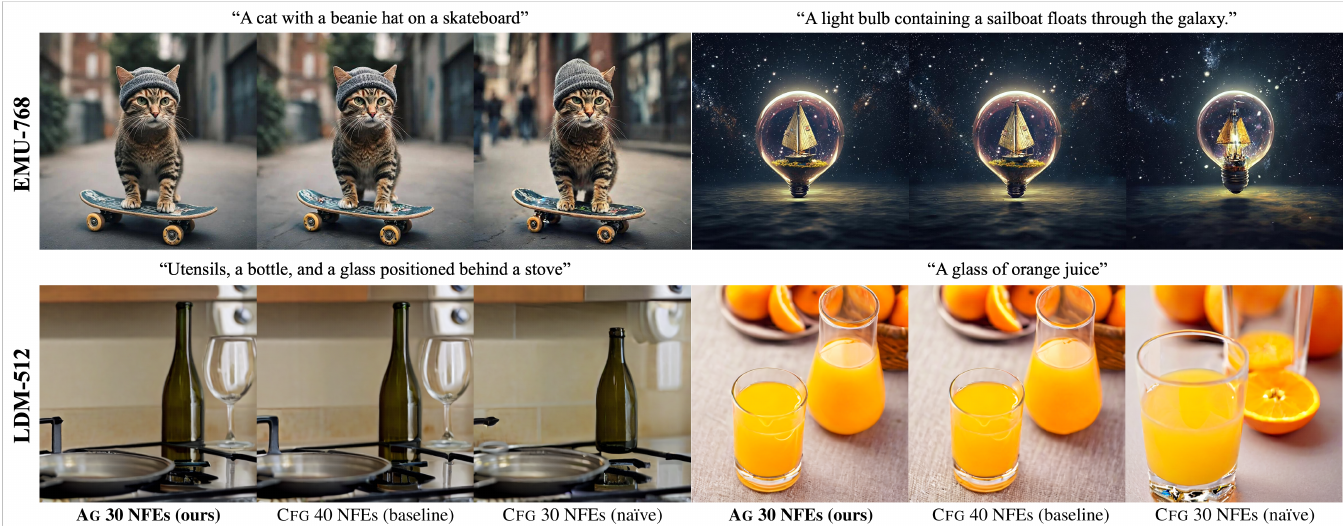}
% \vspace{-.7cm}
}
\captionof{figure}{
\textbf{Accelerating Guided Diffusion Models with Adaptive Guidance: }\footnotesize  By casting diffusion guidance as a Neural Architecture Search problem, we present Adaptive Guidance (\ourmethod), an efficient variant of Classifier-Free Guidance that saves $25\%$ of total NFEs without compromising image quality. \ourmethod constitutes a training-free, plug-and-play alternative to Guidance Distillation that achieves $50\%$ of its speed-ups while offering the ability to handle dynamic negative prompts. As depicted above, our approach (left) replicates the baseline one-to-one and furthermore outperforms a na\"{i}ve reduction of diffusion steps (right).} % \juan{dont forget to change ``TG'' in the bottom of the images}}
\vspace{.3cm}
\label{fig:teaser}
\end{strip}
% \end{figure*}    
% \begin{abstract}
% This paper presents a comprehensive study on the role of classifier-free guidance in text-conditioned diffusion models, particularly focusing on enhancing inference efficiency. Our investigation reveals that the conventional method of applying a consistent guidance strength throughout all diffusion steps is not the most effective strategy since not all diffusion steps contribute equally to the semantic structure of the generated images. Inspired by the literature on differentiable neural architecture search, we formulate the discovery of guidance policies that yield an ideal balance between latency and performance as a gradient-based search problem. Out of the found policies, we distill two very simple variations of classifier-free guidance that can be seamlessly integrated into existing diffusion models. The first approach, termed ``truncated guidance," achieves $30-50\%$ of the speed improvements attributable to guidance distillation, yet it requires neither re-training nor re-evaluation AND VERY EASY TO IMPLEMENT. MENTION OLS.  Our findings offer interesting insights into the evolution of the diffusion process and provide widely applicable improvements in inference efficiency, contributing to more practical and swift deployment of text-conditioned diffusion models.

% TRADE OFF !!!
% \end{abstract}

\begin{abstract}
\vspace{-.8cm}
This paper presents a comprehensive study on the role of Classifier-Free Guidance (\cfg) in text-conditioned diffusion models from the perspective of inference efficiency. 
% This paper presents a comprehensive study on improving the efficiency of diffusion models that leverage Classifier-Free Guidance (\cfg) for conditional generation. % in text-conditioned diffusion models, emphasizing inference efficiency and flexibility. 
%This paper studies the efficiency of diffusion models that use Classifier-Free Guidance (\cfg) for conditional generation.
% This paper presents Truncated Guidance (\ourmethod), a simple and flxible imp
In particular, we relax the default choice of applying \cfg in all diffusion steps and instead search for efficient guidance policies. We formulate the discovery of such policies in the differentiable Neural Architecture Search framework.
% Inspired by Neural Architecture Search (NAS), we formulate the discovery of guidance policies as gradient-based search, and probe for a balance between computational cost and image quality.
%Our findings suggest that not all denoising steps contribute equally to image semantics and quality. 
Our findings suggest that the denoising steps proposed by \cfg become increasingly aligned with simple conditional steps, which renders the extra neural network evaluation of \cfg redundant, especially in the second half of the denoising process. 
Building upon this insight, we propose ``Adaptive Guidance'' (\ourmethod), an efficient variant of \cfg, that adaptively omits network evaluations when the denoising process displays convergence.
Our experiments demonstrate that \ourmethod preserves \cfg's image quality while reducing computation by $25\%$. Thus, \ourmethod constitutes a plug-and-play alternative to Guidance Distillation, achieving $50\%$ of the speed-ups of the latter while being training-free and retaining the capacity to handle negative prompts. %Furthermore, the samples of \ourmethod replicate the baseline model one-to-one, thereby making the need for re-evaluations obsolete.
% Targeting even more computationally-constrained scenarios, we further propose \ourlinearmethod, a technique that exploits linear predictability in the denoising process to further .
Finally, we uncover further redundancies of \cfg in the first half of the diffusion process, showing that entire neural function evaluations can be replaced by simple affine transformations of past score estimates. This method, termed
\ourlinearmethod, offers even cheaper inference at the cost of deviating from the baseline model.
% In summary, 
Our findings provide insights into the efficiency of the conditional denoising process that contribute to more practical and swift deployment of text-conditioned diffusion models.
\vspace{-.2cm}
\end{abstract}
    
\vspace{-0.25cm}
\section{Introduction}
\label{sec:intro}
\vspace{-.5cm}
Diffusion Models (DMs)~\cite{ho2020denoising} exhibit outstanding generative capacities across domains such as images \cite{rombach2022high}, video~\cite{ho2022imagen}, audio \cite{kong2020diffwave}, human pose estimation~\cite{castillo2023bodiffusion}, and even cosmological simulations~\cite{schanz2023stochastic}. 
DMs generate data by sampling a noise instance and iteratively denoising the instance with a neural network.
The sequential nature of this denoising operation makes sampling from DMs a slow and expensive process.
In particular, the time required to sample from a given DM is a function of \textit{(i)}~the latency of each denoising iteration, and \textit{(ii)}~the total number of denoising steps.

Many practical applications entail ``conditional generation'', where DMs create samples conditioned on specific criteria such as a class, a text, or an image~\cite{nichol2021glide}.
DMs achieve conditional generation by replacing regular (\ie, unconditional) denoising steps with conditional ones, in which the neural network processes both the input and the condition.
While conditional denoising steps provide competitive results, Ho \etal proposed the technique of Classifier-Free Guidance (\cfg) \cite{ho2022classifier} to enhance sample quality.
\cfg enriches the conditional denoising process by leveraging implicit priors of the diffusion model itself.
Despite its simplicity, \cfg significantly improves sample quality in tasks such as text-to-image \cite{nichol2021glide,peebles2023scalable,dai2023emu}, image editing \cite{brooks2023instructpix2pix,meng2021sdedit,sheynin2023emu}, and text-to-3D \cite{poole2022dreamfusion,lin2023magic3d}.
Yet, the benefits of \cfg come at the cost of \textit{duplicating} the Number of Function Evaluations (NFEs), since each denoising iteration requires evaluating the neural network both conditionally and unconditionally.
Adding to the problem, neural networks used in practice for DMs max out the parallelization capacity of production-grade GPUs\footnote{By saturating memory bandwidth and/or CUDA cores. For example, using bfloat$16$ and batch size of 1, an EMU-768 model requires $1'553$ ms on an A100 GPU without \cfg. With \cfg, latency almost doubles to $2'865$ ms.}, preventing simultaneous computation of the conditional and unconditional function evaluations. 

% This paper examines \cfg and introduces methods to lower its computational demands without sacrificing image quality in text-to-image applications.
In this paper, we improve the efficiency of text-to-image diffusion models that use Classifier-Free Guidance (\cfg).
% Our analysis reveals that not all denoising steps are crucial, leading us to a more flexible approach than traditional \cfg.
Our analysis reveals that not all denoising steps contribute equally to image quality, suggesting that the traditional policy of applying \cfg in all steps is sub-optimal.
Instead, we search for policies offering more desirable trade-offs between quality and NFEs by employing techniques from differentiable Neural Architecture Search (NAS) \cite{liu2018darts}. 
Our NAS-based search suggests unnecessary computations take place in the latter part of the denoising process.
% a simple policy with minimal harm to quality, whereby \cfg is truncated to the initial denoising iterations.
We draw upon this finding, and propose an adaptive version of \cfg that we call ``Adaptive Guidance'' (\ourmethod).
Our \ourmethod policy is an efficient variant of \cfg that enjoys the image quality of \cfg despite requiring 25\% fewer NFEs. %  while preserving .
Please refer to Fig.~\ref{fig:teaser} for an illustration of the generation quality of \ourmethod. 
% We further boost the quality of this policy back to that of CFG by noting that unconditional updates can be predicted with negligible error via inexpensive \textit{linear combinations} of previous updates.
% We thus incorporate these linear combinations into the denoising process and term the resulting policy \textit{``Truncated Guidance''} (\ourmethod).
% Finally, we demonstrate the capacity of \ourmethod to replicate the quality of \cfg while requiring $\approx$~25\% fewer NFEs.
Compared to efficiency-oriented techniques like guidance distillation~\cite{meng2023distillation}, \ourmethod is easy to implement, is training-free, and preserves the capacity to handle negative prompts.
% Importantly, \ourmethod is very easy to implement, making it accessible for practitioners. 
Finally, we propose \ourlinearmethod, a fast version of \ourmethod that estimates updates required by \ourmethod as a linear combination of past iterates.
\ourlinearmethod provides further reductions in computation at the cost of imperceptible losses in sample quality.

In summary, our contributions are threefold:
\begin{itemize}
    \item We show that techniques from gradient-based Neural Architecture Search (NAS) can be leveraged in the context of sampling from denoising diffusion models to discover efficient guidance policies.
    \item We propose an efficient and general plug-and-play alternative to Guidance Distillation that achieves $50\%$ of the speed-ups while offering the ability to handle dynamic negative prompts and image editing.
    \item We discover that regularities across diffusion paths enable the replacement of certain NFEs in \cfg with affine transformations of past iterates. This observation enables further runtime reductions and constitutes an interesting starting point for future research.
    
    % \item Our study offers valuable insights into the denoising mechanism, revealing that truncating \cfg beyond a specific point can closely approximate a DM with negligible error.
    
    % \item We present ``Truncated Guidance'' (\ourmethod), a novel method that only uses \cfg where crucial and predicts unconditional updates in the denoising process using inexpensive linear combinations. 
    % \ourmethod reduces network evaluations by about 25\% while maintaining the high-quality output of \cfg, showcasing a significant advancement in diffusion model efficiency. 
    % Our findings further indicate that the guidance policies identified in a given model have broad applicability, allowing their plug-and-play use in a wide range of classifier-free guided diffusion models.
\end{itemize}

% Upon acceptance of the paper, we will release our PyTorch~\cite{NEURIPS2019_9015} implementation.
 Find our full project and more resources on \href{https://bcv-uniandes.github.io/adaptiveguidance-wp/}{bcv-uniandes.github.io/adaptiveguidance-wp/.
}

% TRADE OFF

% VERY EASY TO IMPLEMENT

% Importantly, we also show that the guidance policies found in a given model generalize beyond this particular model. Thus enabling their plug-and-play applicability for practitioners to the entire class of classifier-free guided diffusion models.

% (Defend against guidance -> get an estimate of the time [64GPUs] DOUBLE CHECK WITH ARTSIOM.)

% NUMBERS FOR BS1 and BS2
\section{Related Work}
\label{sec:related_works}

\subsection{Fast Inference with Diffusion Models}
% Broadly speaking
Diffusion models \cite{sohl2015deep,ho2020denoising,nichol2021improved} achieve density estimation and sampling by modeling a reversible transport map $T$ that pushes forward a base distribution $p_b$ that is % easy to sample 
tractable
(usually a standard Gaussian) to a target distribution $p_*(\mathbf{x})$, \ie, $T \# p_b = p_*(\mathbf{x})$. 
In contrast to traditional measure transport approaches (\eg, \cite{dinh2016density,kingma2018glow,chen2018neural}), diffusion models do not parameterize $T$ explicitly but rather learn it implicitly from the reverse direction of a gradual noising process. 
This approach has the benefit of the transport $T$ being learnable without the need for simulation. % -free. 
However, it also suffers from having higher inference costs due to the iterative nature of the sampling process.

Thus, a large body of work has focused on producing faster and more efficient ways of sampling from diffusion models.
One angle of attack is the solver employed for integrating the differential equations that underlie the diffusion process. 
For example, methods based on exponential integrator \cite{lu2022dpm,lu2022dpmplus}, higher order solvers \cite{zhao2023unipc, karras2022elucidating,zhang2022fast} or model-specific bespoke solvers \cite{shaul2023bespoke,zheng2023dpm} have been proposed. 
Orthogonal to these efforts, \cite{shih2023parallel} proposes parallelizing sampling via fixed-point iterations. 
Another common goal of exploration is reducing the size of the neural network that performs denoising \cite{peebles2023scalable,yang2023diffusion,li2023snapfusion}. 
For example, \cite{yang2023diffusion} explores ways of distilling a large teacher network into a smaller, more efficient, student. 
Yet, another set of papers explores ways of reducing the size of the diffusion's latent space \cite{gu2022vector,ho2022cascaded,rombach2022high,rampas2022fast}. 
Recently, a line of research explored reformulations of the diffusion process in order to reduce curvature in both the forward (noising) \cite{albergo2023stochastic,lipman2022flow} and backward (de-noising) trajectories \cite{liu2022flow,pooladian2023multisample,lee2023minimizing, karras2022elucidating}, which allows for larger solver steps even when employing lower-order solvers. 
Along these lines, \cite{salimans2022progressive} proposes to progressively reduce the number of diffusion steps by distillation. 

Within the field of accelerating diffusion models, AutoDiffusion \cite{li2023autodiffusion} is conceptually similar to our study in the sense that they employ a neural architecture search-inspired algorithm to improve the runtime of a pre-trained diffusion model. 
In contrast to AutoDiffusion, our method employs a more efficient gradient-based search instead of an evolutionary one. 
Furthermore, we optimize per-step guidance options, while AutoDiffusion focuses on time schedule and network architecture.

\subsection{Conditioning Diffusion Paths}
For both image generation and editing, the most challenging and practical cases involve some form of conditioning. 
Inspired by the success of class-conditioning in GANs (\eg,~\cite{odena2017conditional}), \cite{dhariwal2021diffusion} proposes to enhance the estimates of the diffusion probability path $p_t(\mathbf{x}|\mathbf{c})$ with the gradient of an image classifier $p_\theta(\mathbf{c}|\mathbf{x})$. 
Similarly, \cite{nichol2021glide} proposes to use CLIP guidance for text-to-image generation with diffusion models. 
Yet, both approaches are prone to adversarial outcomes (\ie, degenerate solutions) and struggle with the domain shift between the noisy images of the diffusion sampling process and the clean images on which the guidance models are trained.

In their seminal work, Ho~\etal~\cite{ho2022classifier} show that the diffusion process can be successfully conditioned in a ``classifier-free'' manner by leveraging implicit priors of the diffusion model itself. 
Toward this end, Ho~\etal jointly train a network to predict both unconditional and conditional scores. 
During generation, the two scores are combined, giving rise to the technique known as \cfg, to pinpoint samples with high conditional probability, as given by the inverted diffusion model as implicit classifier\footnote{While implicit classifiers are generally imperfect, especially when the model does not perfectly capture the data distribution (see \cite{grandvalet2004semi}, for instance), the efficacy of \cfg remains unambiguously evident in practice.}. 
Unfortunately, by definition, the \cfg scheme requires two, instead of one, NFEs per step, which doubles the sampling latency of the diffusion process on state-of-the-art models that max out GPU parallelization on a single sample.

Guidance Distillation (\textsc{Gd}) \cite{meng2023distillation} elegantly mitigates the need for an additional unconditional forward pass. However, \textsc{Gd} requires re-training as well as re-evaluation, both of which are resource-intensive.\footnote{To achieve comparable performance to \cfg, Guidance Distillation on EMU-768 requires around 10k iterations with a batch size of $32$, which amounts to roughly four GPU days on A100.} 
Moreover, this technique cannot handle dynamic negative prompts, which are an important asset for responsible AI. It also does not work with compositional guidance \cite{liu2022compositional}, which is, for instance, used in text-to-3D generation \cite{poole2022dreamfusion}. Finally, it is unclear how to generalize \textsc{Gd} to multimodal conditioning employed, for example, in image editing \cite{brooks2023instructpix2pix,sheynin2023emu}.

In this work, we propose plug-and-play alternatives to Guidance Distillation that achieve $50\%$ of the speed-ups at equal sample quality while conceptually omitting the aforementioned problems. For example, \ourmethod accommodates negative prompts (in Sec. \ref{sec:truncated_guidance}), image editing (in Appendix~\ref{appx:editing}), is training-free, and exactly replicates the outputs of a given baseline such that no re-evaluation is needed.

%guidance-rescale \cite{lin2023common}
\subsection{Neural Architecture Search}
Neural Architecture Search (NAS) aims at automating the design of neural network architectures by conceptualizing the network as a Directed Acyclic Graph (DAG) and exploring different layers as its nodes \cite{zoph2016neural,zoph2018learning,pham2018efficient,liu2018progressive,brock2017smash}. 
We focus this review on differentiable NAS methods \cite{liu2018darts,li2020sgas,li2022lc,wu2019fbnet}.
The DARTS framework~\cite{liu2018darts} is particularly relevant to our work, as it introduces a continuous relaxation of the layer representation, allowing architecture search to be differentiable and, hence, more efficient. 
Here, we leverage analogies between neural network design and the diffusion process by unrolling the diffusion process' graph in the time dimension, and thus considering each step as a distinct node in the DAG. % by considering each diffusion step as a distinct node in the directed graph and searching for the optimal guidance option of each node.
This allows us to directly apply DARTS to search for an optimal guidance option at each node.

\section{Background on Diffusion Models}

\newcommand{\nicewidth}{0.24}

As introduced in Sec. \ref{sec:related_works}, diffusion models generate images by reversing a pre-defined noising process. 
In particular, when the noising process is an Ornstein-Uhlenbeck process, the continuous time limit of the forward SDE reads as $d\mathbf{x} = \mathbf{f}(\mathbf{x},t)\:dt  + g(t)\:d\mathbf{w}$, where $f(\mathbf{x},t): \mathbb{R}^d\rightarrow \mathbb{R}^d$ is a vector-valued drift coefficient, $g(t): \mathbb{R}\rightarrow\mathbb{R}$ is the diffusion coefficient of $\mathbf{x}(t)$ and $\mathbf{w}$ is standard Brownian motion.  
Anderson's Theorem \cite{anderson1982reverse} states that, under mild assumptions, this SDE satisfies a reverse-time process:
\begin{equation}\label{eq:reverse_sde}
d\mathbf{x} = \left[ \mathbf{f}(\mathbf{x}, t) - g(t)^2 \nabla_{\mathbf{x}} \log p_t(\mathbf{x}) \right] dt + g(t)\:d\mathbf{\bar{w}},
\end{equation}
where $\mathbf{\bar{w}}$ is the reverse-time Brownian motion.
As shown in~\cite{hyvarinen2005estimation, song2020score}, the marginal transport map % $T$ 
can be learned~(in expectation) by maximum likelihood estimation of the scores of individually diffused data samples $\nabla_\mathbf{x} \log p_t(\mathbf{x})$ in a simulation-free manner. 
This map is commonly learnt by optimizing the parameters $\theta$ of a time-conditioned neural network that produces score estimates $\epsilon_{\theta}(\mathbf{x}_t,t)$.\footnote{For brevity's sake, we omit the conditioning of $\mathbf{\epsilon}$ on $t$ going forward.}

As shown in \cite{song2020score}, the SDE in Eq.~\eqref{eq:reverse_sde} has a deterministic counterpart (\ie, an ODE) that enjoys equivalent marginal probability densities:
\setlength{\abovedisplayskip}{3pt}
\setlength{\belowdisplayskip}{3pt}
\begin{equation}\label{eq:pflowode}
d\mathbf{x} = \left[\mathbf{f}(\mathbf{x},t) - \frac{1}{2} g(t)^2 \nabla_{\mathbf{x}} \log p_t(\mathbf{x})\right]dt.
\end{equation} % reference in sect w/ experiments!!!
%This formulation has the advantage of allow for exact likelihood computation via the change of variable formula \cite{chen2018neural}.
Solving Eq.~\eqref{eq:pflowode} generally yields better results when fewer discretization steps are taken~\cite{karras2022elucidating}.

% \albert{All SDE formulation is introduced to end up with ODE. Reader doesn't need all the SDE extra formulation to understand the paper.}
% \jonas{Yes but it sounds cool and our methods are not restricted to ODE case.}
%(https://arxiv.org/pdf/2307.04102.pdf#:~:text=Generative%20models%2C%20such%20as%20normalizing,Gaussian)%20to%20a%20target%20distribution.)
 
% \begin{equation}
% dx = \left[ \mathbf{f}(\mathbf{x}, t) - \frac{1}{2} g(t)^2 \nabla \log p_{t}(\mathbf{x}|\mathbf{c}) \right] dt.
% \end{equation}

\paragraph{Conditional generation with diffusion models.}
The diffusion framework can be extended to allow for conditional generation by learning the score $\log p_{t}(\mathbf{x}|\mathbf{c})$, where $\mathbf{c}$ is, for example, a class- or text-condition. %, allows to sample from $p(\mathbf{x}|\mathbf{c})$.
% Inspired by the success of class-conditioning in GANs (\eg, \cite{odena2017conditional}), \cite{dhariwal2021diffusion} proposes to enhance the conditional score estimate with the gradient of an image classifier $p_\theta(\mathbf{c}|\mathbf{x})$. 
% Along similar lines, \cite{nichol2021glide} put forward the use of CLIP guidance for text-to-image generation. 
% Yet, both approaches are prone to adversarial outcomes and struggle with the domain shift between the noisy images used by the diffusion sampling process and the clean images on which the guidance models are trained.
Current state-of-the-art models for conditional generation employ ``Classifier-Free Guidance''~(\cfg)
% In their seminal work, Ho~\etal~
\cite{ho2022classifier},
% show that the diffusion process can be successfully conditioned on priors in a ``classifier-free'' manner. 
a technique in which both the conditional and unconditional scores are linearly combined to denoise the sample.
% Namely, they jointly learn unconditional and conditional scores by randomly setting $\mathbf{c}$ to an unconditional prompt token $\emptyset$. 
% During sampling, the trajectory is incentivized to pinpoint samples with high conditional probability $ \log p(\mathbf{c}|\mathbf{x})$ by following the score estimate
In particular, \cfg proposes to follow the score estimate given by
\begin{equation}\label{eq:cfg-score}
    \mathbf{\epsilon}_{\text{cfg}}(\mathbf{x}_t,\mathbf{c},s) =\mathbf{\epsilon}_\theta(\mathbf{x}_t, \emptyset) + s \cdot (\mathbf{\epsilon}_\theta(\mathbf{x}_t, \mathbf{c}) - \mathbf{\epsilon}_\theta(\mathbf{x}_t, \emptyset)), 
\end{equation}
where $\emptyset$ is the unconditional prompt token, and $s>1$ indicates the guidance strength. 
While this new score may not directly reflect the gradient of a classifier's log-likelihood, it is inspired by the gradient of an implicit classifier $p'(\mathbf{c}|\mathbf{x}) \propto p(\mathbf{x}|\mathbf{c})/p(\mathbf{x})$. 
As a result, $\nabla_x \log p(\mathbf{c}|\mathbf{x}) \propto \nabla_\mathbf{x}\log p(\mathbf{x}|\mathbf{c}) - \nabla_\mathbf{x}\log p(\mathbf{x}) $ and hence $  \epsilon_{\lambda}(\mathbf{x}_t,\mathbf{c}) \propto \epsilon(\mathbf{x}_t, \emptyset) + s \cdot \nabla_\mathbf{x} \log p(\mathbf{x}|\mathbf{c})$. 
In that sense, \cfg shifts probability mass toward data where an implicit classifier $p'(\mathbf{c}|\mathbf{x})$ assigns a high likelihood to the condition $\textbf{c}$.

%Geometrically, classifier free guidance can be interpreted as amplifying the conditioning by moving score estimate towards and beyond the conditional score from the viewpoint of the unconditional.
Notably, evaluating Eq.~\eqref{eq:cfg-score} introduces an extra NFE compared to unguided sampling, which may up to double the latency. 
Next, we search for efficient ways of guiding the denoising process, aiming at reducing NFEs while retaining the benefits of \cfg. 
In the following sections, we discuss these approaches along with their respective results. 

\begin{figure*}[ht]
    \centering
    % First row
    \raisebox{1.1\height}{\begin{sideways}\ourmethod\textbf{(ours)}\end{sideways}}
    \begin{subfigure}{\nicewidth\textwidth}
        \includegraphics[width=\linewidth]{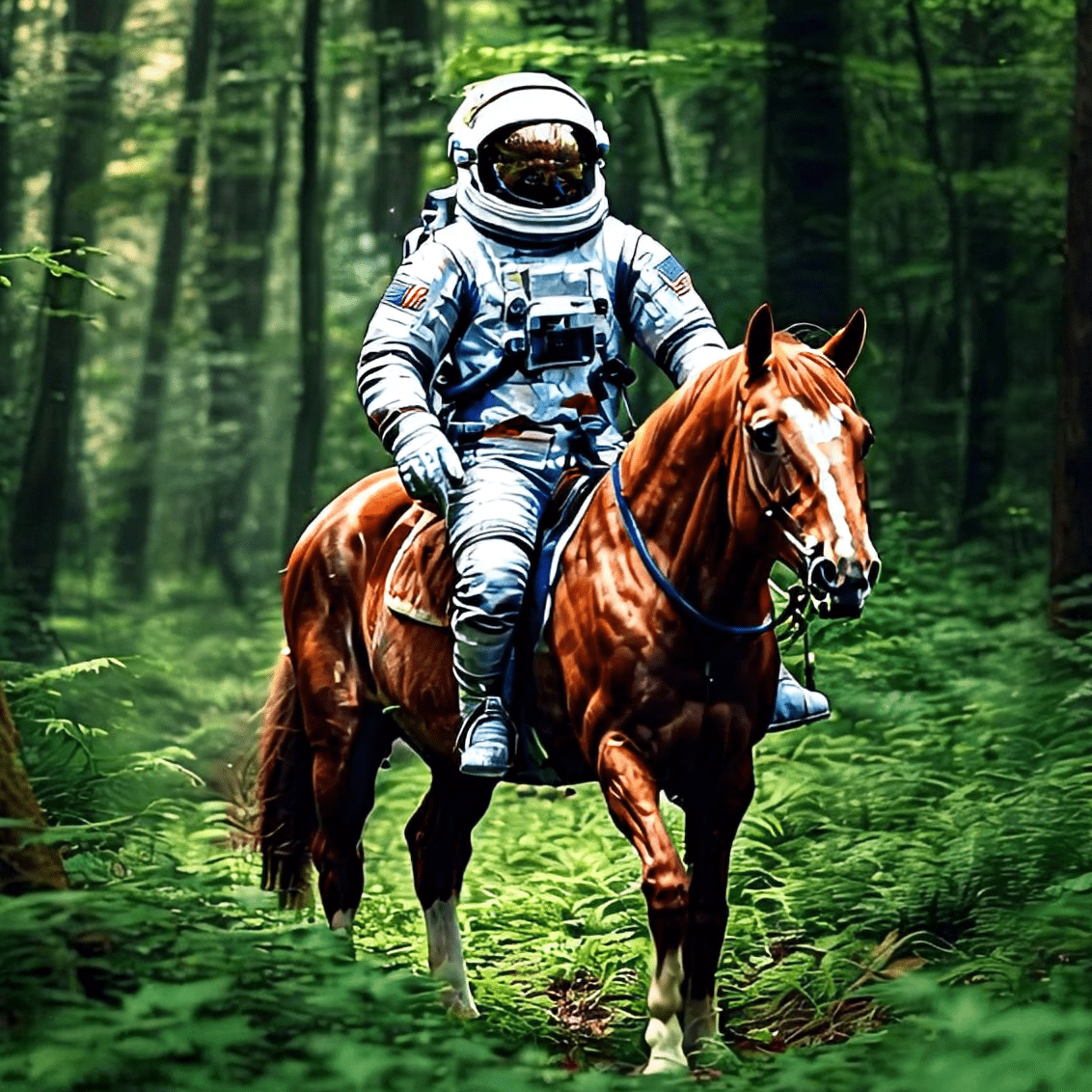}
        \caption{$\Bar{\gamma}=\infty$ (40NFEs)}
    \end{subfigure}
    \hfill\small
    \begin{subfigure}{\nicewidth\textwidth}
        \includegraphics[width=\linewidth]{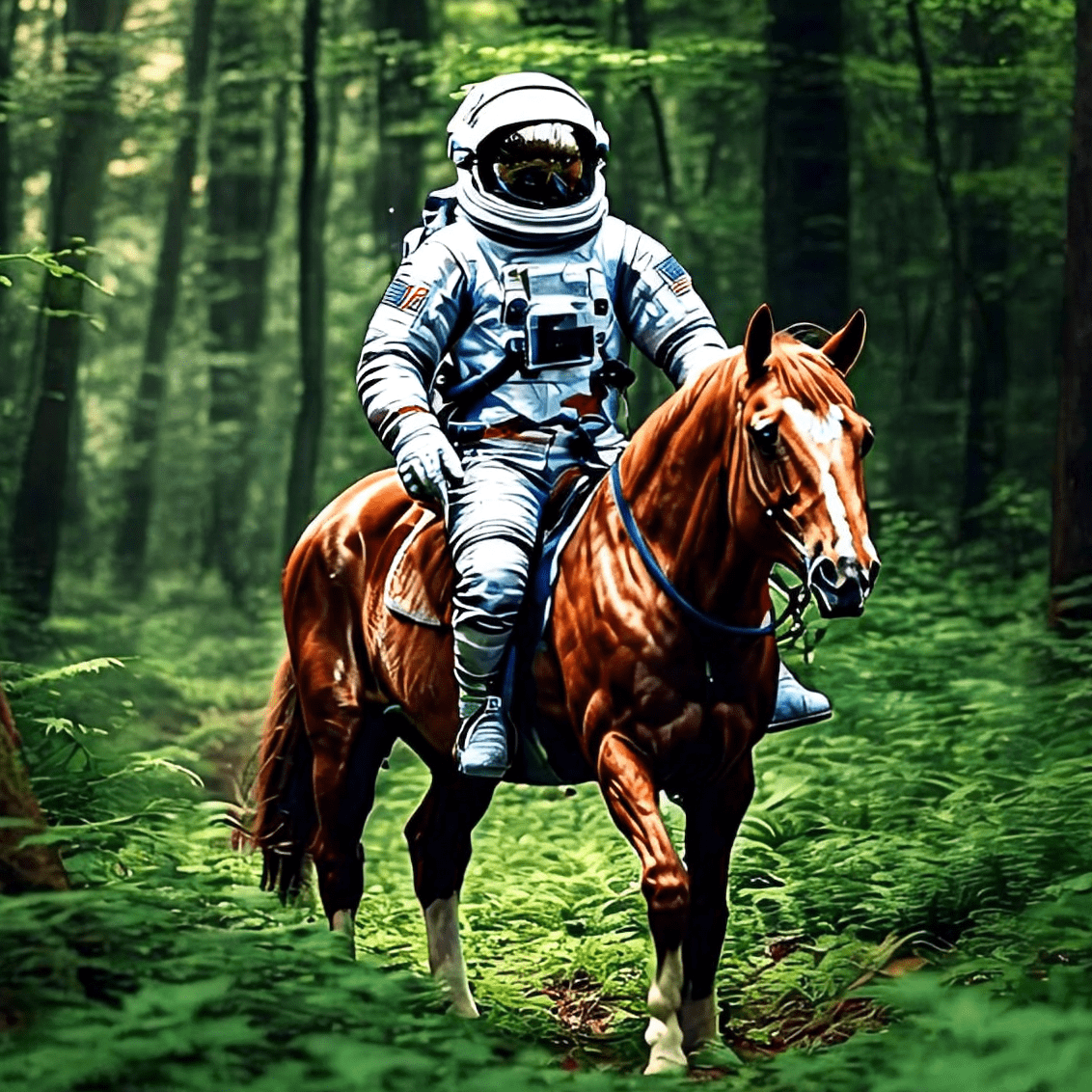}
        \caption{$\Bar{\gamma}=0.993$ (32NFEs)}
    \end{subfigure}
    \hfill 
    \begin{subfigure}{\nicewidth\textwidth}
        \includegraphics[width=\linewidth]{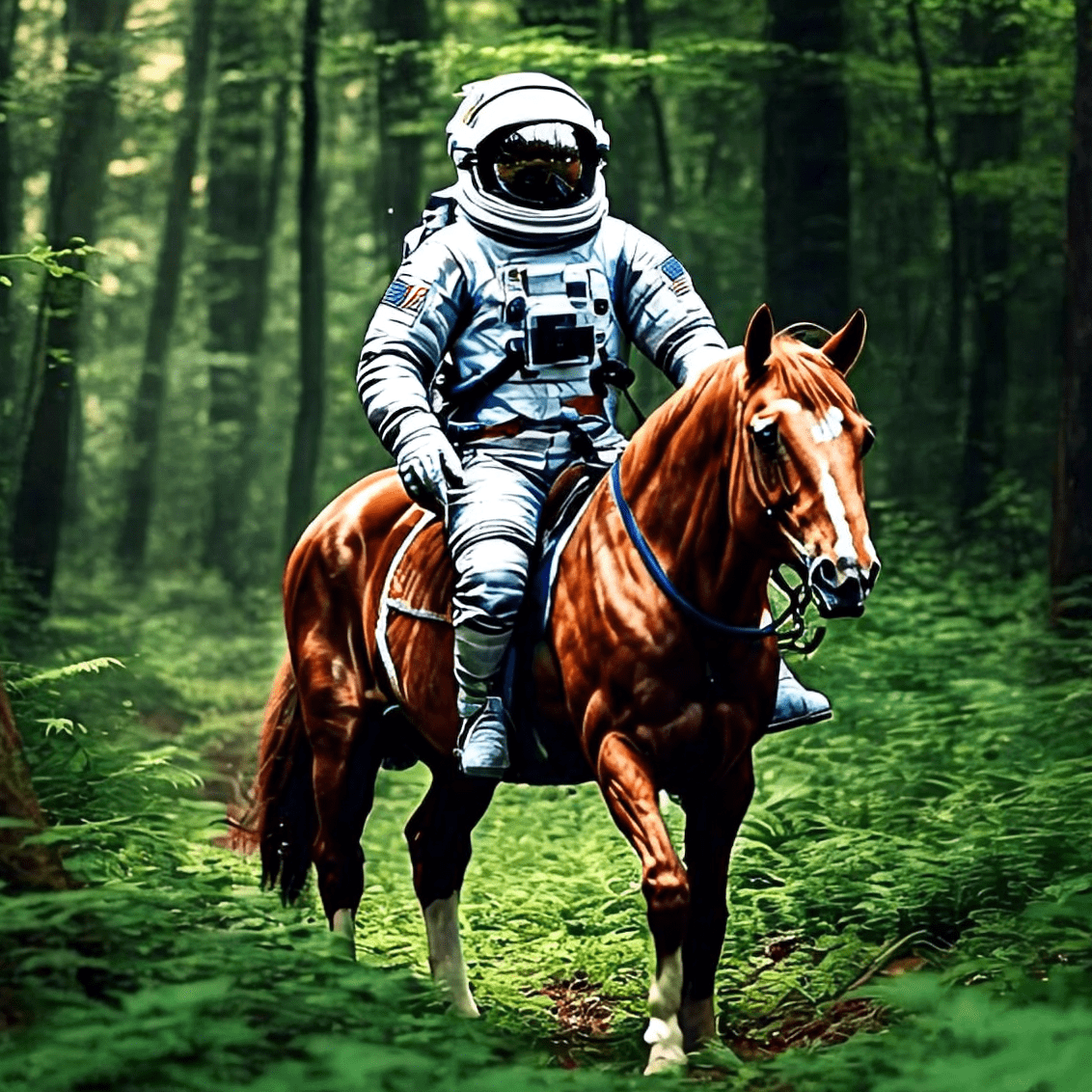}
        \caption{$\Bar{\gamma}=0.991$ (30NFEs)}
    \end{subfigure}
    \hfill
    \begin{subfigure}{\nicewidth\textwidth}
        \includegraphics[width=\linewidth]{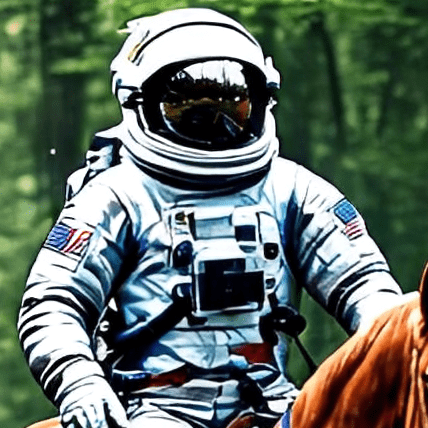}
        \caption{Zoom of image to the left}
    \end{subfigure}
    \hfill
    % \begin{subfigure}{0.19\textwidth}
    %     \includegraphics[width=\linewidth]{figures/truncated_guidance/cow_05.jpg}
    %     \caption{T\textsc{g}, $\Bar{\gamma}=0.99$ (30NFEs)}
    % \end{subfigure}

    \vspace{-0.02cm} % space between rows

    % Second row
    \raisebox{3.6\height}{\begin{sideways}\textsc{Cfg}\end{sideways}}
    \begin{subfigure}{\nicewidth\textwidth}
        \includegraphics[width=\linewidth]{figures/fig1/astronaut_20.png}
        \caption{ $steps=20$ (40NFEs)}
    \end{subfigure}
    \hfill
    \begin{subfigure}{\nicewidth\textwidth}
        \includegraphics[width=\linewidth]{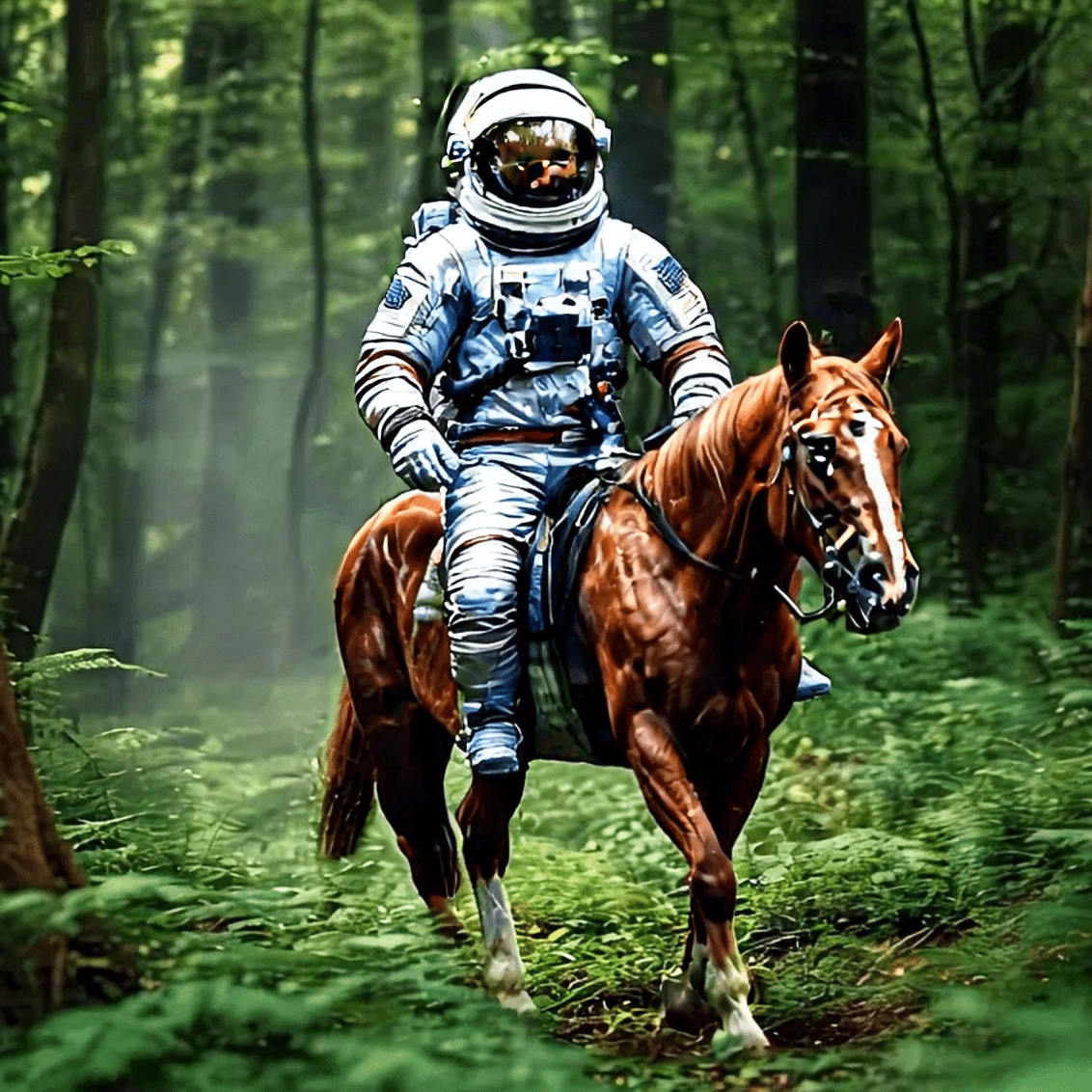}
        \caption{$steps=16$ (32NFEs)}
    \end{subfigure}
    \hfill
    \begin{subfigure}{\nicewidth\textwidth}
        \includegraphics[width=\linewidth]{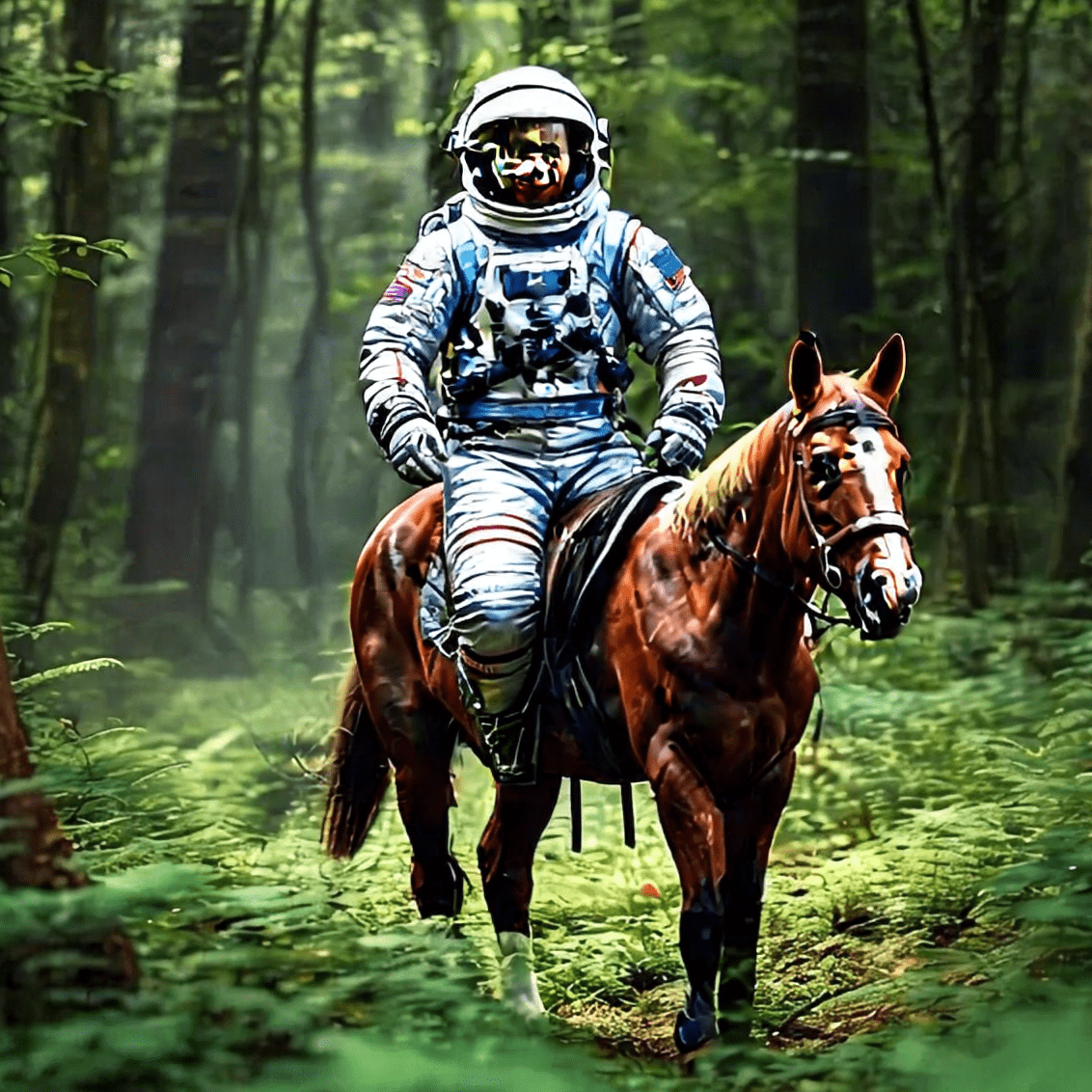}
        \caption{$steps=15$ (30NFEs)}
    \end{subfigure}
    \hfill
    \begin{subfigure}{\nicewidth\textwidth}
        \includegraphics[width=\linewidth]{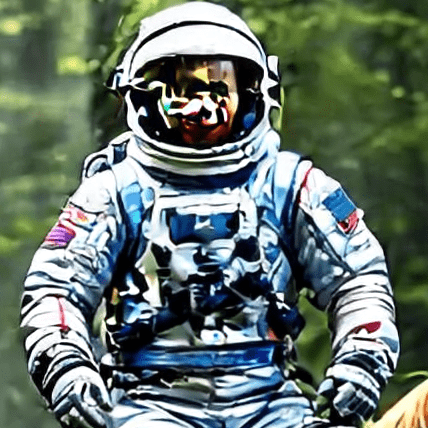}
        \caption{Zoom of image to the left}
    \end{subfigure}
    \hfill
    % \begin{subfigure}{0.19\textwidth}
    %     \includegraphics[width=\linewidth]{figures/truncated_guidance/gt_cow_15.jpg}
    %     \caption{\textsc{Cfg}, $s=15$ (30NFEs)}
    % \end{subfigure}
    \vspace{-0.3cm}
    \caption{\textbf{Adaptive Guidance (\ourmethod) \textit{vs.} Classifier-Free Guidance (\textsc{Cfg}) for multiple Number of Function Evaluations (NFEs).} \footnotesize For \ourmethod we keep the number of denoising iterations constant but reduce the number of steps using \cfg by increasing the threshold $\bar{\gamma}$ (top). \cfg simply reduces the total number of diffusion steps (bottom). Vertically aligned samples require the exact same number of NFEs. As can be seen, \ourmethod replicates the baseline very closely while \textsc{Cfg} with less steps introduces artifacts.} %\juan{couldnt we just call the first $\gamma = 1$ instead of $\infty$?} jonas: NO, can be 1 before last step.}
\label{fig:figure1}
\end{figure*}

\section{Gradient Search along Diffusion Dynamics}
\paragraph{Design space for guided diffusion steps.}
We assume access to a pre-trained diffusion model $\Phi : X \times C \rightarrow X$
% \begin{equation}
% \Phi : X \times C \rightarrow X
% \end{equation}
working in latent space \( X = \mathbb{R}^{H \times W \times C} \) and condition space \( C \), where \( \mathbf{c} \in C \) is a condition, \eg, a text prompt. 
Initializing \( \mathbf{x}_T \sim p_b \), where \( p_b \) represents a % i.i.d. 
Gaussian distribution, setting a condition \( \mathbf{c} \in C \) and a time-schedule $\tau=\{T,T-1,...,0\}$, the diffusion model builds a sequence of latent codes
\begin{equation}\label{eq:generation}
\{\mathbf{x}_t\}_{t=0}^T\: \text{s.t.} \:  \mathbf{x}_{T}\sim p_b,\: \mathbf{x}_{t-1} = \Phi(\mathrm{solver}(\bar{\mathbf{x}}_t)),
\end{equation}
where $\mathrm{solver}$ represents an ODE solver for Eq.~\eqref{eq:pflowode}. 
The model $\Phi$ operates under classifier-free guidance as given in Eq.~\eqref{eq:cfg-score}, \ie, $\bar{\mathbf{x}}_t= \epsilon_\text{cfg}(\mathbf{x_t},\mathbf{c},s)$ and $s$ is constant over time. 
While this setup is the default in most popular diffusion models \cite{rombach2022high,dai2023emu,saharia2022photorealistic,nichol2021glide}, we highlight that multiple alternatives exist for \( \bar{\mathbf{x}}_t \) at any given \( t \), each associated with different computational costs:
% Definition of the choice sets

\begin{tabular}{@{\textbullet\hspace{\tabcolsep}} l l c}
    Unconditional score: & $\epsilon_\theta(\mathbf{x}_t, \emptyset)$ & (1 NFE) \\
    Conditional score: & $\epsilon_\theta(\mathbf{x}_t, \mathbf{c})$ & (1 NFE) \\
    \cfg score: & $\epsilon_{\text{cfg}}(\mathbf{x}_t, \mathbf{c}, s_t)$ & (2 NFEs) \\
\end{tabular}

Here, $\epsilon_{\theta}$ represents a neural network parameterized by frozen weights $\theta$, and $s_t$ is no longer constant in time. Denote by $f_t$ the particular step \textit{choice} at time $t$ with
$f_t\in 
\mathcal{F}_t= \left\{ \epsilon_\theta(\mathbf{x}_t, \emptyset), \epsilon_\theta(\mathbf{x}_t, \mathbf{c}), \epsilon_{\text{cfg}}(\mathbf{x}_t,\mathbf{c},s_t) \right\}.$ 
Then, the search space for the complete diffusion process is given by:
$
\mathcal{S} = \prod_{t=0}^{T} \mathcal{F}_t,
$ where the product symbol denotes the Cartesian product over sets. 

As a result, $\mathcal{S}$ is the set of all possible sequences of choices $\zeta=(f_0, f_1, \dots, f_T)$, which we henceforth refer to as \textit{policies}. 
Clearly, $\mathcal{S}$ is unbounded as long as $s_t\in\mathbb{R}$. 
Although this fact is not problematic in itself for gradient-based search, we constrain $s_t$ to be in a bounded and finite set $\mathcal{S}=\{s^1,...,s^k\}$ in order to obtain simpler and more generalizable policies. 
As a result, the search spaces contain a total of $|\mathcal{S}|=| \prod_{t=0}^{T} \mathcal{F}_t|=(2+k)^{T+1}$ different policies. 

\paragraph{Enabling backpropagation with soft alphas.}
Searching $\mathcal{S}$ for policies with a good performance-latency trade-off constitutes a large-scale combinatorial problem, especially since $T$ is usually in the range of $20$ to $50$. 
Thus, inspired by the literature on NAS, we relax the discrete search into a continuous one. 
This decision allows for effectively using gradients to navigate the high-dimensional search space. 
In particular, for each set of choices $\mathcal{F}_t$, we introduce a trainable vector $\bm{\alpha}_t \in \mathbb{R}^{k+2}$ and obtain the solver input as a softmax weighting of the individual options %\juan{are we gonna mention the temperature hyper-parameter?} jonas: I vote no for simplicity
\setlength{\abovedisplayskip}{3pt}
\setlength{\belowdisplayskip}{3pt}
\begin{equation}\label{eq:options}
   \bar{\mathbf{x}}_t:= \text{softmax}\left(\bm{\alpha}\right)^\intercal \mathcal{F}_t %\sum_{i=1}^{k+2} \frac{e^{\bm{\alpha}_t^i}}{\sum_{j=1}^{k+2} e^{\bm{\alpha}_t^j}} f_t^i
\end{equation}
Once trained, the score matrix $\bm{\alpha}:= [\bm{\alpha}_T^\intercal,...,\bm{\alpha}_0^\intercal]$ represents a multinominal distribution over the per-iteration options~$(\mathcal{F}_T,...,\mathcal{F}_0)$ from which we can sample concrete policies $\zeta$. In the following, we define a differentiable objective to guide our search for efficient and effective guidance policies.

\paragraph{Search objective.}
We seek a policy $\zeta$ that gives rise to a diffusion model that replicates $\Phi$ as closely as possible, as quantified by a differentiable metric $d: X \times X \rightarrow [0, \infty)$ that measures the distance between the endpoints of the two diffusion paths ($\mathbf{x_0}'$ and $\mathbf{x_0}$, respectively). 
Our goal is to achieve replication with fewer NFEs than the reference policy $f_t=\epsilon_{cfg}(\mathbf{x_t},\mathbf{c},s), \forall\:t$. 
Towards this end, we optimize:
\setlength{\abovedisplayskip}{3pt}
\setlength{\belowdisplayskip}{3pt}
\begin{equation}\label{eq:replication}
\bm{\alpha}^* = \text{argmin}_{\bm{\alpha}} \left[d(\mathbf{x}_0,\mathbf{x}_0'(\zeta(\bm{\alpha})) +\lambda g(\zeta(\bm{\alpha}))\right],
\end{equation}
where $\lambda>0$ and $g(\zeta(\bm{\alpha}))$ regularizes the sum of the scores obtained by passing $\bm{\alpha}$ through a Gumbel-softmax~\cite{maddison2016concrete} weighted by the per-choice costs ($1$ for unconditional/conditional steps and $2$ for \cfg steps with $s_t>1$). 
Thus, $g$ represents a (differentiable) proxy for the total NFEs of the policy $\zeta(\bm{\alpha})$. 
We employ a ReLU offset to a target cost limit $\bar{c}$ under which no penalty is employed.  

For the policy search, we initialize $\bm{\alpha}$ as i.i.d. uniform random variables. 
Subsequently, in each training iteration, we sample $\mathbf{x}_T\sim \mathcal{N}(0,I)$ and use our baseline model $\Phi$ to generate a target image $\mathbf{x}_0$. 
The same starting noise tensor is then being fed through a student model $\Phi'$ that mimics $\Phi$ but employs a soft alpha-weighted forward pass according to Eq.~\eqref{eq:options} to obtain $\mathbf{x}_0'(\zeta(\bm{\alpha}))$. 
Given these two images, we compute the differentiable loss in Eq.~\eqref{eq:replication} and backpropagate through $\Phi'$ w.r.t. $\bm{\alpha}$.\footnote{To cope with limited memory resources, we re-run certain forward-pass segments during backward (``activation checkpointing'').}

% \begin{figure}[htbp]
%     \centering
%     \includegraphics[width=\linewidth]{figures/synthesis_and_deltas/iterations.png}
%     \caption{Iterations}
%     \label{fig:image1}
% \end{figure}

% \begin{figure}[htbp]
%     \centering
%     \includegraphics[width=\linewidth]{figures/synthesis_and_deltas/differences.png}
%     \caption{Second Image}
%     \label{fig:image2}
% \end{figure}

\subsection{Experimental Setup}\label{sec:set-up}
We perform our guidance search in the context of text-to-image generation using the popular Stable Diffusion architecture~\cite{rombach2022high}, which we refer to as \textbf{LDM-512}.\footnote{We train LDM-512 from scratch on a commissioned dataset of images.}
This model has $900\text{M}$ parameters and generates images at a $512 \times 512$ resolution via a latent space of shape $4\times64\times64$.
To showcase that our findings generalize beyond the model they were searched on, we validate the found policies on a state-of-the-art EMU model \cite{dai2023emu}, which we refer to as \textbf{EMU-768}.
This model has $2.7\text{B}$ parameters, produces photorealistic images at a resolution of $768\times 768$, and uses a latent space of shape $16\times96\times96$. 

For training, we generate $10,000$ noise-image pairs randomly selected from the CC3M dataset \cite{song2019mass} using our \textbf{LDM-512} with $T=20$ DPM++~\cite{lu2022dpm} solver steps and a fixed guidance strength of $s=7.5$. 
In our search space $\mathcal{S}$ we include $k=3$ guidance strengths, which gives a total of five discrete choices: unconditional $\epsilon_\theta(\mathbf{x}_t, \emptyset)$, conditional $\epsilon_\theta(\mathbf{x}_t, \mathbf{c}),$ as well as $\epsilon_{\text{cfg}}(\mathbf{x}_t,\mathbf{c},a \cdot 7.5)$ for $a \in \{\tfrac{1}{2}, 1, 2\}$. 
We optimize Eq.~\eqref{eq:replication} with the Lion optimizer~\cite{chen2023symbolic} for $5$ epochs. All evaluation metrics are computed on a subset of $1,000$ prompts from the Open User Input (OUI) dataset 
\cite{dai2023emu}.

\subsection{Search Results}\label{sec:search_results}

\begin{figure}[t!]
    \centering
        \includegraphics[width=0.9\linewidth]{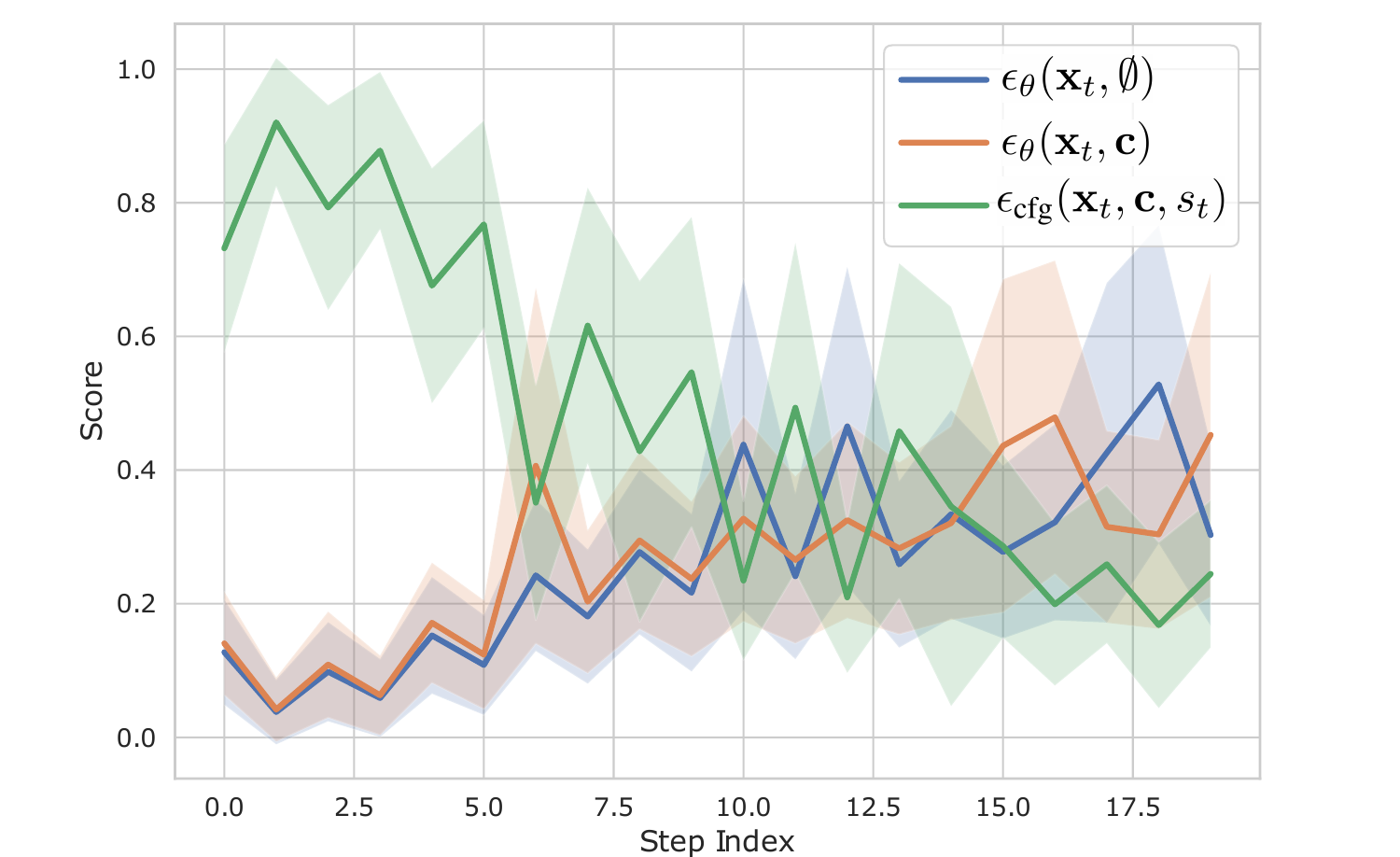}
    \vspace{-0.2cm}
    \caption{\textbf{Search results.} \footnotesize Average scores and standard deviations over steps in the diffusion process for the different guidance options. The 30 best searches are represented. As can be seen, \cfg is most important in the beginning, and the score decreases over time. }%\juan{make epsilon notation consistent. 30 best don't contain 2s and s/2 guidance}}
    \label{fig:replication_scores}
\end{figure}

Upon completion of our search, we find that the best-performing policies focused essentially on three guidance choices: conditional, unconditional, and \cfg with $s=7.5$.\footnote{In hindsight, this is not surprising as replicating a baseline model requires following the entire diffusion trajectory. Yet, by the design of the measured transport $T$, paths cannot cross, and there is no way of returning to the baseline once stepped off using a different guidance scale.} The score distribution of these policies is summarized in Figure~\ref{fig:replication_scores}. 
% We find that the evolution of the per-choice scores over the diffusion process. 
Notably, a distinct pattern emerges in the search results: namely, the importance assigned to \cfg is high in the first half of the denoising process but drops significantly in the second half. This fact follows intuition: text-conditioning is particularly important for determining the overall semantic structure of the image, and this semantic structure is set up early on in the diffusion process, while the later steps focus more on generating local information and high-frequency details (see \textit{e.g.}, Fig.~\ref{fig:semantics} in the Appendix).

\begin{figure}[t]
    \centering
    \begin{subfigure}{0.495\columnwidth} 
        \includegraphics[width=\linewidth, trim=10 10 50 50, clip]{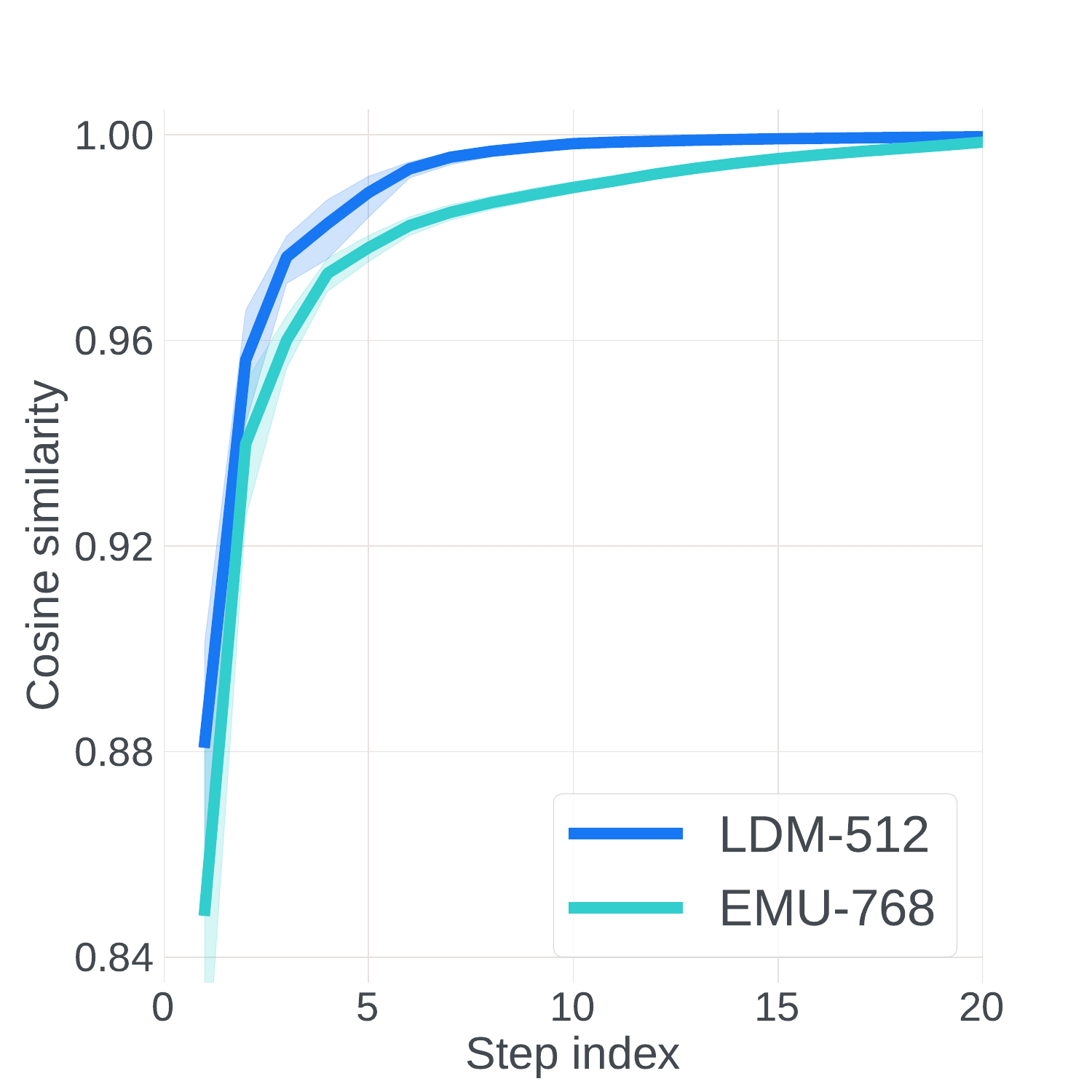}
        
    \end{subfigure}
    \begin{subfigure}{0.495\columnwidth}
        \includegraphics[width=\linewidth, trim=10 10 50 50, clip]{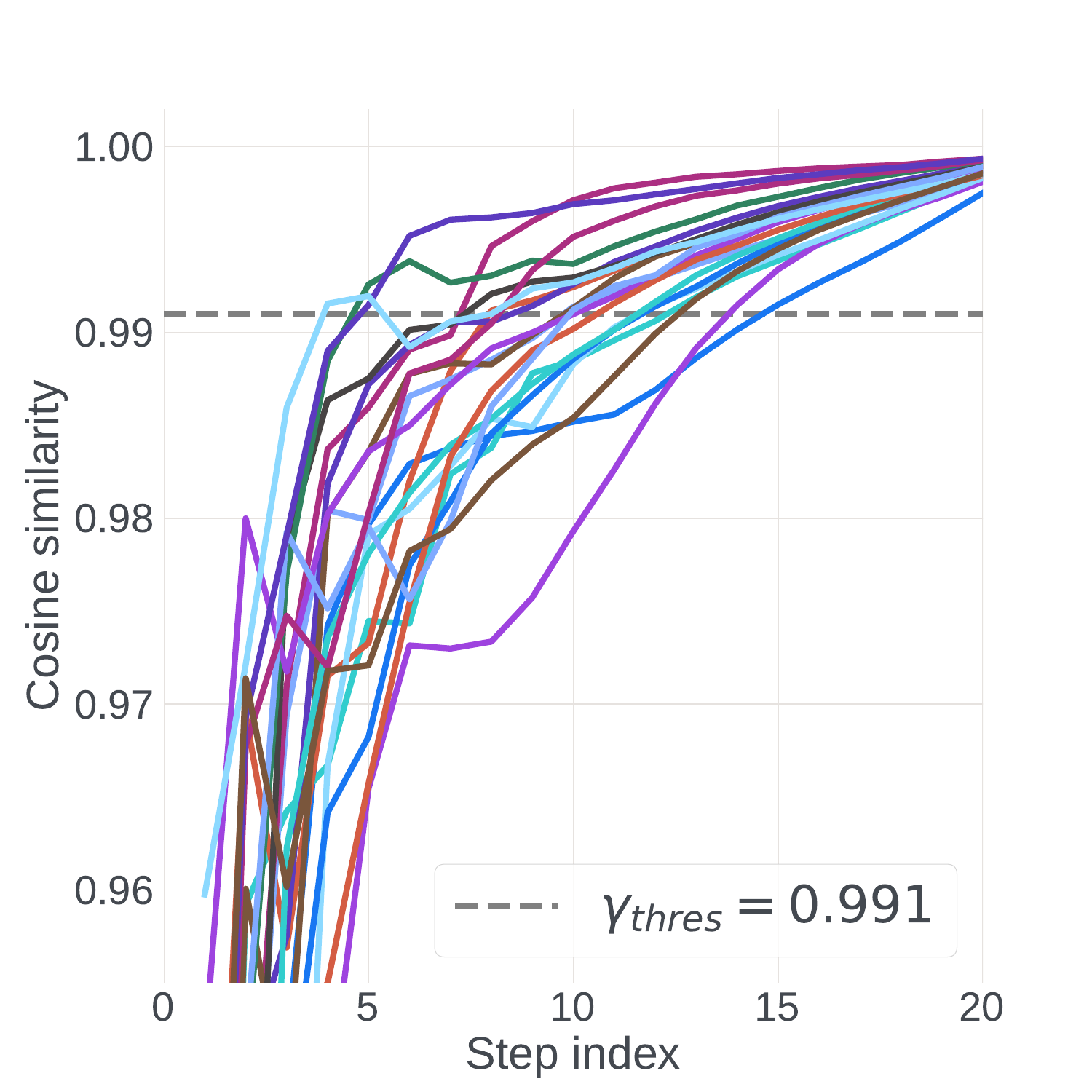}
        
    \end{subfigure}

    \caption{\textbf{Cosine similarities over time.} \footnotesize Left: Average cosine similarity $\gamma$ and $99\%$ confidence interval over $1,000$ IOU prompts for EMU and LDM. Right: Zoom to y-values in $[0.955,1.0]$ for 21 EMU samples. }%\albert{I would try to avoid these and the "search results" plot to be next to each other for aesthetics}}
    \label{fig:cos}
\end{figure}

Interestingly, this generative structure is mirrored in the inner workings of the diffusion process. Namely, the cosine similarity $\gamma_t$ between the conditional ($\epsilon(\mathbf{x_t},\mathbf{c})$) and unconditional ($\epsilon(\mathbf{x_t},\emptyset)$) network predictions increases almost monotonically over time. As shown in Fig.~\ref{fig:cos}, $\gamma_t$ achieves almost perfect alignment towards the end of the diffusion process.
That is, we empirically observe
\setlength{\abovedisplayskip}{3pt}
\setlength{\belowdisplayskip}{3pt}
\begin{equation}\label{eq:cos_sim}
\lim_{t \to 0} \left[ \gamma_t := \frac{\epsilon_{\theta}(\mathbf{x}_t,\mathbf{c}) \cdot \epsilon_{\theta}(\mathbf{x}_t,\emptyset)}{\| \epsilon_{\theta}(\mathbf{x}_t,\mathbf{c}) \| \| \epsilon_{\theta}(\mathbf{x}_t,\emptyset) \|} \right] = 1.
\end{equation}
In light of this finding, \ourmethod works because it stops guiding precisely when the conditional and unconditional update steps have converged, and guiding hence no longer introduces shifts in direction.

\section{Adaptive Guidance}\label{sec:truncated_guidance}
\paragraph{Definition.}
Section~\ref{sec:search_results} found that the conditional and unconditional updates become increasingly correlated over time.
This fact suggests an intuitive way to save NFEs by stopping \cfg computation when this correlation is high.
% simply truncating \textsc{Cfg} in the second half of the diffusion process can reduce computational redundancies without introducing significant deviations from the denoising path. 
We thus expand on this intuition to propose ``Adaptive Guidance'' (\ourmethod), a principled technique to decrease sampling cost while maintaining high image quality. 
In particular, \ourmethod adaptively switches from \cfg updates to (cheaper) conditional updates % and resort to the cheaper option of simple conditional steps ($\epsilon(\mathbf{x_t},\mathbf{c})$) as soon as 
when $\gamma_t$ (Eq.~\eqref{eq:cos_sim}) exceeds a threshold $\Bar{\gamma}$, where $\Bar{\gamma}\in [0,1]$ is the only hyper-parameter of \ourmethod. 
As a result, \ourmethod results in uncomplicated policies such as
\begin{equation*}
\zeta_{\ourmethod}=[\epsilon_{\text{cfg}}(\mathbf{x}_T, \mathbf{c}), \text{...}, \epsilon_{\text{cfg}}(\mathbf{x}_t, \mathbf{c}), \epsilon_\theta(\mathbf{x}_{t-1}, \mathbf{c}), \text{...},\epsilon_\theta(\mathbf{x}_{0}, \mathbf{c})],
\end{equation*}
where the truncation point $t$ is a function of  $\Bar{\gamma}$, the starting seed $\mathbf{x}_T$ and the conditioning $\mathbf{c}$. 

We highlight that $\zeta_{\ourmethod}$ is independent of the particular time schedule $\tau$ (Eq.~\eqref{eq:generation}) and solver used for the sampling process. Hence, it can be within a wide class of diffusion models and for arbitrary numbers of diffusion steps. Importantly, as shown in Fig.~\ref{fig:cos}, the cosine similarity trend found on LDM-512 generalizes to the much larger EMU-768 model.

\begin{figure}[t!]
    \centering
        \includegraphics[width=\linewidth,trim=25 5 25 25, clip]{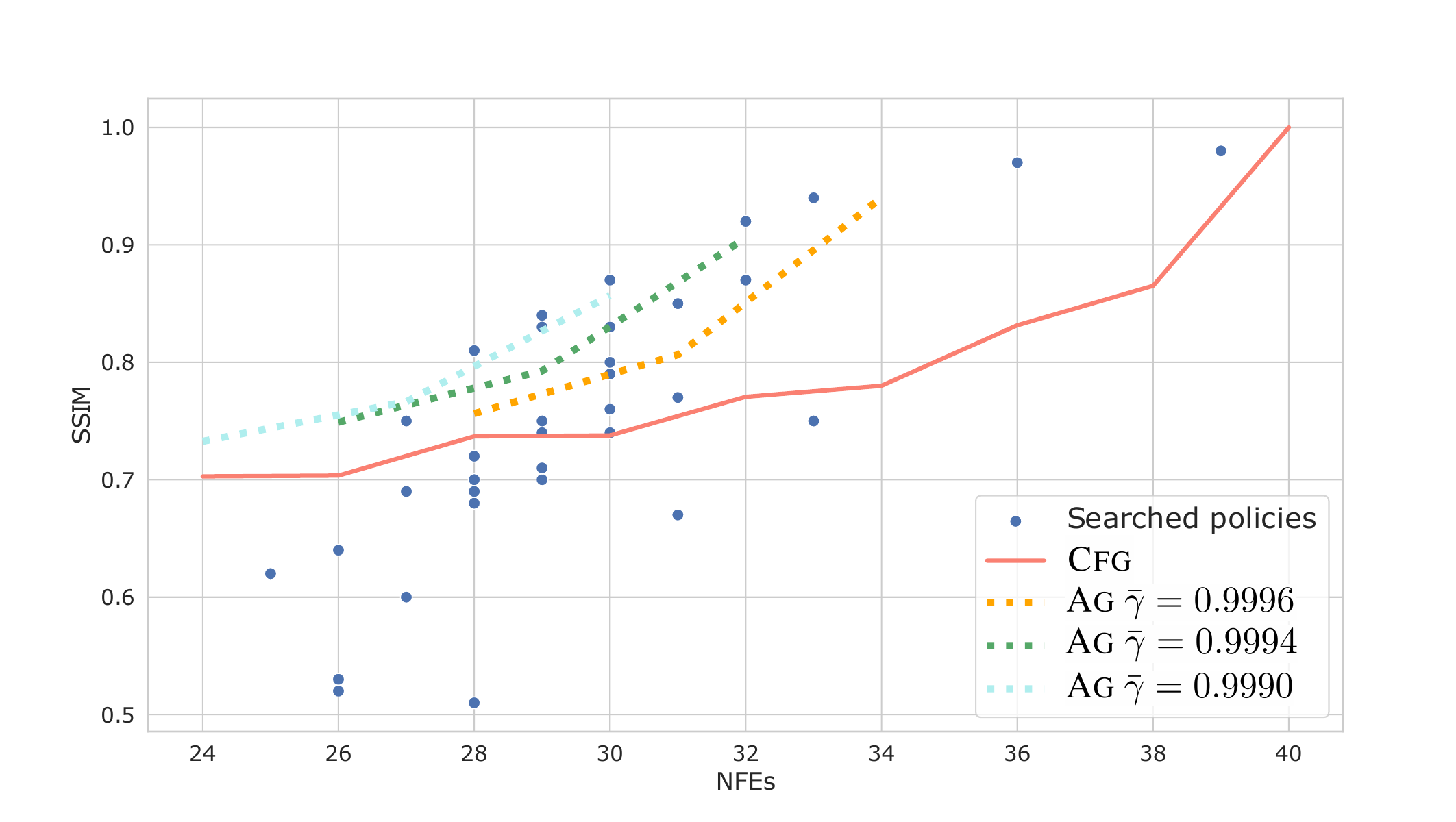}
    \vspace{-0.1cm}
    \caption{\textbf{Search results:} \footnotesize SSIM of different searched policies (dots) compared to the 20 step \cfg baseline on LDM-512. Also depicted are results of \ourmethod for different truncation threshold $\Bar{\gamma}$ (dashed lines) as well \cfg with na\"{i}ve step reduction (solid lines). The total number of steps reduces from right to left. As can be seen, \ourmethod is strictly better at replicating the baseline than a na\"{i}ve step reduction. Furthermore, it matches most individual searched policies, while being simpler and scalable.}      
    \label{fig:benchmark}
\end{figure}

\paragraph{Quantitative evaluation.}
We compare \ourmethod and \cfg w.r.t. their capacity to reconstruct a set of $1,000$ target images (computed from a baseline model with 20 \cfg steps, \ie, 40 NFEs).
We run this experiment for LDM-512 and report results in Figure~\ref{fig:benchmark} for various values of $\Bar{\gamma}$.
We find that \ourmethod can replicate the target images to higher accuracy than \cfg for the entire regime we considered (from 22 to 40 NFEs). %  how the policies found in our search can reasonably approximate the expensive \cfg policy with fewer NFEs.
% These results were computed on a comprehensive benchmark on $1,000$ samples generated with LDM-512 (see Appendix for a replication on EMU-768).
Again, as detailed in Appendix~\ref{appx:human_eval}, these findings generalize to the much larger EMU-768 model.

\paragraph{Qualitative evaluation.}
Figure~\ref{fig:figure1} depicts samples generated with \ourmethod for different $\Bar{\gamma}$ values. 
Our results suggest that up to $50\%$ of the diffusion steps can be performed without \cfg at no cost to image quality. 
Moreover, Figures~\ref{fig:teaser} and \ref{fig:figure1} showcase samples where \ourmethod outperforms the na\"{i}ve alternative of reducing the total number of diffusion steps. 

\begin{table}[t]
\centering
\footnotesize
\begin{tabular}{l|ccccc}
\toprule
\textbf{EMU-756} & \textbf{SSIM} $\uparrow$ & \textbf{Win} $\uparrow$  & \textbf{Lose} $\downarrow$ &  \textbf{NFEs} $\downarrow$ \\ \hline
\textsc{Cfg}        &  \multirow{2}{*}{$0.91\pm 0.03$} & 502      &     498   &  40            \\ %\cline{1-1}\cline{3-5}
\ourmethod $\Bar{\gamma}=0.991$ &  & $498$ & $502$ & $29.6\pm 1.3$ \\ 
\bottomrule\end{tabular}
\caption{\textbf{Evaluation results.} \footnotesize Comparison of \ourmethod ($\Bar{\gamma}=0.991$, approximately 30 NFE) and the 20 step (\ie 40 NFE) \cfg baseline. Avg. SSIM and majority voting of five trained human evaluators, each voting on $1,000$ images generated from OUI prompts. \ourmethod achieves equal visual quality despite using $25\%$ fewer NFEs.}
\label{tab:human_eval}
\end{table}

\paragraph{Human evaluation.}
We further validate \ourmethod's capacity to generate images of \cfg-level  quality % while reducing the total number of NFEs by up to $25\%$, we conducted 
via a thorough human evaluation. 
Our assessment involved five trained human annotators, who were tasked with voting for the most visually appealing instance from a pair of images.
One image was sampled from \cfg (with 40 NFEs), and the other from \ourmethod with $\Bar{\gamma}=0.991$, inducing approximately a 25\% reductionin NFEs.
We ran this test on $1,000$ prompts from the OUI dataset and report results in Table~\ref{tab:human_eval}.
Statistical analysis revealed a mean difference in votes of $-0.047$ ($\text{SD}=2.543$), indicating no significant overall preference. 
The distribution of votes was nearly even, with \ourmethod favored in $498$ cases and \cfg in 502 cases (majority voting). 
We further conducted a two-sided Wilcoxon Signed-Rank Test, yielding a $p$-value of $0.603$, with a test statistic of $244,590$, indicating \textit{no significant difference in visual appeal} between the two models ($p > 0.05$).
These findings suggest that, despite the efficiency of \ourmethod, the generated images are of comparable quality to \textsc{Cfg}, as judged by human annotators searching for visual aesthetics. 
Figure~\ref{fig:human_eval} depicts some exemplary samples of this evaluation.

\begin{figure}[t!]
    \centering
    \setlength{\tabcolsep}{2pt}

    \begin{tabular}{c c}
        % Row title A
        \multicolumn{2}{c}{\footnotesize  ``A whale breaching near a mountain'' (win)} \\
    
        % First pair of images
        \begin{subfigure}{0.48\columnwidth} 
            \includegraphics[width=\linewidth]{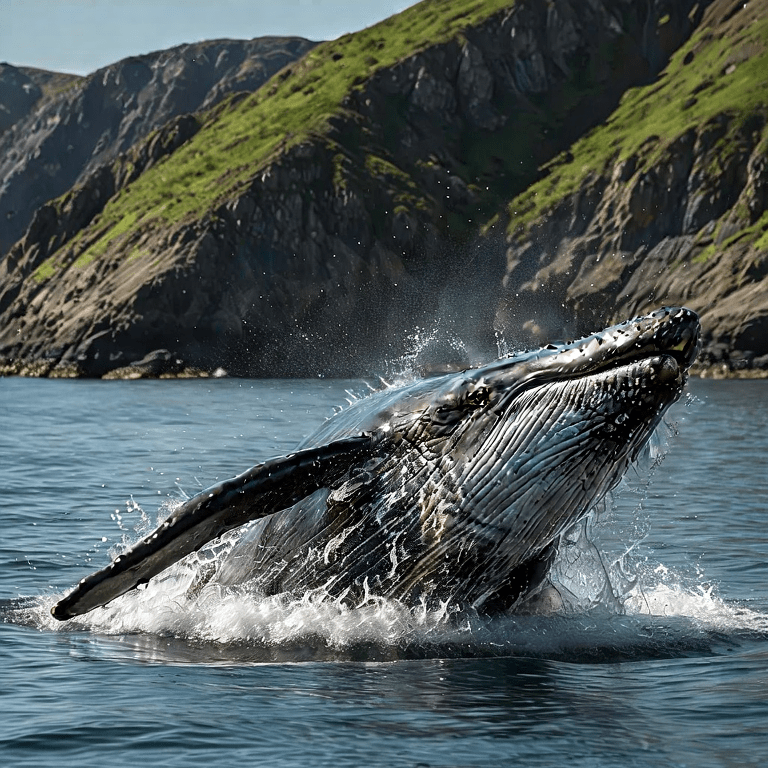}
            \caption{\textsc{Cfg} (40NFEs)}
        \end{subfigure} &
        \begin{subfigure}{0.48\columnwidth}
            \includegraphics[width=\linewidth]{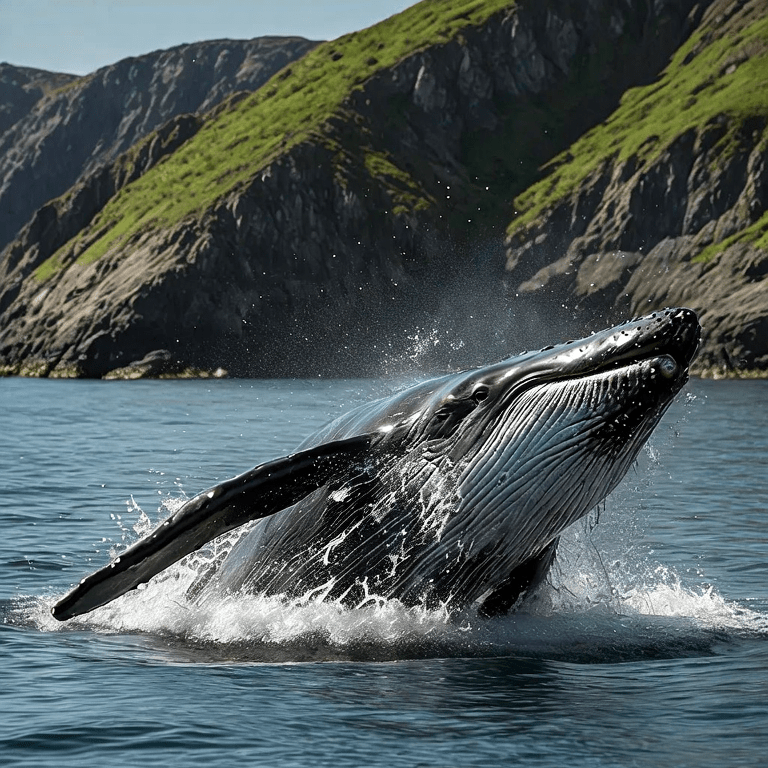}
            \caption{\ourmethod, $\Bar{\gamma}=0.991$ (30NFEs)}
        \end{subfigure} \\
        \multicolumn{2}{c}{\footnotesize ``A red laptop with earbuds sitting on a table'' (lose)} \\

        % Third pair of images
        \begin{subfigure}{0.48\columnwidth}
            \includegraphics[width=\linewidth]{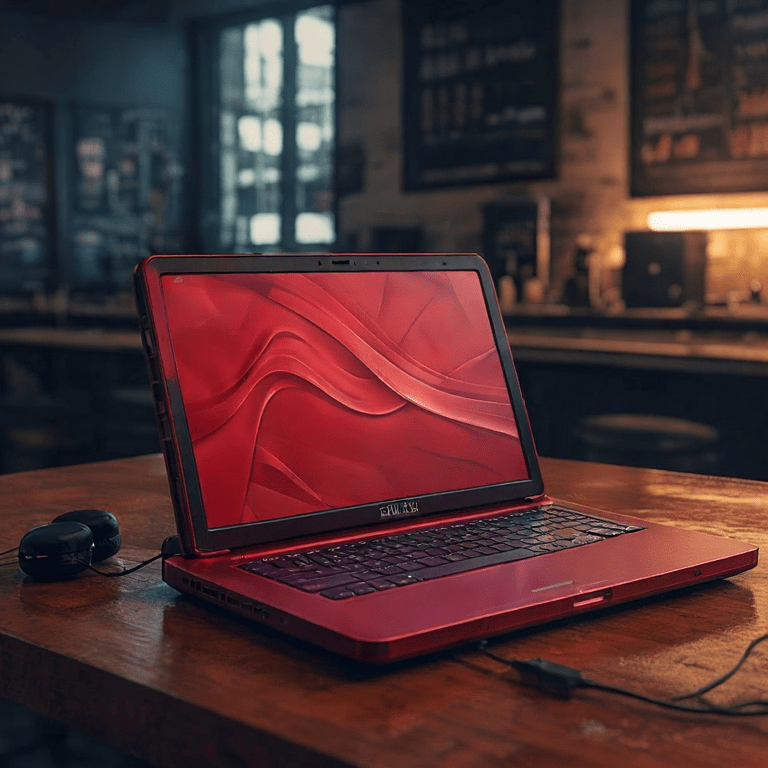}
            \caption{\textsc{Cfg} (40NFEs)}
        \end{subfigure} &
        \begin{subfigure}{0.48\columnwidth}
            \includegraphics[width=\linewidth]{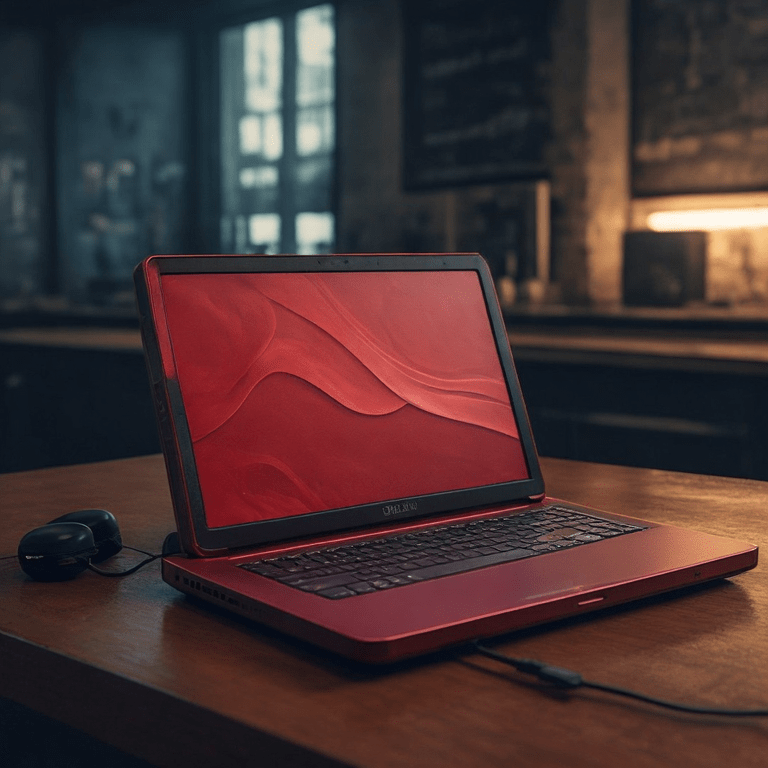}
            \caption{\ourmethod, $\Bar{\gamma}=0.991$ (29NFEs)}
        \end{subfigure}

    \end{tabular}
    \vspace{-0.5cm}
    \caption{\textbf{Human evaluation examples.} \footnotesize An exemplary sample for win (top) and lose (bottom) of \ourmethod with $\Bar{\gamma}=0.991$ \textit{vs.} \cfg. The baseline \cfg tends to produce higher frequencies, which can be for better (bottom) or worse (top). We report further examples in Appendix~\ref{appx:human_eval}.}
\label{fig:human_eval}
\end{figure}

\subsection{Replacing NFEs with Affine Transformations}\label{sec:ols}
In the previous sections, we found that, in the latter stages of denoising, \cfg updates can be replaced with conditional steps. 
Yet, for the first half of the denoising steps, guidance is of particular importance. 
Indeed, as shown in the first column of Fig.~\ref{fig:ols}, reducing the number of guidance steps to as few as five (followed by $15$ conditional steps) significantly degrades image quality. 
At the same time, the smooth alignment of conditional and unconditional steps over time and the high concentration around the mean of the cosine similarities depicted in Fig.~\ref{fig:cos} suggest a high regularity in diffusion paths. 
Intrigued by this observation, we probe for linear patterns in the diffusion path.
Indeed, we find that unconditional network evaluations $\epsilon(\mathbf{x}_t,\emptyset)$ can often be estimated with high accuracy via \textit{affine} transformations of network evaluations at previous iterations. To compute the parameters of these affine transformations, we generate a small dataset of $200$ images from EMU-768 and store the intermediate iterations. Subsequently, we model a given unconditional step at any $t<T$ as a linear combination of the previous iterations in the diffusion chain as
\setlength{\abovedisplayskip}{3pt}
\setlength{\belowdisplayskip}{3pt}
\begin{equation}\label{eq:ols}
    \hat{\epsilon}(\mathbf{x}_t,\emptyset) = \sum_{i=T}^{t}\beta_i^\mathbf{c}\epsilon_{\theta}(\mathbf{x}_i,\mathbf{c}) + \sum_{i=T}^{t+1}\beta_i^{\emptyset}\epsilon_{\theta}(\mathbf{x}_i,\emptyset),
\end{equation}
where $\beta_i^\mathbf{c}$ and $\beta_i^{\emptyset}$ are \textit{scalars}. We learn these Linear Regression (LR) coefficients for each step by solving a simple Ordinary Least Squares problem on the set of 200 trajectories. 
Together with the time required for generating the dataset, we obtain LR coefficients for all steps in under 20 minutes. 
During sampling, computing $\hat{\epsilon}(\mathbf{x}_t,\emptyset)$ is essentially for free.

\begin{figure}[t!]
\setlength{\tabcolsep}{0.05em} % width separation

    \centering
    \begin{tabular}{c c c}
    \multicolumn{3}{c}{\footnotesize ``A delicious chocolate cake with fruits''} \\ % Add the title for the first row here
    \begin{subfigure}{0.32\columnwidth} 
        \includegraphics[width=\linewidth, trim=1 1 1 1, clip]{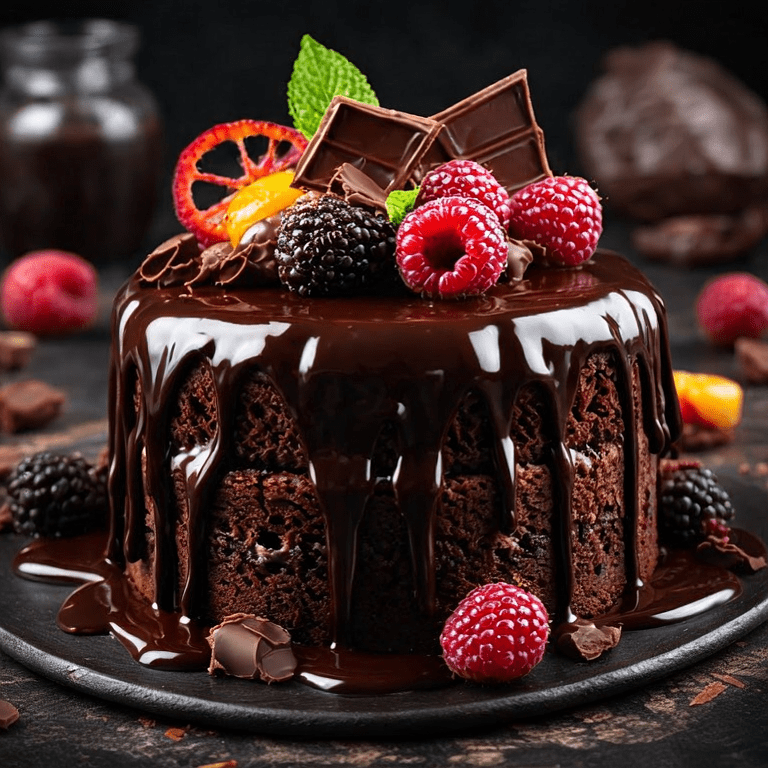}
        \caption{ \cfg \\ (40NFEs)}
    \end{subfigure}    &
    \begin{subfigure}{0.32\columnwidth}
        \includegraphics[width=\linewidth, trim=1 1 1 1, clip]{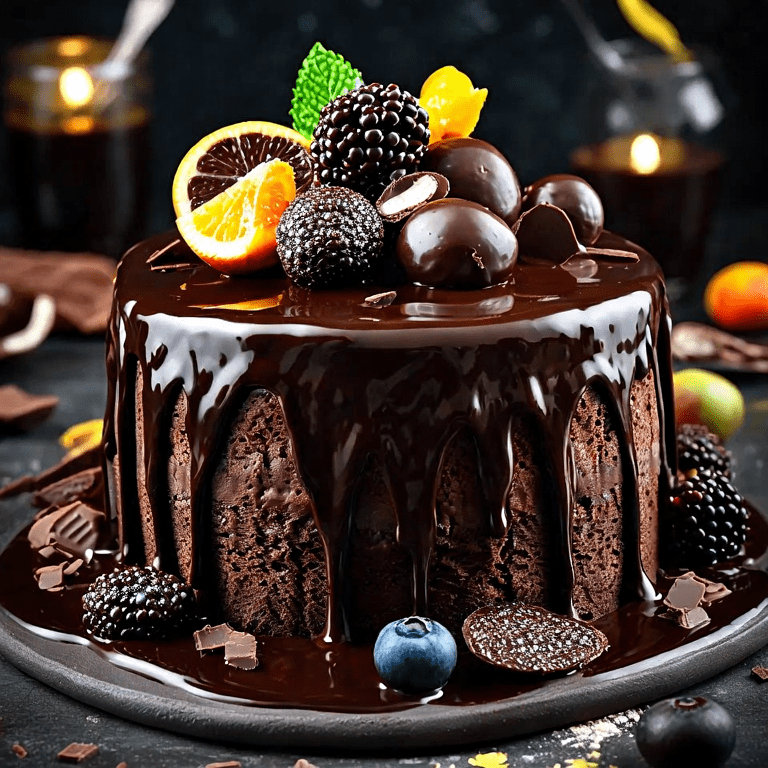}
        \caption{\centering -``raspberry'' \cfg  (40NFEs)}
    \end{subfigure} &
     \begin{subfigure}{0.32\columnwidth}
        \includegraphics[width=\linewidth, trim=1 1 1 1, clip]{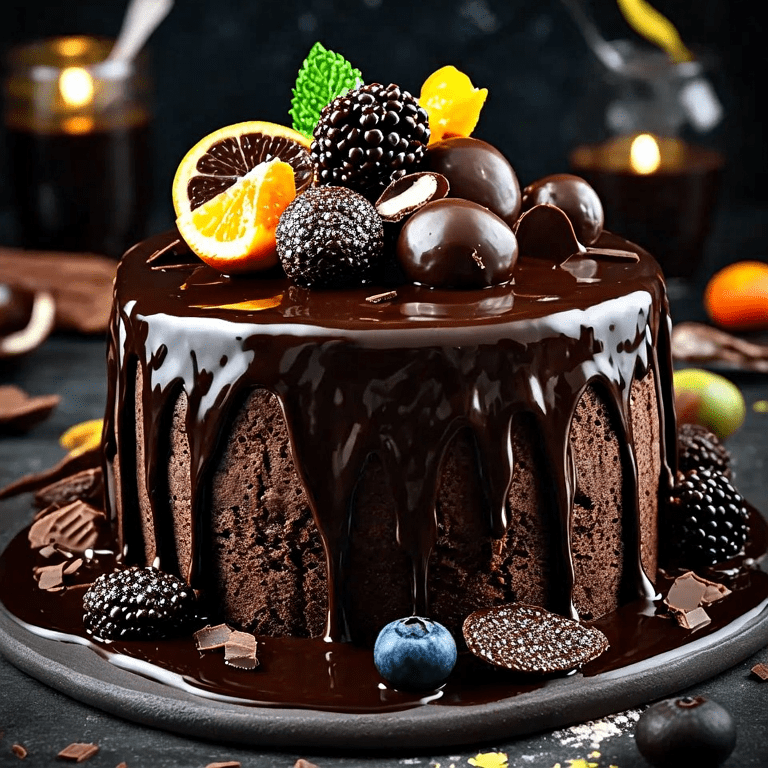}
        \caption{\centering -``raspberry'' \ourmethod  (30NFEs)}
    \end{subfigure} 
    \\ 
    \end{tabular}
    \vspace{-0.4cm}
    \caption{\textbf{Negative prompts.} \footnotesize As can be seen, \ourmethod produces similar results to \cfg when using non-empty negative prompts, again highlighting the importance of only the first $\tfrac{T}{2}$ diffusion steps for semantic structure. More such examples can be found in Figure~\ref{fig:negative_apx} in the Appendix.}
    \label{fig:negative}
\end{figure}

Perhaps surprisingly, we find that this estimator displays remarkable capacity to predict unconditional steps. Of course, for any unconditional score replaced by an LR predictor, the following denoising step will no longer have ground truth past information and errors accumulate auto-regressively. 
Yet, interleaving \cfg steps with ``approximated'' \cfg steps (where the $\epsilon_\theta(\mathbf{x}_t,\emptyset)$ is replaced by its linear estimator $\hat{\epsilon}(\mathbf{x}_t,\emptyset)$), reduces the rate of error accumulation. 
We term this strategy \ourlinearmethod. 
When $T=20$, \ourlinearmethod performs ten steps, alternating between \cfg (2 NFEs) and LR-based \cfg (1 NFE), followed by ten LR-based \textsc{Cfg} steps. More details can be found in App.~\ref{appx:ols}.

As depicted in Fig.~\ref{fig:ols}, \ourlinearmethod drastically improves image quality over \ourmethod with very low $\Bar{\gamma}$. 
Furthermore, it shows that the LR successfully recognizes patterns along the path since it compares favorably to the na\"{i}ve alternative of simply alternating between \cfg and conditional steps for the first half (followed by $\tfrac{T}{2}$ conditional steps). Finally, as reported in Figures~\ref{fig:negative} and~\ref{fig:negative_apx}, \ourlinearmethod can even handle negative prompts to a certain extent.

\begin{figure}[t!]
\setlength{\tabcolsep}{0.05em} % width separation
    \centering
    \begin{tabular}{c c c}
    \multicolumn{3}{c}{\footnotesize ``A happy cow in the Swiss alps''} \\ % Add the title for the first row here
    \begin{subfigure}{0.33\columnwidth} 
        \includegraphics[width=\linewidth, trim=1 1 1 1, clip]{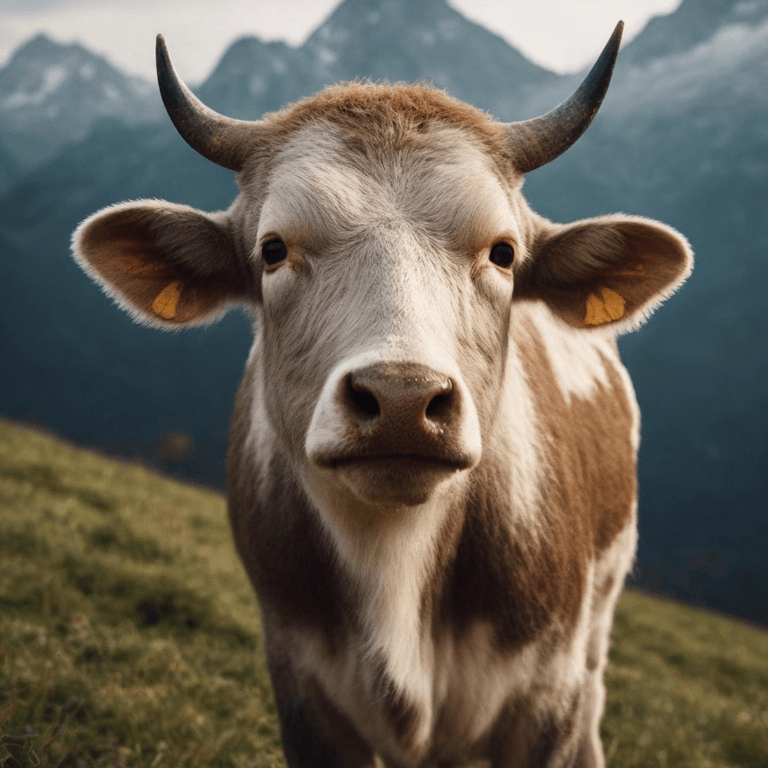}
    \end{subfigure}    &
    \begin{subfigure}{0.33\columnwidth}
        \includegraphics[width=\linewidth, trim=1 1 1 1, clip]{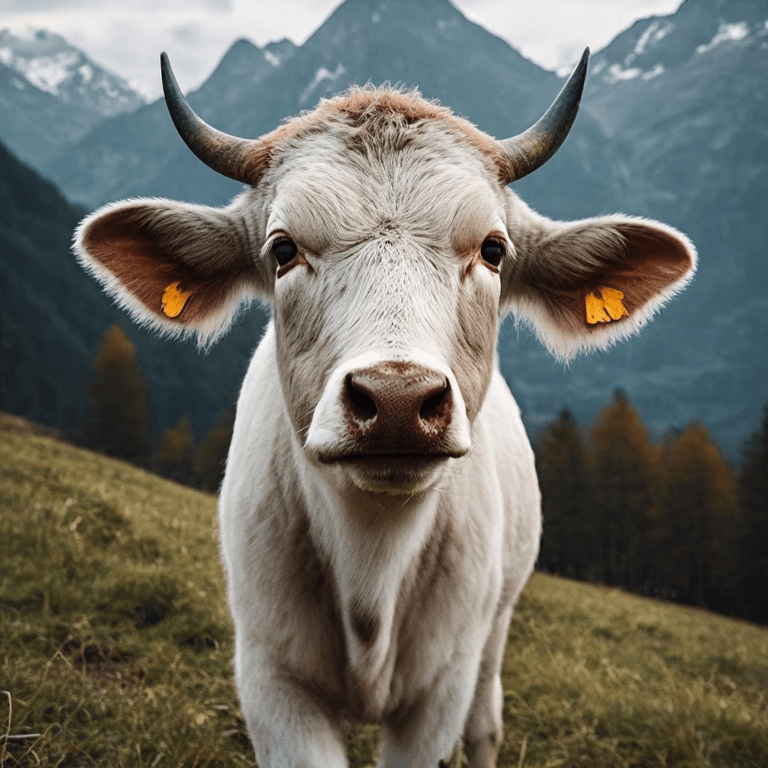}
    \end{subfigure} &
     \begin{subfigure}{0.33\columnwidth}
        \includegraphics[width=\linewidth, trim=1 1 1 1, clip]{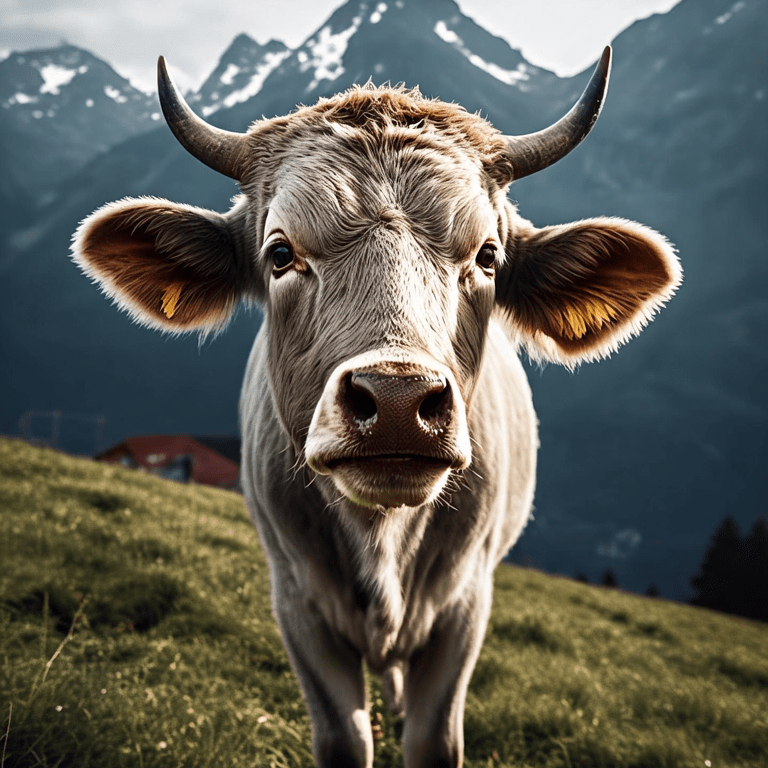}
    \end{subfigure}   \\
    \multicolumn{3}{c}{\footnotesize ``A traditional tea house in a garden with cherry blossom trees''} \\ % Add the title for the second row here
    \begin{subfigure}{0.33\columnwidth} 
        \includegraphics[width=\linewidth, trim=1 1 1 1, clip]{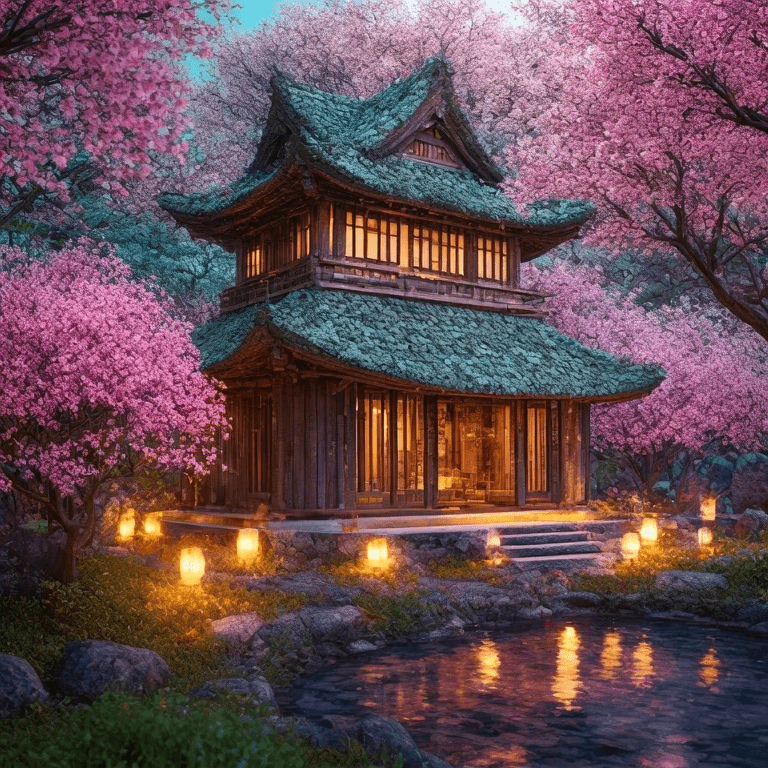}
        \caption{\centering \ourmethod $\Bar{\gamma}=0.975$ (25NFEs)}
    \end{subfigure}    &
    \begin{subfigure}{0.33\columnwidth}
        \includegraphics[width=\linewidth, trim=1 1 1 1, clip]{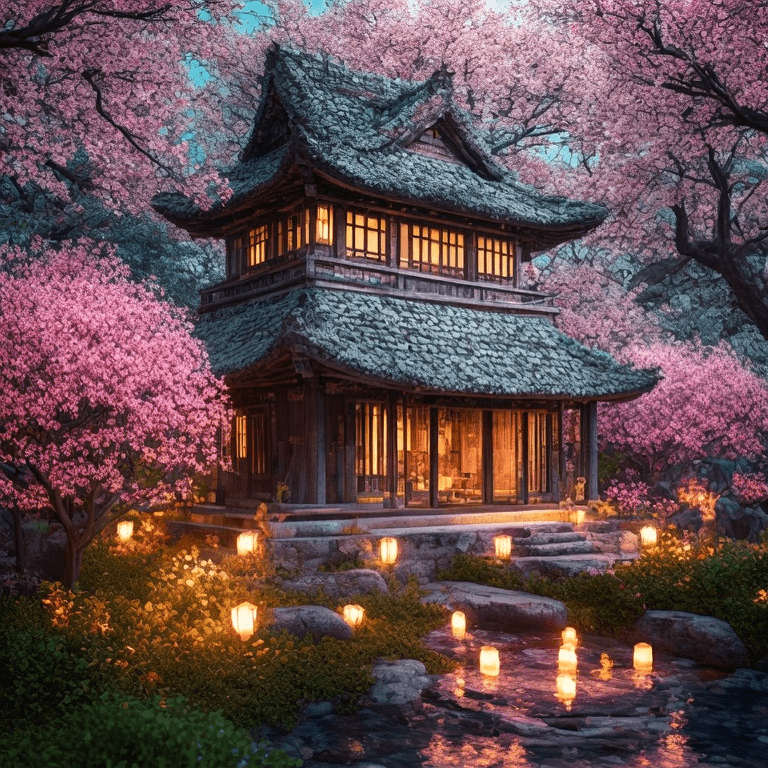}
        \caption{\centering na\"{i}ve interleaving \cfg (25NFEs)}
    \end{subfigure} &
     \begin{subfigure}{0.33\columnwidth}
        \includegraphics[width=\linewidth, trim=1 1 1 1, clip]{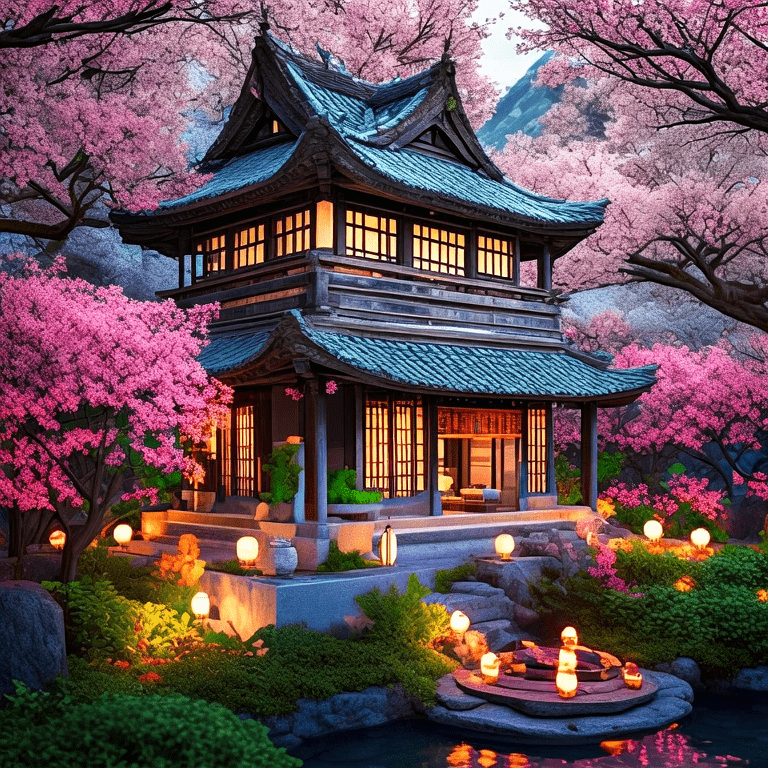}
        \caption{\centering \textcolor{white}{...}  \ourlinearmethod \textcolor{white}{...} (25NFEs)}
    \end{subfigure}   
    \end{tabular}
    \vspace{-0.4cm}
    \caption{\textbf{Replacing \cfg in the first half of diffusion steps.} \footnotesize Three different approaches to reduce the number of NFEs in the first $50\%$ of diffusion steps. As can be seen, \ourlinearmethod samples show increased sharpness, dynamic lightning with higher contrast, and more vivid colors.}
    \label{fig:ols}
\end{figure}

\section{Conclusions}
In this paper, we have leveraged the gradient-based NAS framework to bring about principled insights into the denoising process of conditional diffusion models. In particular, we found that Classifier-Free Guidance involves computational redundancies of different sorts in both the first and second parts of the diffusion process. Leveraging these insights, we first proposed Adaptive Guidance, a very general and efficient plug-and-play variant of \textsc{Cfg} that is able to closely replicate a baseline model while reducing the number of 
% Neural Function Evaluations 
NFEs needed for guidance by up to $50\%$. Compared to Guidance Distillation, \ourmethod is training-free, extremely easy to implement and it offers considerable flexibility, for example when it comes to negative prompts or image editing.

Second, we proposed an even faster variant of \cfg, termed \ourlinearmethod, that increases the guidance NFE savings of \textsc{Ag} to $75\%$ by replacing entire network evaluations with surprisingly simple linear transformations of past information. However, these extra runtime savings come at the price of \ourlinearmethod no longer replicating the baseline one-to-one, which entails the need for extensive evaluations. As such, \ourlinearmethod is to be considered more as a proof of concept as well as an interesting starting point for future research on finding ways to effectively leverage smoothness in and regularity across diffusion paths for efficient inference.

{
    \small
    \bibliographystyle{ieeenat_fullname}
    \bibliography{main}
}
\appendix

\clearpage
\setcounter{page}{1}
\maketitlesupplementary
% \section{Rationale}
% \label{sec:rationale}
% % 
% Having the supplementary compiled together with the main paper means that:
% % 
% \begin{itemize}
% \item The supplementary can back-reference sections of the main paper, for example, we can refer to \cref{sec:intro};
% \item The main paper can forward reference sub-sections within the supplementary explicitly (e.g. referring to a particular experiment); 
% \item When submitted to arXiv, the supplementary will already included at the end of the paper.
% \end{itemize}
% % 
% To split the supplementary pages from the main paper, you can use \href{https://support.apple.com/en-ca/guide/preview/prvw11793/mac#:~:text=Delete%20a%20page%20from%20a,or%20choose%20Edit%20%3E%20Delete).}{Preview (on macOS)}, \href{https://www.adobe.com/acrobat/how-to/delete-pages-from-pdf.html#:~:text=Choose%20%E2%80%9CTools%E2%80%9D%20%3E%20%E2%80%9COrganize,or%20pages%20from%20the%20file.}{Adobe Acrobat} (on all OSs), as well as \href{https://superuser.com/questions/517986/is-it-possible-to-delete-some-pages-of-a-pdf-document}{command line tools}.

\section{Evaluations on EMU-768} \label{appx:human_eval}
\begin{figure}[t]
    \centering
        \includegraphics[width=\linewidth,trim=15 5 25 25, clip]{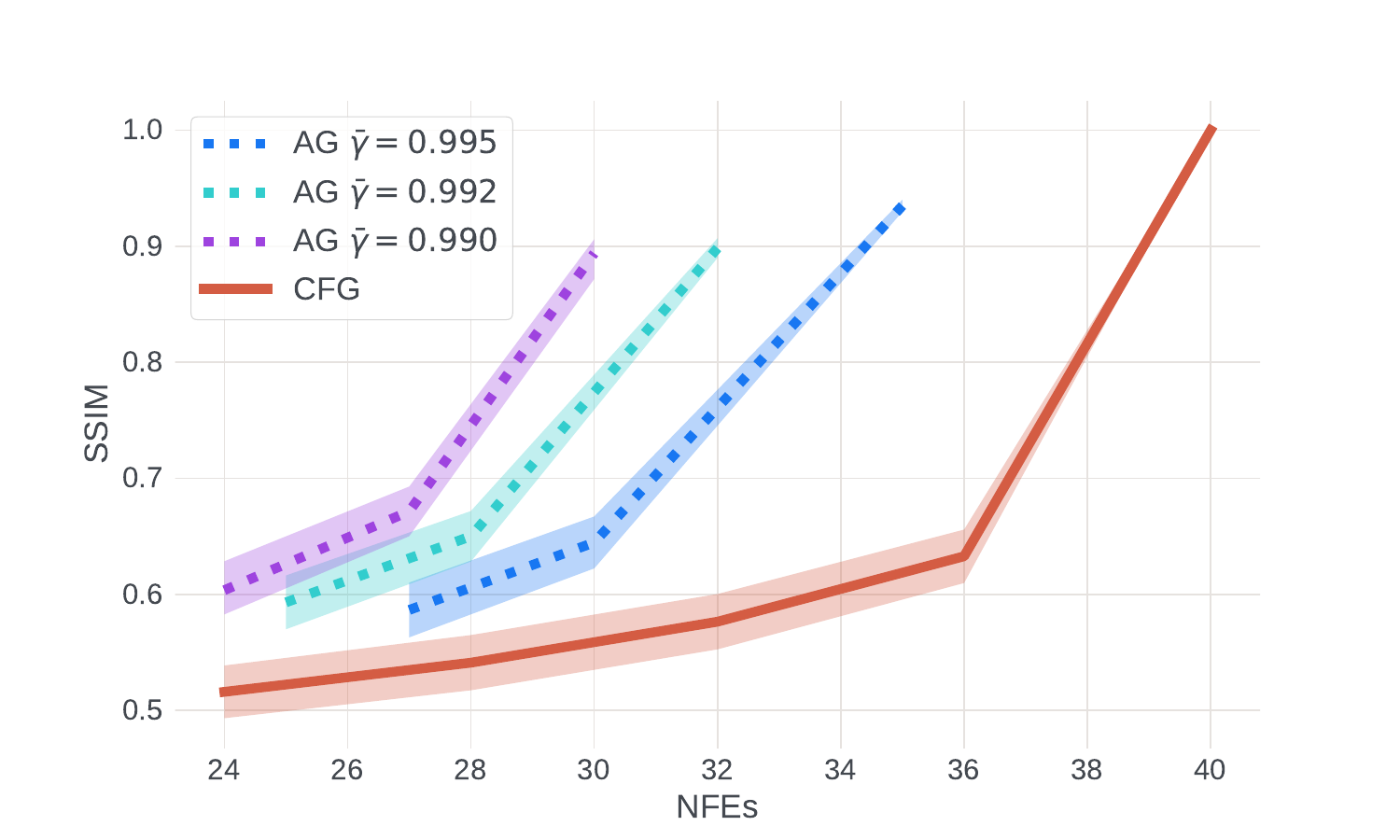}
    \vspace{-0.1cm}
    \caption{\textbf{\ourmethod vs \cfg:} SSIM (mean and $95\%$ CI) compared to the 20 step \cfg baseline on EMU-768. Depicted are results of \ourmethod for different truncation threshold $\Bar{\gamma}$ (dashed lines) as well \cfg with na\"{i}ve step reduction (solid lines). The total number of steps reduces from right to left. As can be seen, \ourmethod is strictly better at replicating the baseline than a na\"{i}ve step reduction. Similar results for LDM-512 can be found in Fig. \ref{fig:benchmark}}      
    \label{fig:benchmark_apx}
\end{figure}

As stated in the main paper, our policy search was performed on the LDM-512 model. Importantly, we find that the resulting adaptive guidance policies generalize to the much bigger and more powerful EMU-768 model. For example \ref{fig:benchmark_apx}, similar to Fig. \ref{fig:benchmark} on LDM, shows that \ourmethod scales more favorably than \cfg for different numbers of NFEs on EMU. 

For our human evaluation results, we generated images using 20 \cfg steps as well as 20 \ourmethod steps with $\Bar{\gamma}=0.991$, which gave rise to an average of 29.6 NFEs (that is, the average sample was generated with around 10 guided steps, followed by 10 unguided (conditional) steps). We used the same seed sequence for both models on a subset of 1000 prompts from OUI.

\begin{figure}[t]
    \centering
        \includegraphics[width=\linewidth,trim=10 5 25 25, clip]{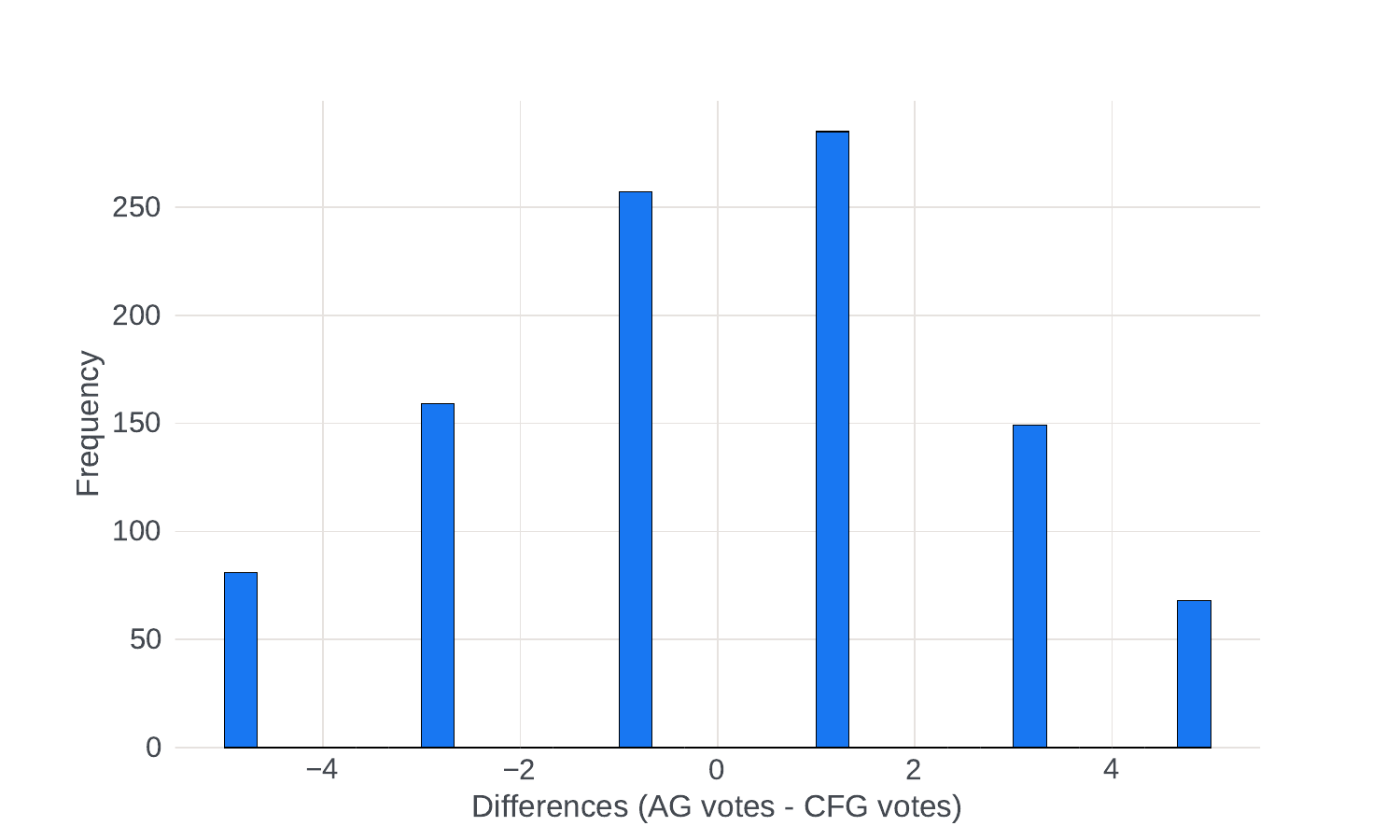}
    \vspace{-0.1cm}
    \caption{\textbf{Voting distribution.} Distribution of voting difference of five annotators for \ourmethod vs. \cfg for 1000 samples. As can be seen, the distribution is very symmetric around zero. Hence, paired difference tests can find no significant difference in the model performance.}      
    \label{fig:voting_dist}
\end{figure}

After generation, for each prompt, the images of both models were shown side-by-side to a random subset of 5 out of a pool of 42 trained human evaluators. The order of the images was also random. Annotators had to vote for higher visual appeal. There was no tie option to incentivize active engagement. The vote distribution was symmetric around zero (see Fig. \ref{fig:voting_dist}. Hence, no significant difference in the model performance can be found by paired difference tests.

\newcommand{\negativesize}{0.24}

\begin{figure*}[t!]
\setlength{\tabcolsep}{0.05em} % width separation

    \centering
    \begin{tabular}{c c c c}
   
    \multicolumn{4}{c}{\footnotesize ``A Tuscany villa'' (success)} \\ % Add the title for the first row here
    \begin{subfigure}{\negativesize\textwidth} 
        \includegraphics[width=\linewidth, trim=1 1 1 1, clip]{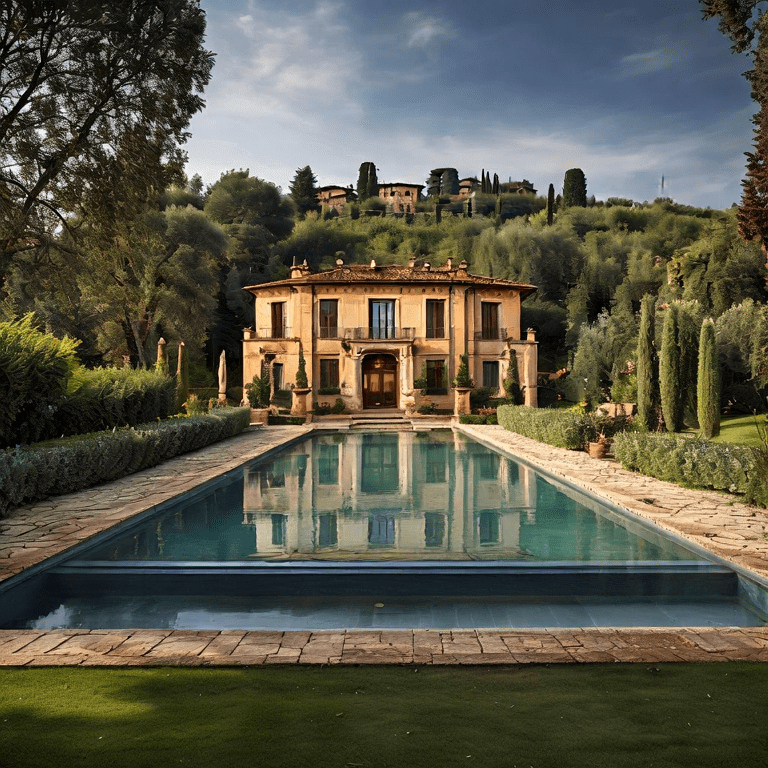}
        \caption{\centering \cfg (40NFEs) }
    \end{subfigure}    &
    \begin{subfigure}{\negativesize\textwidth}
        \includegraphics[width=\linewidth, trim=1 1 1 1, clip]{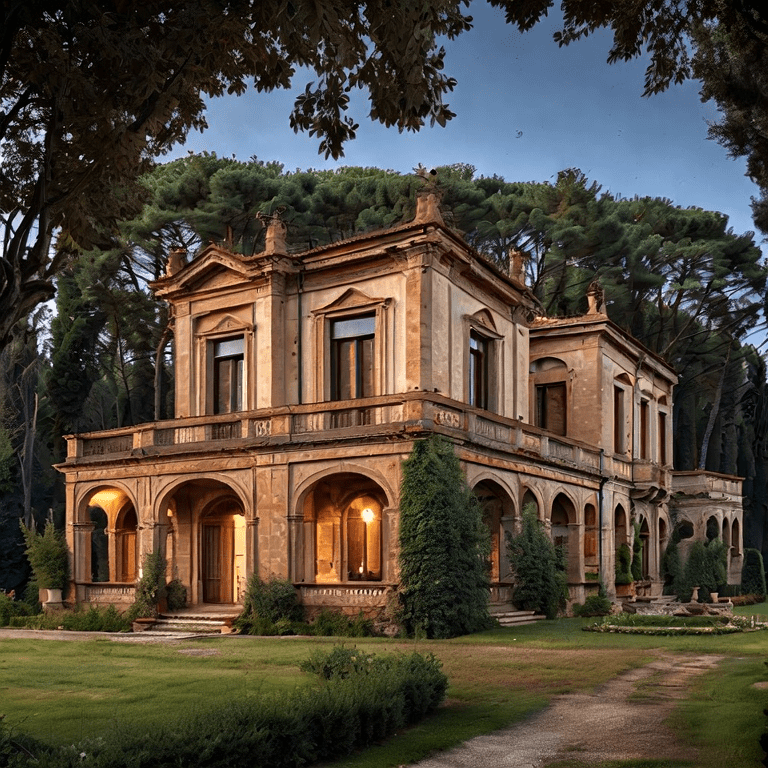}
        \caption{\centering  -``swimmingpool'' \cfg (40NFEs)}
    \end{subfigure}  &
     \begin{subfigure}{\negativesize\textwidth}
        \includegraphics[width=\linewidth, trim=1 1 1 1, clip]{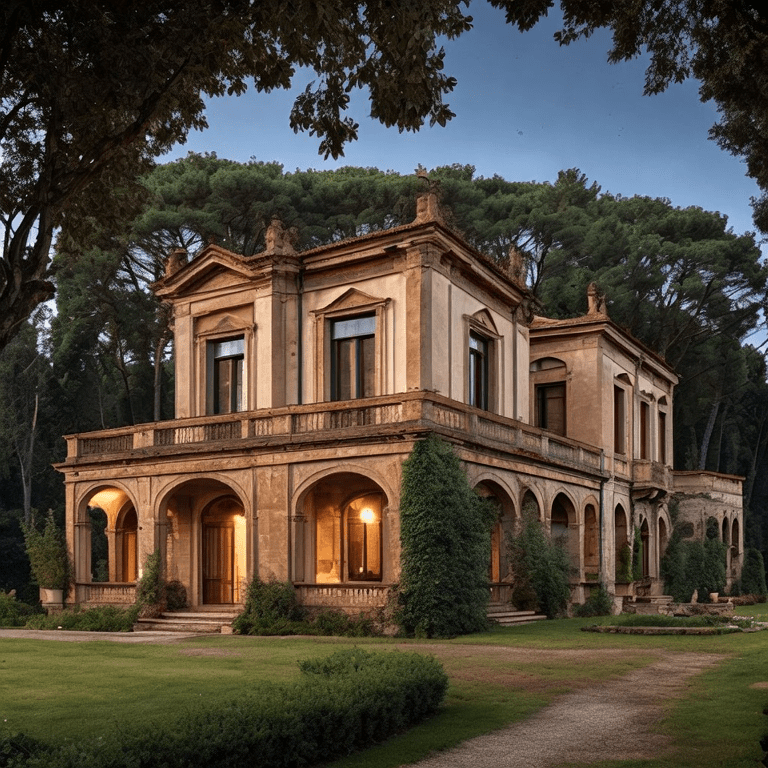}
        \caption{\centering  -``swimmingpool''  \ourmethod (30NFEs)   }
    \end{subfigure} &
     \begin{subfigure}{\negativesize\textwidth}
        \includegraphics[width=\linewidth, trim=1 1 1 1, clip]{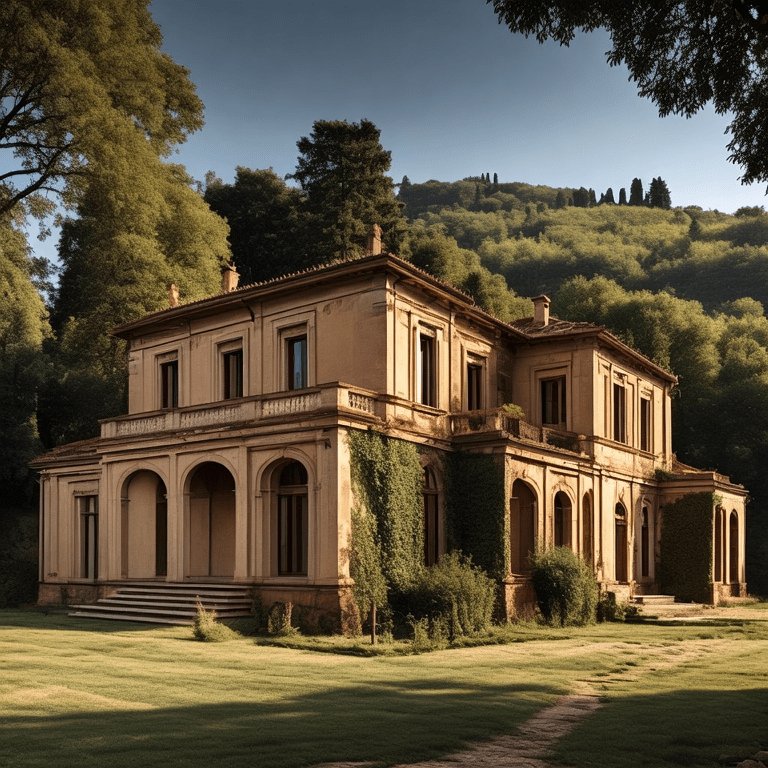}
        \caption{\centering -``sw..pool'' \ourlinearmethod (25NFEs)   }
    \end{subfigure} 
    \\ 
     \multicolumn{4}{c}{\footnotesize ``A healthy bowl of salad'' (success)} \\ % Add the title for the first row here
    \begin{subfigure}{\negativesize\textwidth} 
        \includegraphics[width=\linewidth, trim=1 1 1 1, clip]{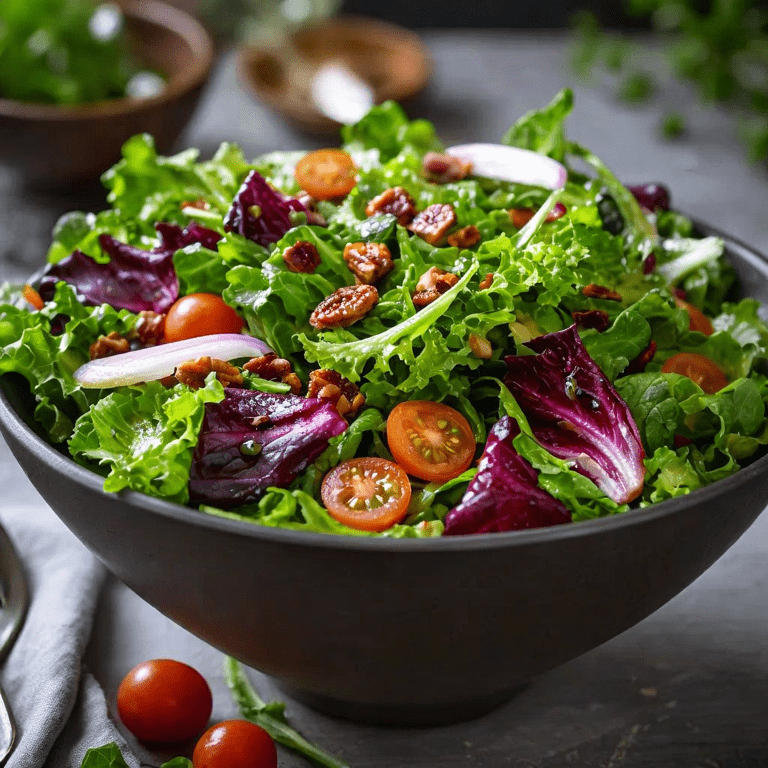}
        \caption{\centering \cfg (40NFEs) }
    \end{subfigure}    &
    \begin{subfigure}{\negativesize\textwidth}
        \includegraphics[width=\linewidth, trim=1 1 1 1, clip]{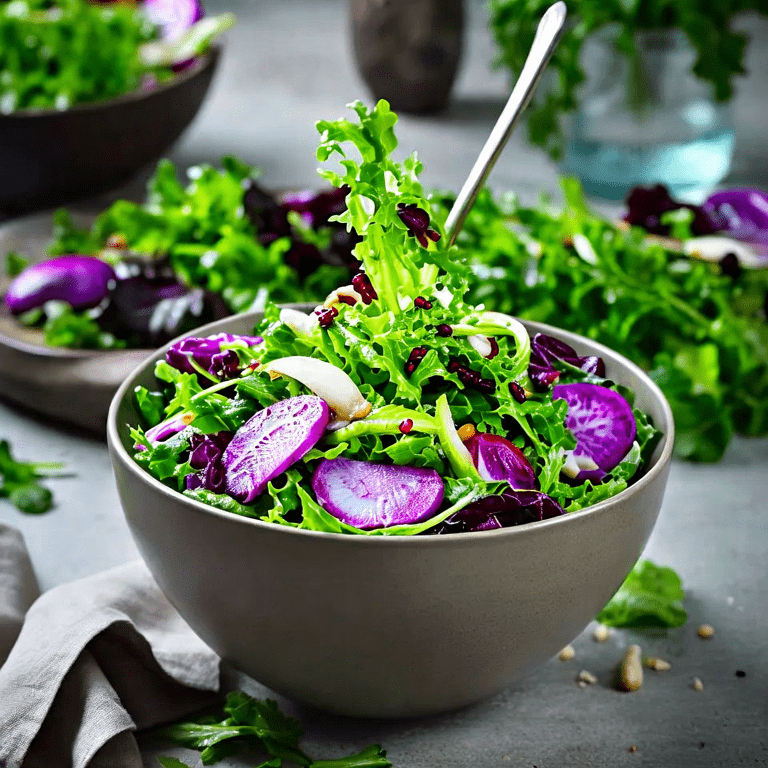}
        \caption{\centering  -``tomatoes'' \cfg (40NFEs)}
    \end{subfigure}  &
     \begin{subfigure}{\negativesize\textwidth}
        \includegraphics[width=\linewidth, trim=1 1 1 1, clip]{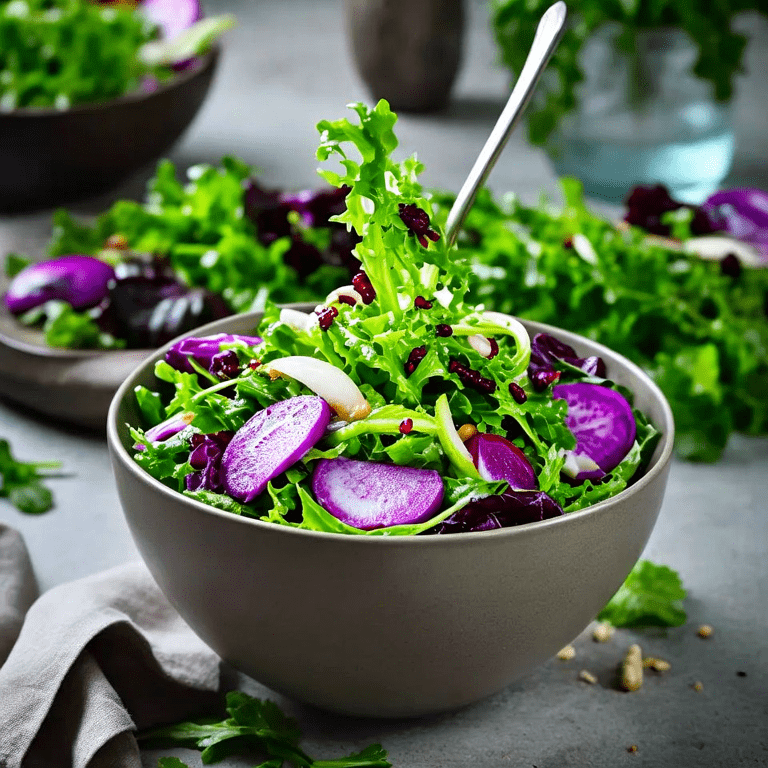}
        \caption{\centering  -``tomatoes''  \ourmethod (30NFEs)   }
    \end{subfigure} &
     \begin{subfigure}{\negativesize\textwidth}
        \includegraphics[width=\linewidth, trim=1 1 1 1, clip]{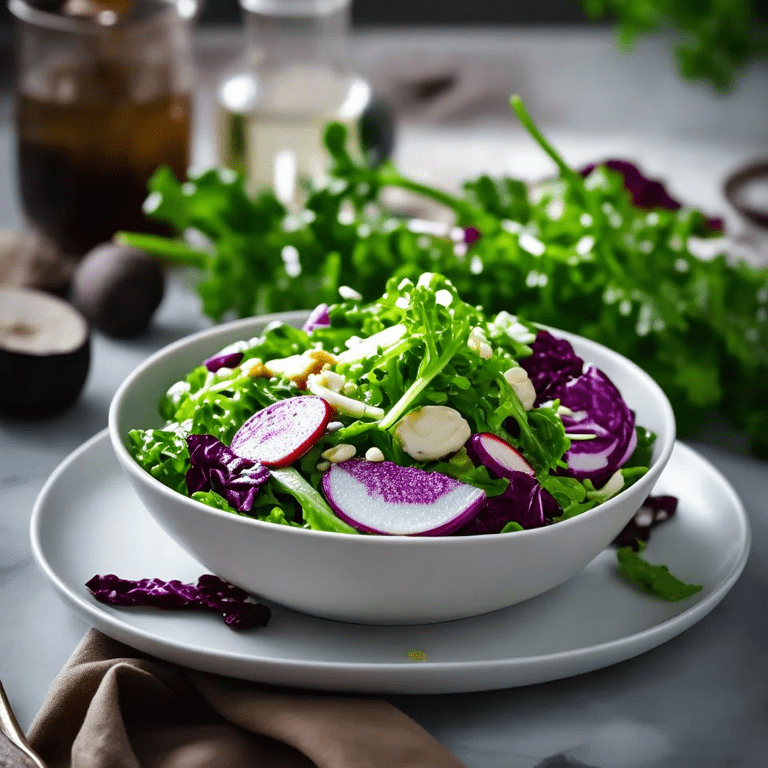}
        \caption{\centering -``tomatoes'' \ourlinearmethod (25NFEs)   }
    \end{subfigure} 
    \\ 
    \multicolumn{4}{c}{\footnotesize ``An Italian pizza'' (failure)} \\ % Add the title for the first row here
    \begin{subfigure}{\negativesize\textwidth} 
        \includegraphics[width=\linewidth, trim=1 1 1 1, clip]{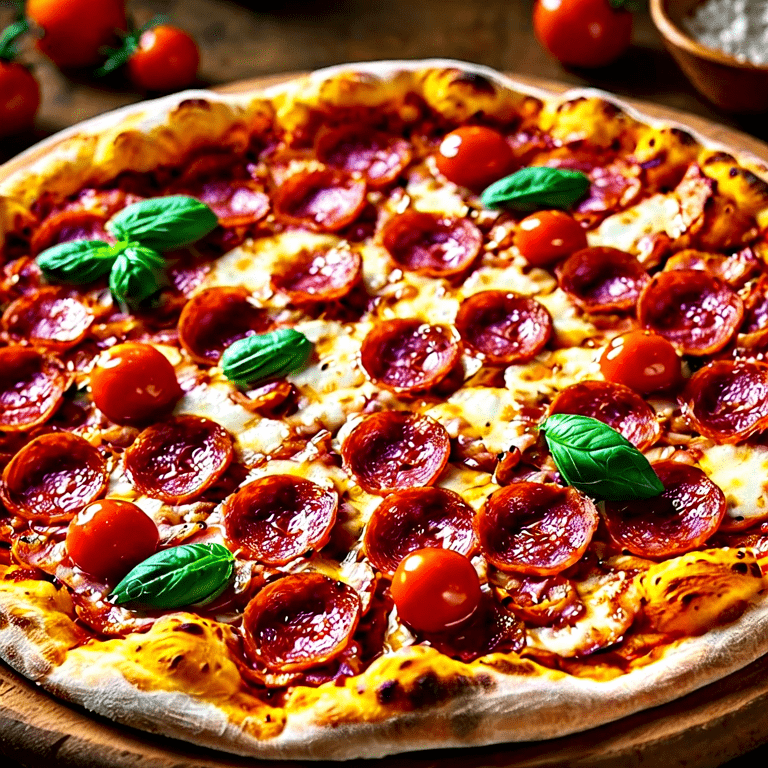}
        \caption{\centering \cfg (40NFEs) }
    \end{subfigure}    &
    \begin{subfigure}{\negativesize\textwidth}
        \includegraphics[width=\linewidth, trim=1 1 1 1, clip]{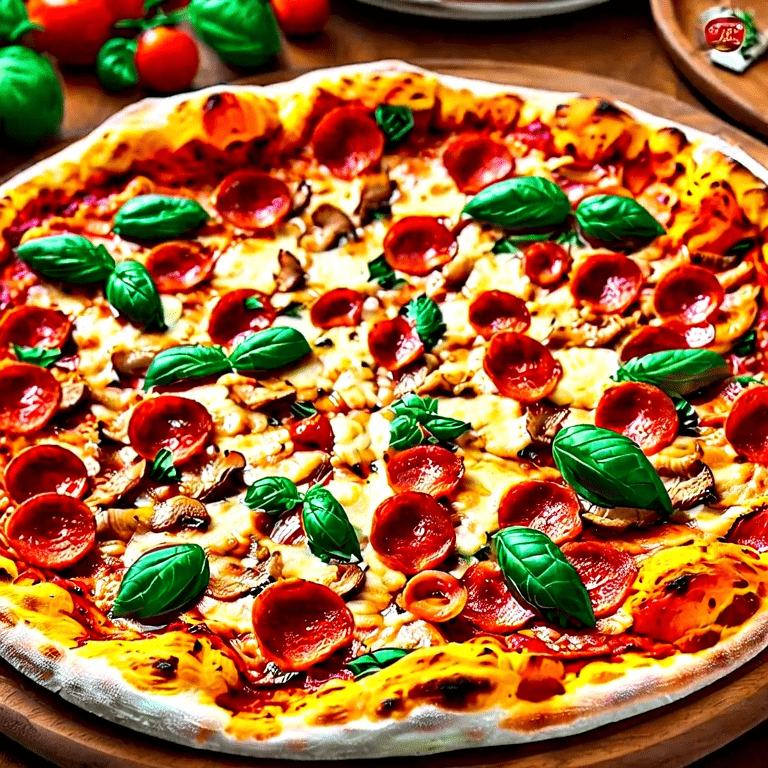}
        \caption{\centering  -``raspberry'' \cfg (40NFEs)}
    \end{subfigure} &
     \begin{subfigure}{\negativesize\textwidth}
        \includegraphics[width=\linewidth, trim=1 1 1 1, clip]{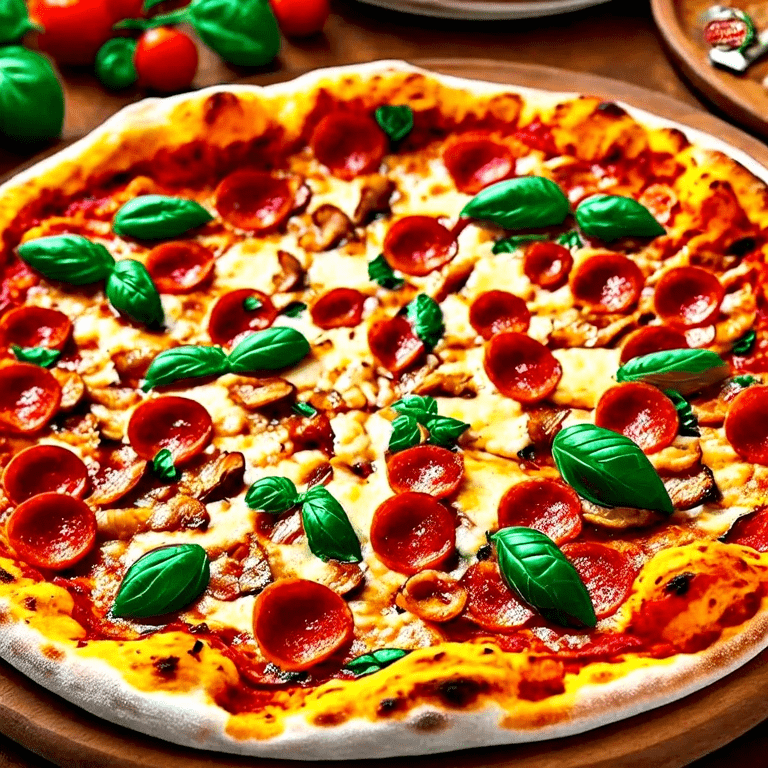}
        \caption{\centering  -``raspberry''  \ourmethod (30NFEs)   }
    \end{subfigure} &
     \begin{subfigure}{\negativesize\textwidth}
        \includegraphics[width=\linewidth, trim=1 1 1 1, clip]{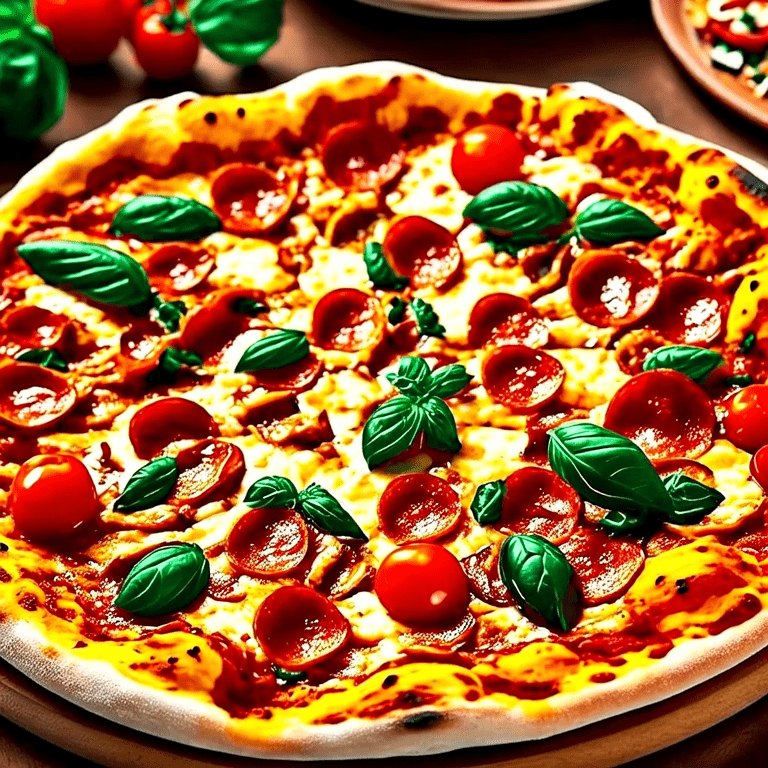}
        \caption{\centering -``raspberry'' \ourlinearmethod (25NFEs)   }
    \end{subfigure} 
    \end{tabular}
    \vspace{-0.4cm}
    \caption{\textbf{More negative prompts.} \footnotesize More examples of using negative prompts with adaptive and linear adaptive guidance. The top and middle rows show successful examples. The bottom row shows a failure case. Importantly, standard \cfg also fails in the latter case.}
    \label{fig:negative_apx}
\end{figure*}

\begin{figure}[t!]
    \centering
    \setlength{\tabcolsep}{2pt}

    \begin{tabular}{c c}
        % Row title A
        \multicolumn{2}{c}{\footnotesize  ``Two violins standing up with their bows on the ground'' (win)} \\
    
        % First pair of images
        \begin{subfigure}{0.48\columnwidth} 
            \includegraphics[width=\linewidth]{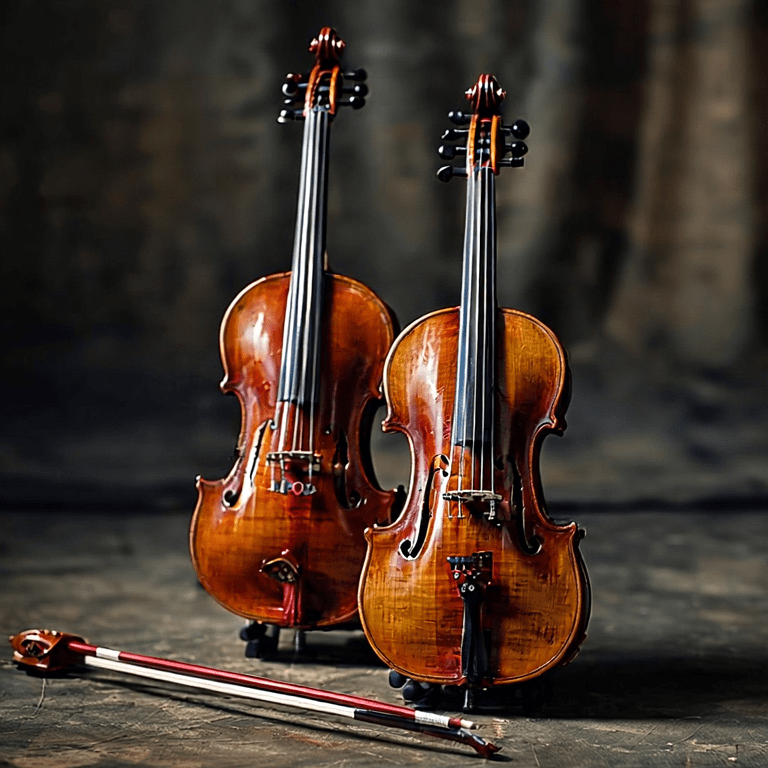}
            \caption{\textsc{Cfg} (40NFEs)}
        \end{subfigure} &
        \begin{subfigure}{0.48\columnwidth}
            \includegraphics[width=\linewidth]{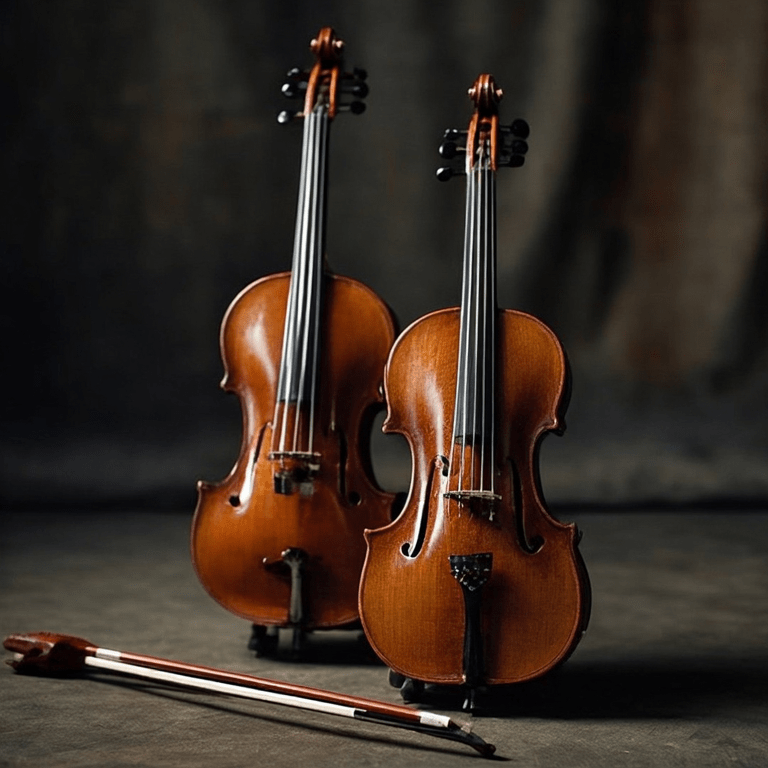}
            \caption{\ourmethod, $\Bar{\gamma}=0.991$ (30NFEs)}
        \end{subfigure} \\
        \multicolumn{2}{c}{\footnotesize  ``A toucan close up, midnight, lake, dark, moon light'' (win)} \\
    
        % First pair of images
        \begin{subfigure}{0.48\columnwidth} 
            \includegraphics[width=\linewidth]{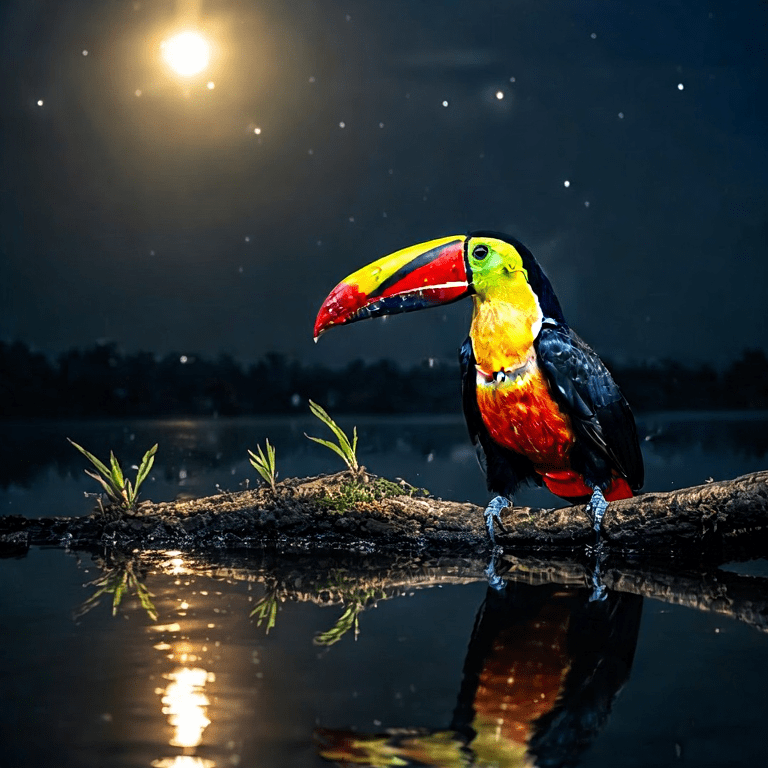}
            \caption{\textsc{Cfg} (40NFEs)}
        \end{subfigure} &
        \begin{subfigure}{0.48\columnwidth}
            \includegraphics[width=\linewidth]{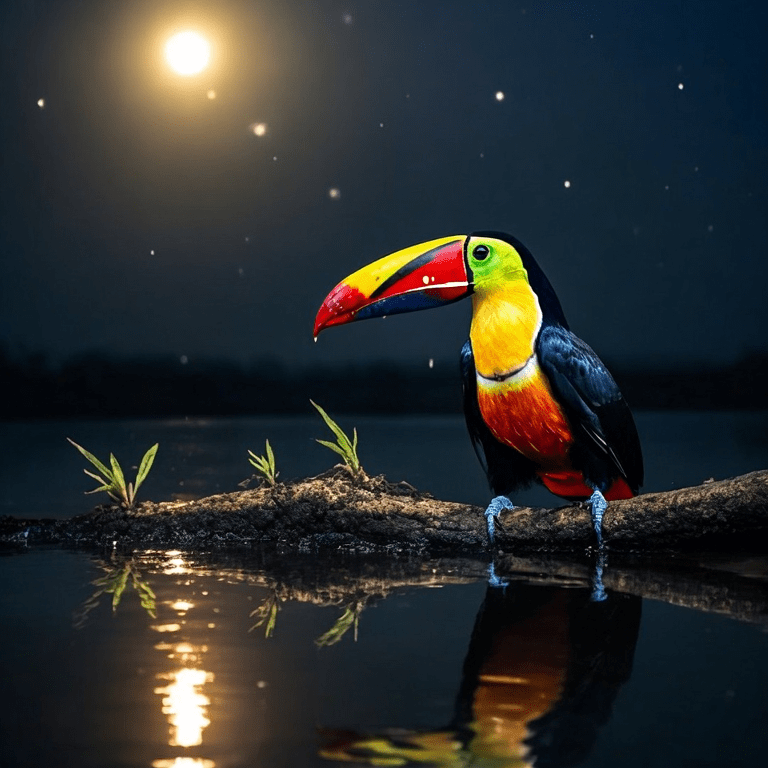}
            \caption{\ourmethod, $\Bar{\gamma}=0.991$ (30NFEs)}
        \end{subfigure} \\
        \multicolumn{2}{c}{\footnotesize ``three wolf moon but with cats instead of wolves'' (win)} \\

        % Third pair of images
        \begin{subfigure}{0.48\columnwidth}
            \includegraphics[width=\linewidth]{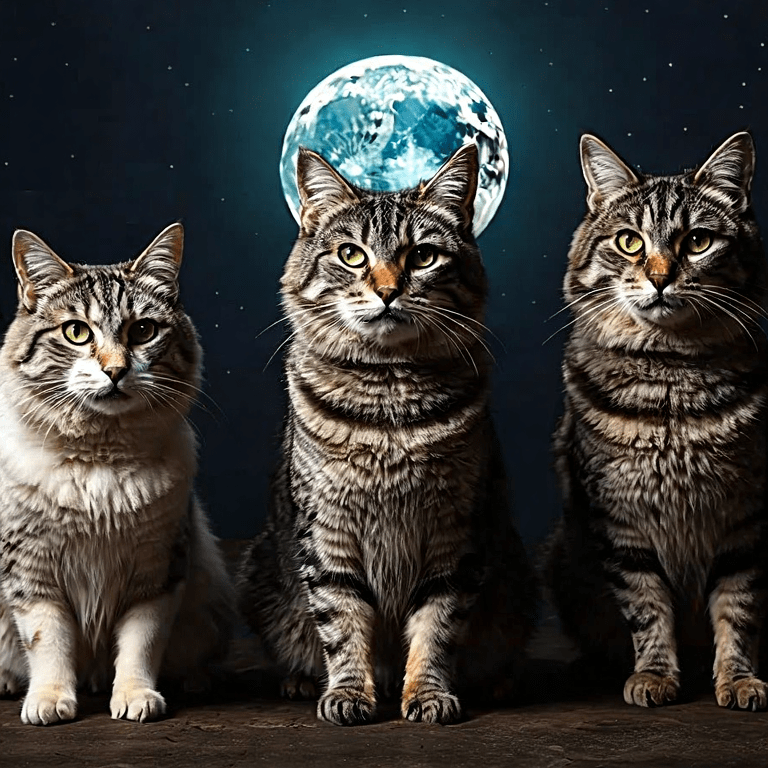}
            \caption{\textsc{Cfg} (40NFEs)}
        \end{subfigure} &
        \begin{subfigure}{0.48\columnwidth}
            \includegraphics[width=\linewidth]{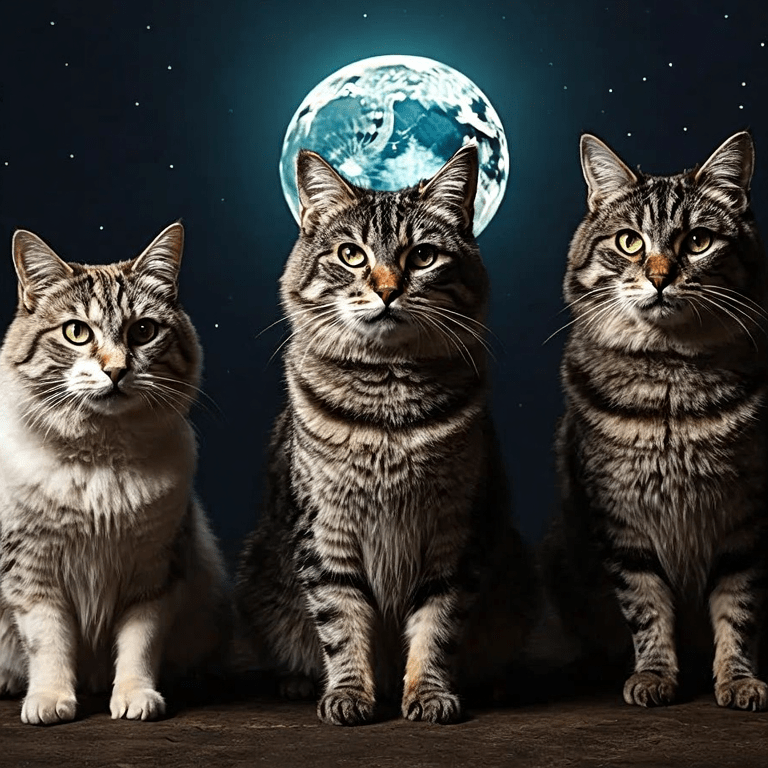}
            \caption{\ourmethod, $\Bar{\gamma}=0.991$ (31NFEs)}
        \end{subfigure} \\

         \multicolumn{2}{c}{\footnotesize ``a realistic medieval castle built for bees in a sunflower field'' (win)} \\

        % Third pair of images
        \begin{subfigure}{0.48\columnwidth}
            \includegraphics[width=\linewidth]{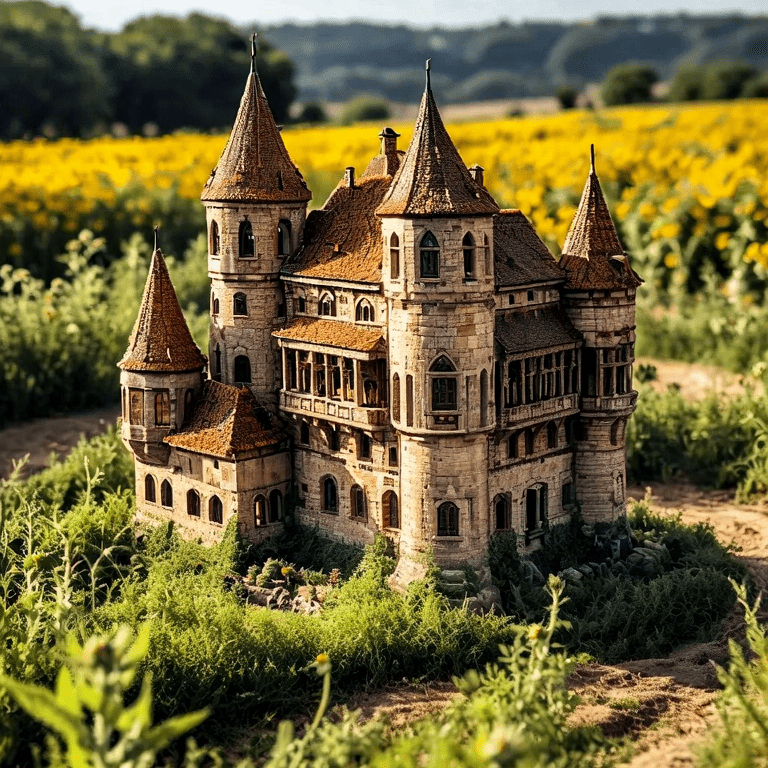}
            \caption{\textsc{Cfg} (40NFEs)}
        \end{subfigure} &
        \begin{subfigure}{0.48\columnwidth}
            \includegraphics[width=\linewidth]{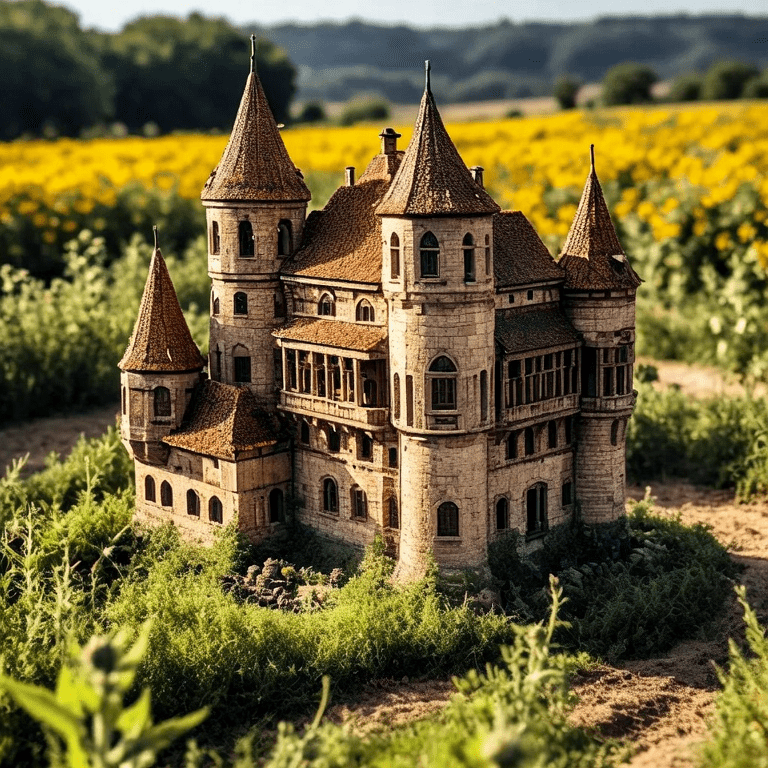}
            \caption{\ourmethod, $\Bar{\gamma}=0.991$ (29NFEs)}
        \end{subfigure}

    \end{tabular}
    \vspace{-0.5cm}
    \caption{\textbf{Human evaluation examples (win).} \footnotesize More samples from the human evaluation trials. The figure depicts a subset biased towards greater visual difference. We emphasize that images drawn uniformly from the dataset almost always look alike. This explains the draw situation depicted in Table \ref{tab:human_eval}.}
\label{fig:human_eval}
\end{figure}

\begin{figure}[t!]
    \centering
    \setlength{\tabcolsep}{2pt}

    \begin{tabular}{c c}
        % Row title A
        \multicolumn{2}{c}{\footnotesize  ``Fast commuter train moving past an outdoor platform.'' (lose)} \\
    
        % First pair of images
        \begin{subfigure}{0.48\columnwidth} 
            \includegraphics[width=\linewidth]{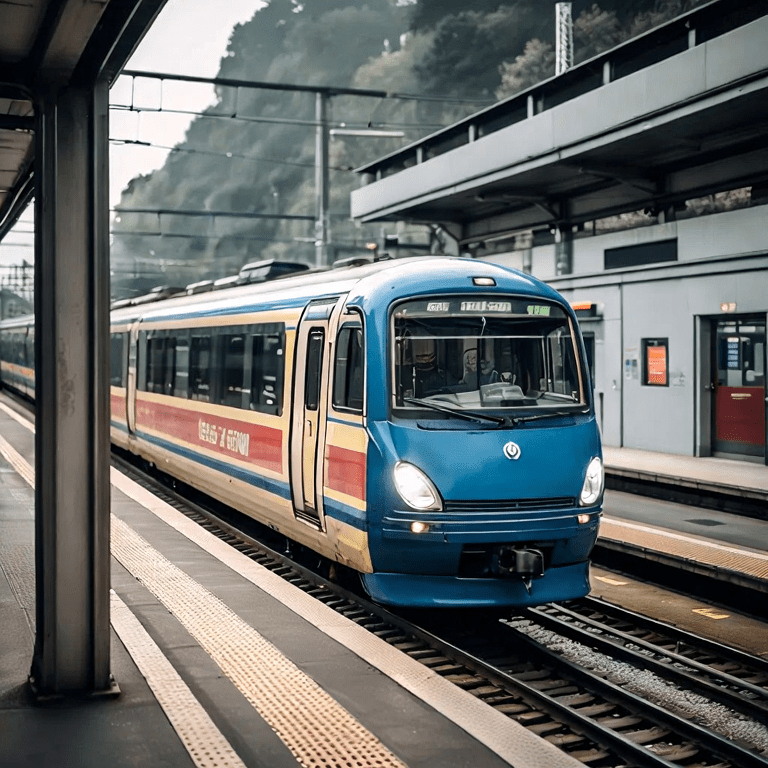}
            \caption{\textsc{Cfg} (40NFEs)}
        \end{subfigure} &
        \begin{subfigure}{0.48\columnwidth}
            \includegraphics[width=\linewidth]{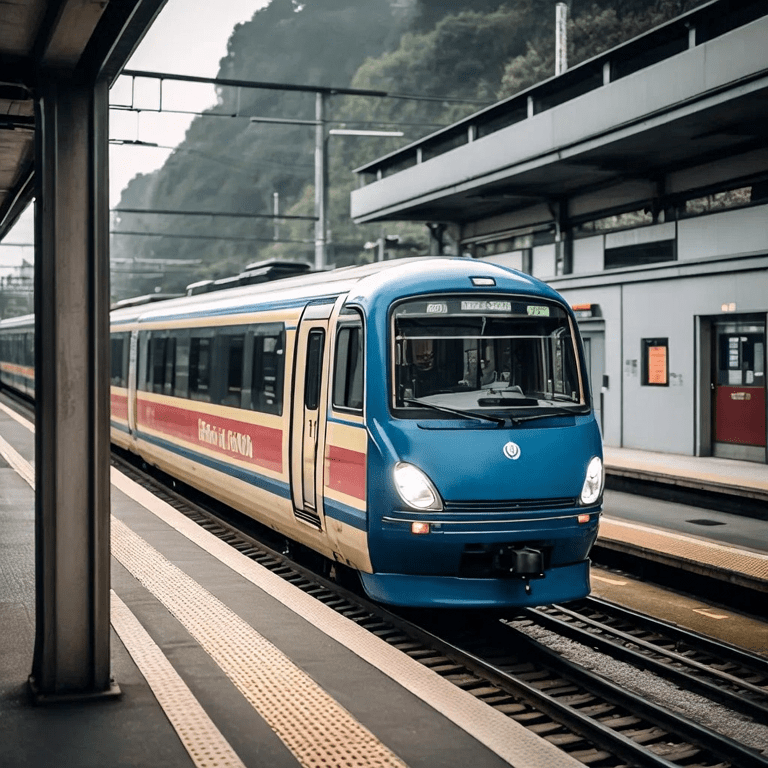}
            \caption{\ourmethod, $\Bar{\gamma}=0.991$ (31NFEs)}
        \end{subfigure} \\
        \multicolumn{2}{c}{\footnotesize  ``Three bears standing in a field outside.'' (lose)} \\
    
        % First pair of images
        \begin{subfigure}{0.48\columnwidth} 
            \includegraphics[width=\linewidth]{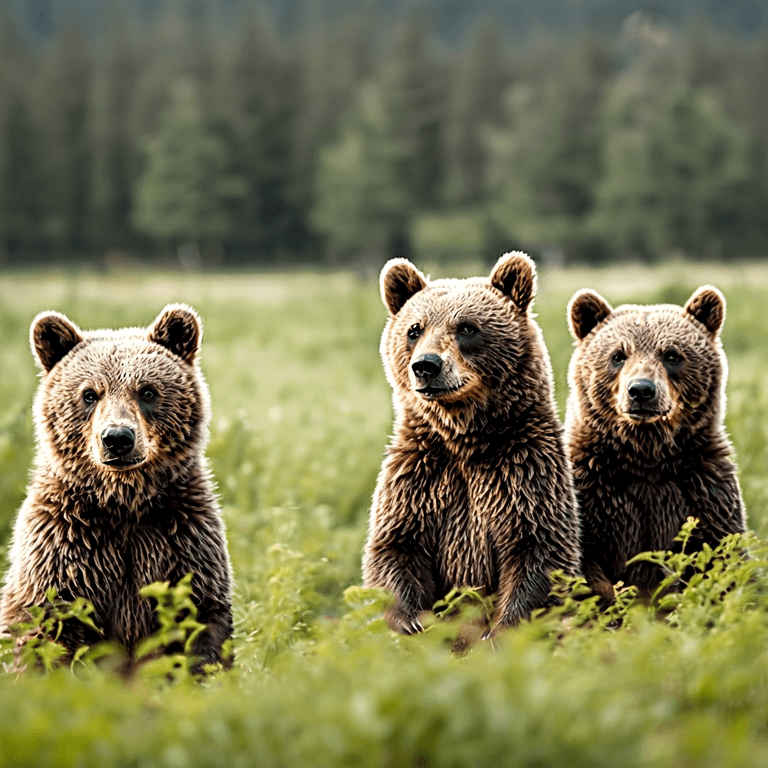}
            \caption{\textsc{Cfg} (40NFEs)}
        \end{subfigure} &
        \begin{subfigure}{0.48\columnwidth}
            \includegraphics[width=\linewidth]{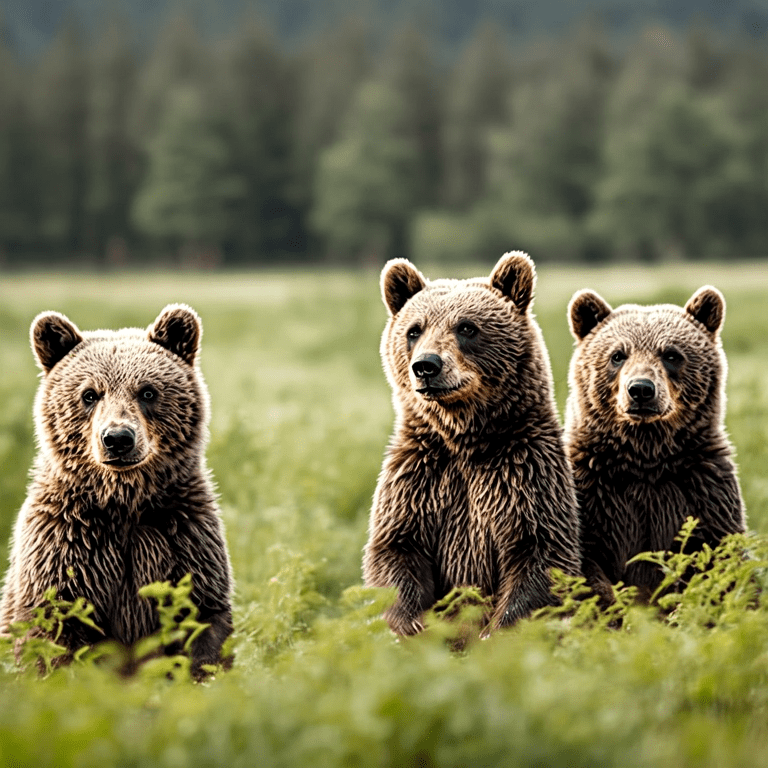}
            \caption{\ourmethod, $\Bar{\gamma}=0.991$ (29NFEs)}
        \end{subfigure} \\
        \multicolumn{2}{c}{\footnotesize ``bee farm, The beatles, bees, honey, honey farm'' (win)} \\

        % Third pair of images
        \begin{subfigure}{0.48\columnwidth}
            \includegraphics[width=\linewidth]{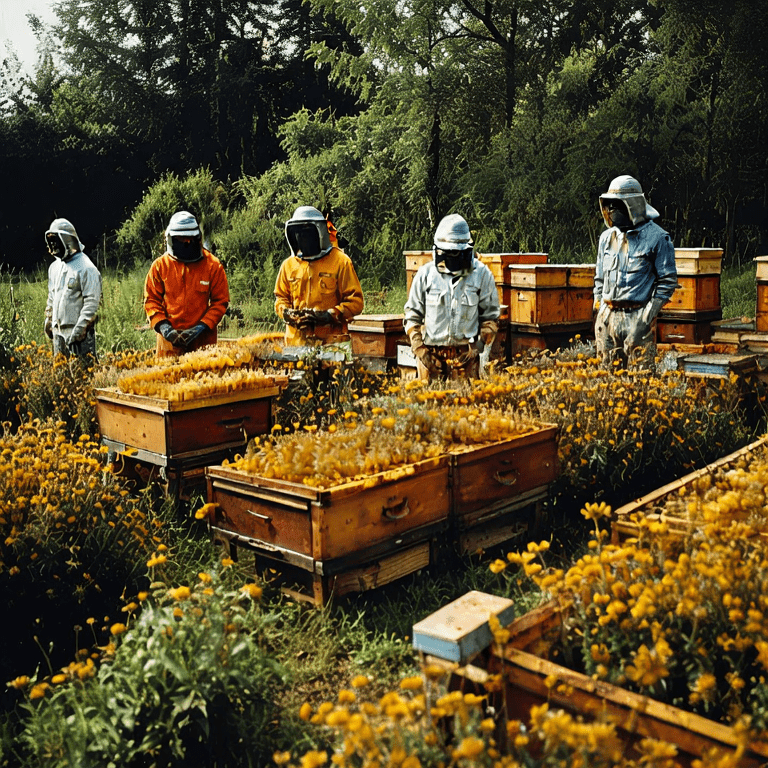}
            \caption{\textsc{Cfg} (40NFEs)}
        \end{subfigure} &
        \begin{subfigure}{0.48\columnwidth}
            \includegraphics[width=\linewidth]{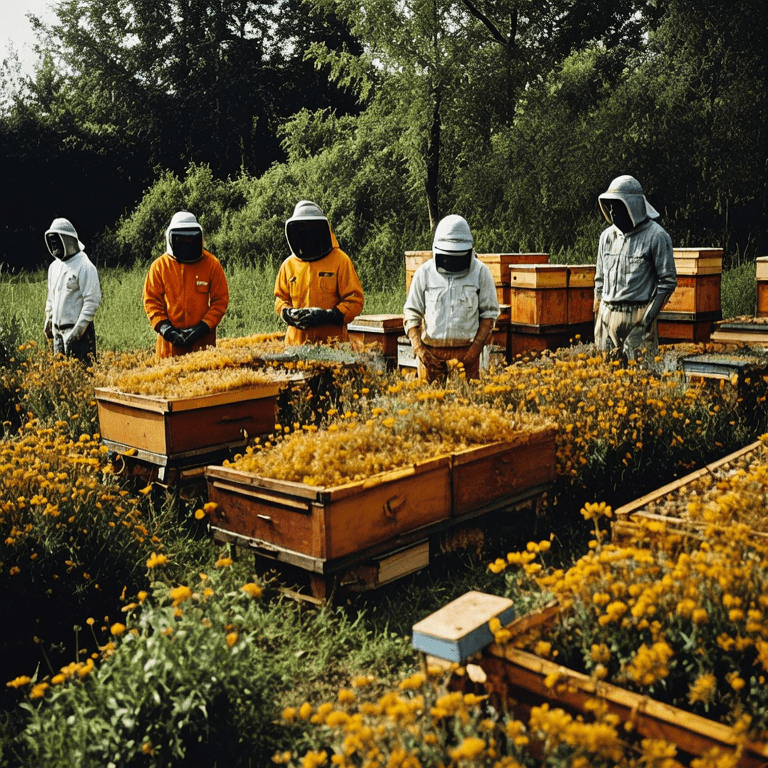}
            \caption{\ourmethod, $\Bar{\gamma}=0.991$ (30NFEs)}
        \end{subfigure} \\

         \multicolumn{2}{c}{\footnotesize ``two cats patting a magical crystal ball'' (win)} \\

        % Third pair of images
        \begin{subfigure}{0.48\columnwidth}
            \includegraphics[width=\linewidth]{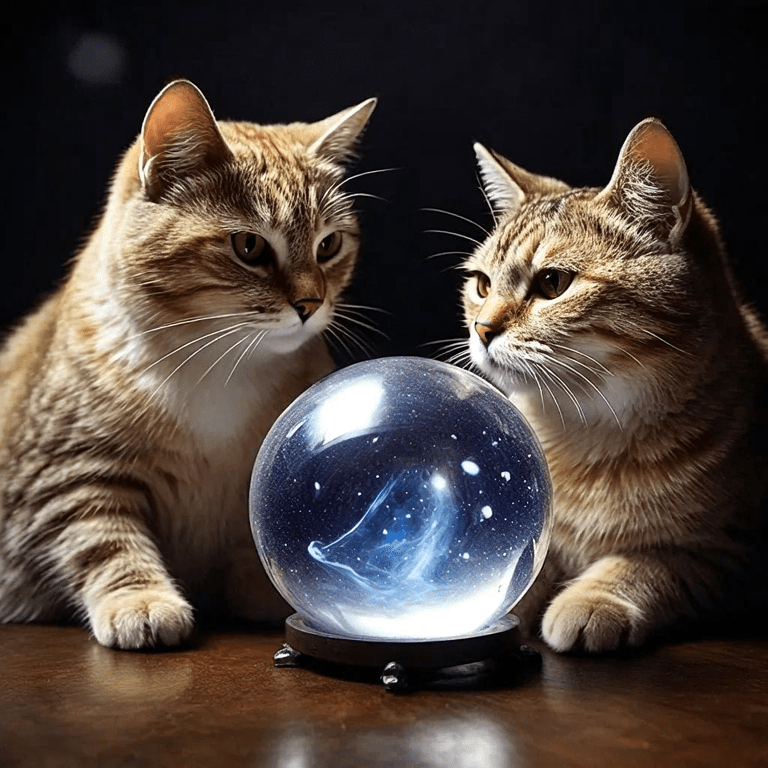}
            \caption{\textsc{Cfg} (40NFEs)}
        \end{subfigure} &
        \begin{subfigure}{0.48\columnwidth}
            \includegraphics[width=\linewidth]{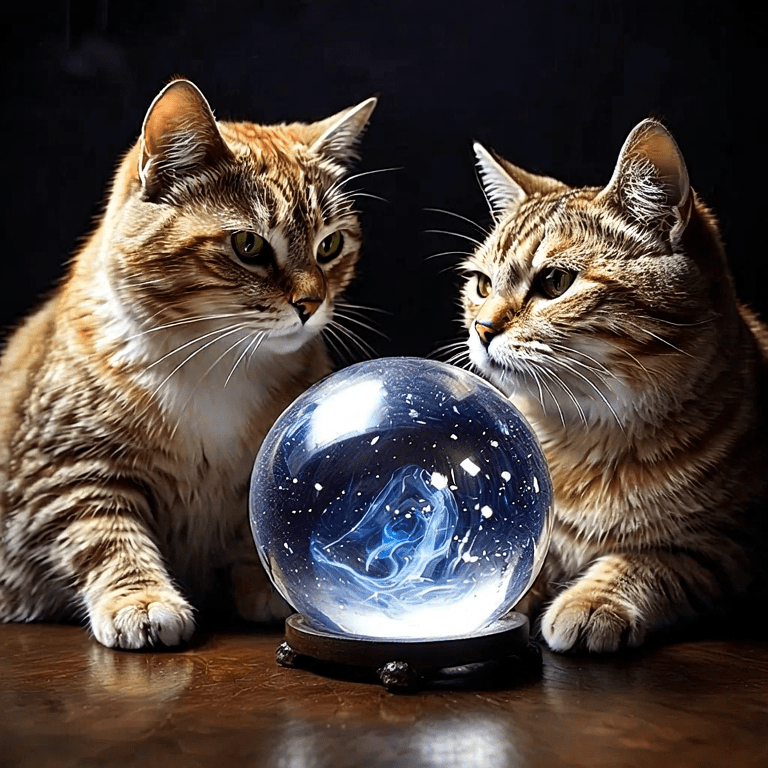}
            \caption{\ourmethod, $\Bar{\gamma}=0.991$ (30NFEs)}
        \end{subfigure}

    \end{tabular}
    \vspace{-0.5cm}
    \caption{\textbf{Human evaluation examples (lose).} \footnotesize More samples from the human evaluation trials. The figure depicts a subset biased towards greater visual difference. We emphasize that images drawn uniformly from the dataset almost always look alike. This explains the draw situation depicted in Table \ref{tab:human_eval}.}
\label{fig:human_eval}
\end{figure}

\section{Image editing}\label{appx:editing}

\begin{figure}[h!]
\setlength{\tabcolsep}{0.05em} % width separation
    \centering
    \begin{tabular}{c c c}
    \multicolumn{3}{c}{\footnotesize ``Turn the horse into a cow''} \\ % Add the title for the first row here
    \begin{subfigure}{0.33\columnwidth} 
        \includegraphics[width=\linewidth, trim=1 1 1 1, clip]{figures/fig1/astronaut_20.png}
    \end{subfigure}    &
    \begin{subfigure}{0.33\columnwidth}
        \includegraphics[width=\linewidth, trim=1 1 1 1, clip]{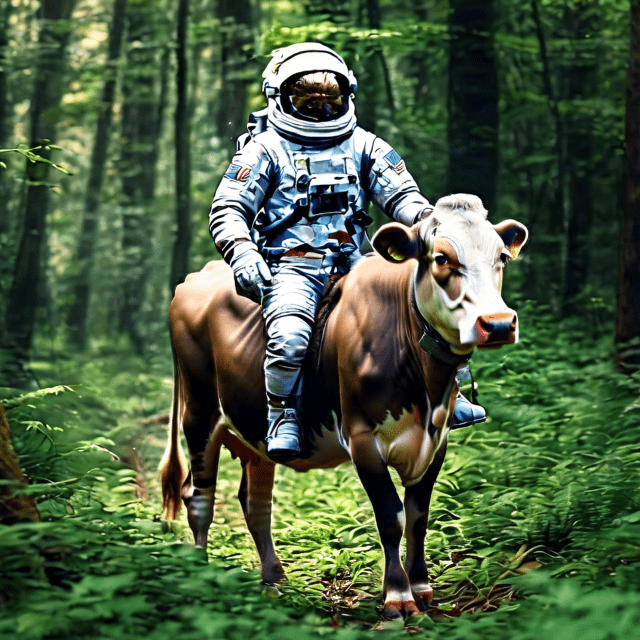}
    \end{subfigure} &
     \begin{subfigure}{0.33\columnwidth}
        \includegraphics[width=\linewidth, trim=1 1 1 1, clip]{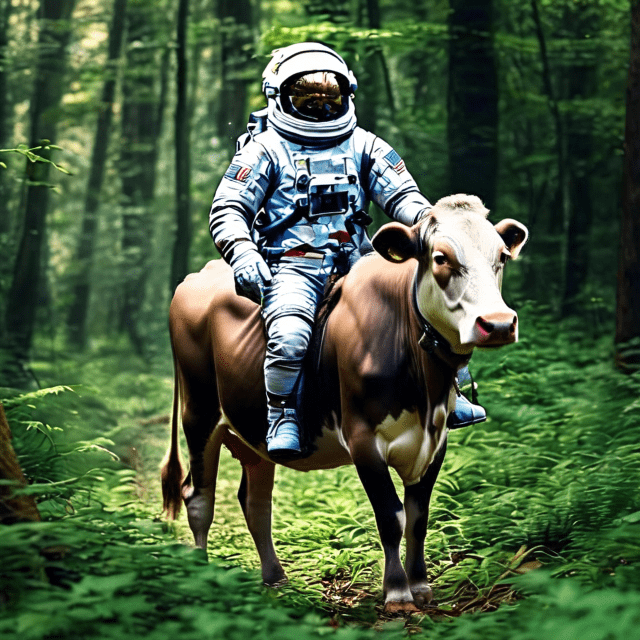}
    \end{subfigure}   \\
    \multicolumn{3}{c}{\footnotesize ``Make it winter''} \\ % Add the title for the first row here
    \begin{subfigure}{0.33\columnwidth} 
        \includegraphics[width=\linewidth, trim=1 1 1 1, clip]{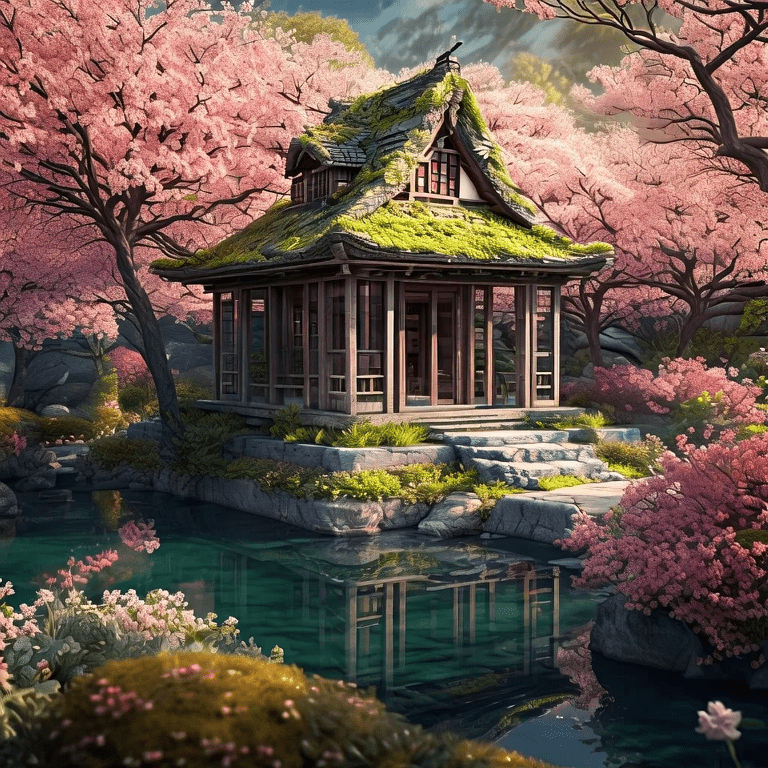}
    \end{subfigure}    &
    \begin{subfigure}{0.33\columnwidth}
        \includegraphics[width=\linewidth, trim=1 1 1 1, clip]{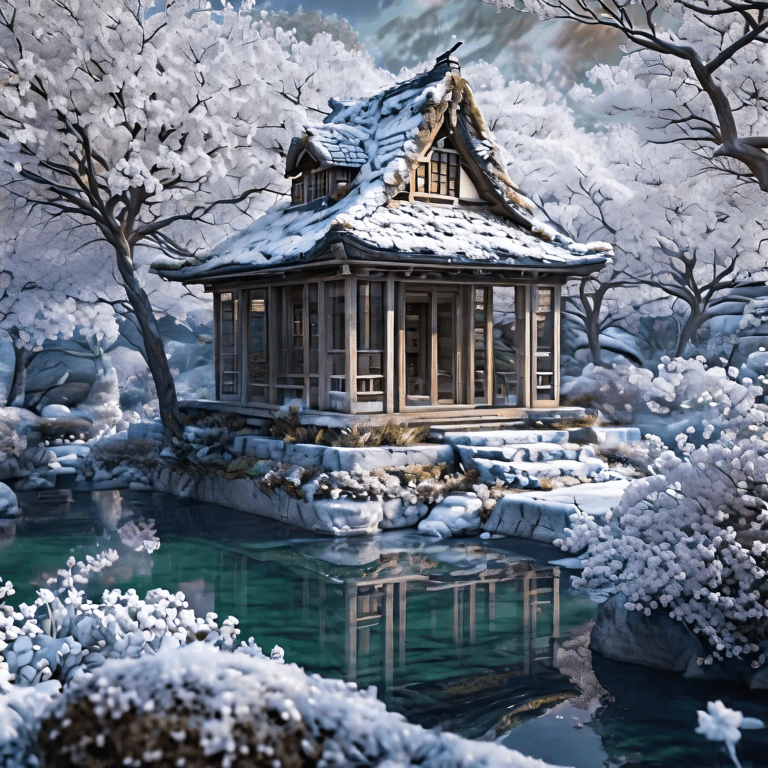}
    \end{subfigure} &
     \begin{subfigure}{0.33\columnwidth}
        \includegraphics[width=\linewidth, trim=1 1 1 1, clip]{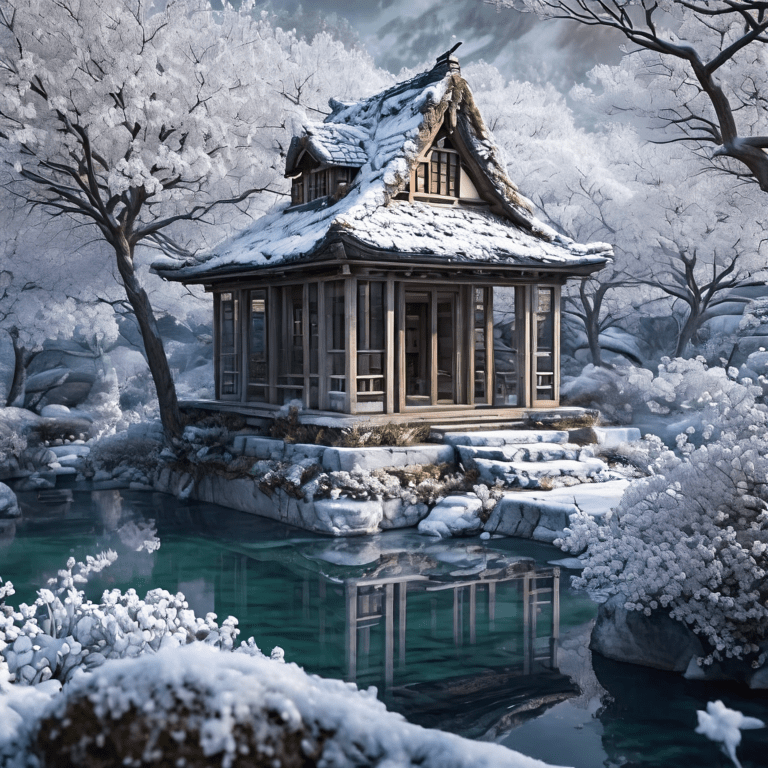}
    \end{subfigure}   \\

    \multicolumn{3}{c}{\footnotesize ``Turn the hat into a wreath of flowers''} \\ % Add the title for the second row here
    \begin{subfigure}{0.33\columnwidth} 
        \includegraphics[width=\linewidth, trim=1 1 1 1, clip]{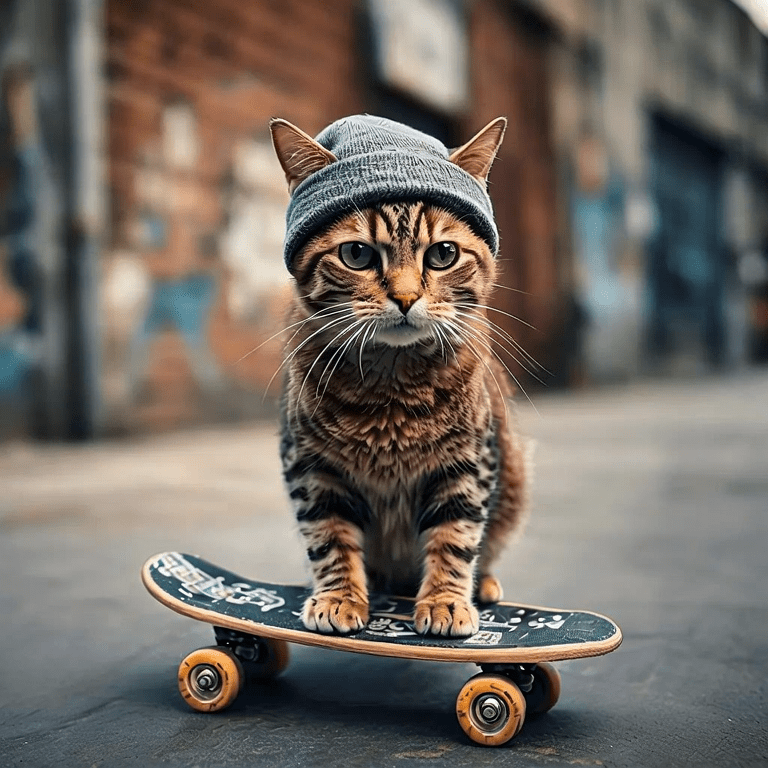}
        \caption{\centering  \textcolor{white}{whit
} Input image \textcolor{white}{for the win}}
    \end{subfigure}    &
    \begin{subfigure}{0.33\columnwidth}
        \includegraphics[width=\linewidth, trim=1 1 1 1, clip]{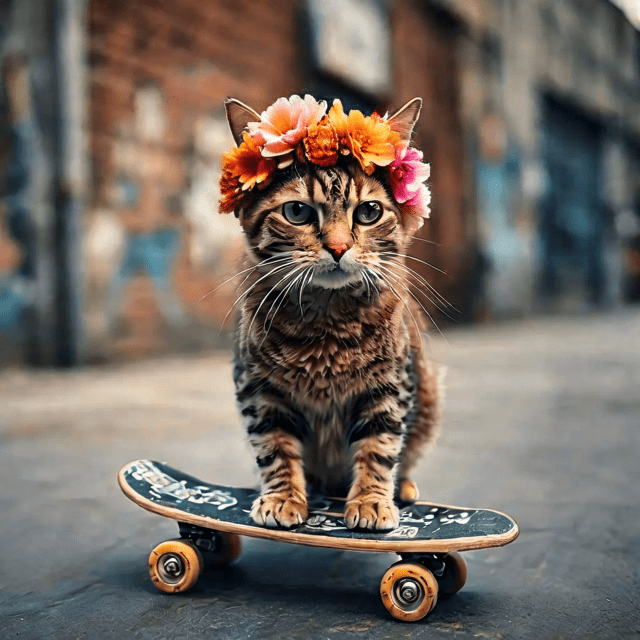}
        \caption{\centering \cfg-based editing (60NFEs)}
    \end{subfigure} &
     \begin{subfigure}{0.33\columnwidth}
        \includegraphics[width=\linewidth, trim=1 1 1 1, clip]{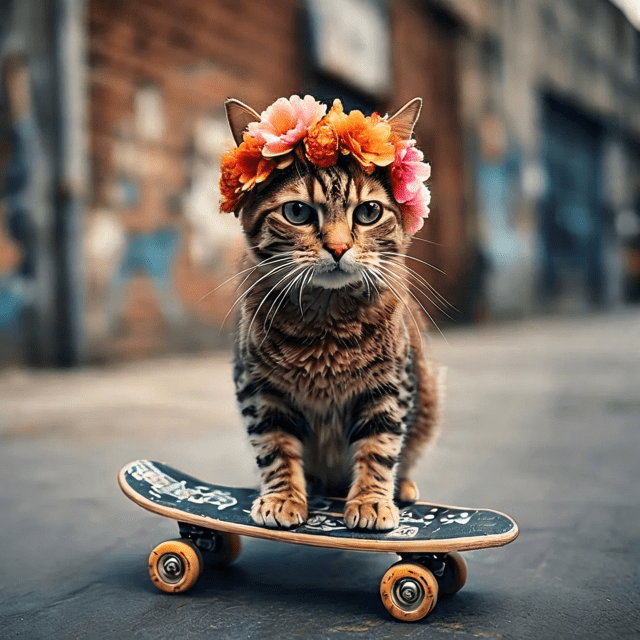}
        \caption{\centering \ourmethod-based editing (40NFEs)}
    \end{subfigure}   
    \end{tabular}
    \vspace{-0.4cm}
    \caption{\textbf{Image editing.} \footnotesize Instruction based editing with EMU Edit \cite{sheynin2023emu}, which builds upon InstructPix2Pix \cite{brooks2023instructpix2pix}. Depicted are the original image (left), classic \cfg editing (Eq.~\ref{eq:cfg-score-pix2pix}) and \ourmethod editing, which gives equal quality results while reducing NFEs by $33.3\%$. Importantly, Guidance Distillation is not directly applicable for this task as the update steps are conditioned on the input image.}
    \label{fig:editing}
\end{figure}

A large body of works proposes to use text-to-image models not only for generation of novel images but also for instruction based editing of existing ones (\eg,~\cite{meng2021sdedit,zhang2023magicbrush,brooks2023instructpix2pix,sheynin2023emu}). One particularly successful approach within this realm, termed InstructPix2Pix \cite{brooks2023instructpix2pix}, achieve successful image editing by augmenting the \cfg paradigm to not only text but image and text conditioning, giving rise to the modified score estimate

\begin{equation}\label{eq:cfg-score-pix2pix}
\begin{aligned}
    \mathbf{\epsilon}_{\text{pix2pix}}(\mathbf{x}_t,\mathbf{c},\mathbf{I}) =& \mathbf{\epsilon}_\theta(\mathbf{x}_t, \emptyset,\emptyset) \\ & + s_c \cdot (\mathbf{\epsilon}_\theta(\mathbf{x}_t, \mathbf{c},\mathbf{I}) - \mathbf{\epsilon}_\theta(\mathbf{x}_t, \emptyset,\mathbf{I})) \\ & + s_T  \cdot (\mathbf{\epsilon}_\theta(\mathbf{x}_t, \emptyset,\mathbf{I}) - \mathbf{\epsilon}_\theta(\mathbf{x}_t, \emptyset,\emptyset)).
\end{aligned}
\end{equation}

This has two important implications. First, a single step in the diffusion process now requires 3 instead of 2 NFEs. Second, Guidance Distillation can no longer be applied as part of the ``unconditional'' update step is now dynamic (\ie, $\mathbf{I}$ changes across samples, akin to the case of negative prompts).

Both effects are unfortunate as fast generation is particularly relevant in the image editing context, where users may want to try various instructions in sequence. Interestingly, we find that -- similar to the case of simple text conditioning -- the terms in Eq.~\ref{eq:cfg-score-pix2pix} converge over time. Hence, as shown in Figure~\ref{fig:editing}, \ourmethod can again be employed to reduce NFEs without noticeable loss of quality. In the depicted images, \ourmethod employs only ten (instead of 20) $ \mathbf{\epsilon}_{\text{pix2pix}}(\mathbf{x}_t,\mathbf{c},\mathbf{I})$ steps, thereby saving $33.3\%$ of the total number of NFEs.

\section{OLS} \label{appx:ols}

\begin{figure}[h!]
    \centering
        \includegraphics[width=\linewidth,trim=15 5 25 55, clip]{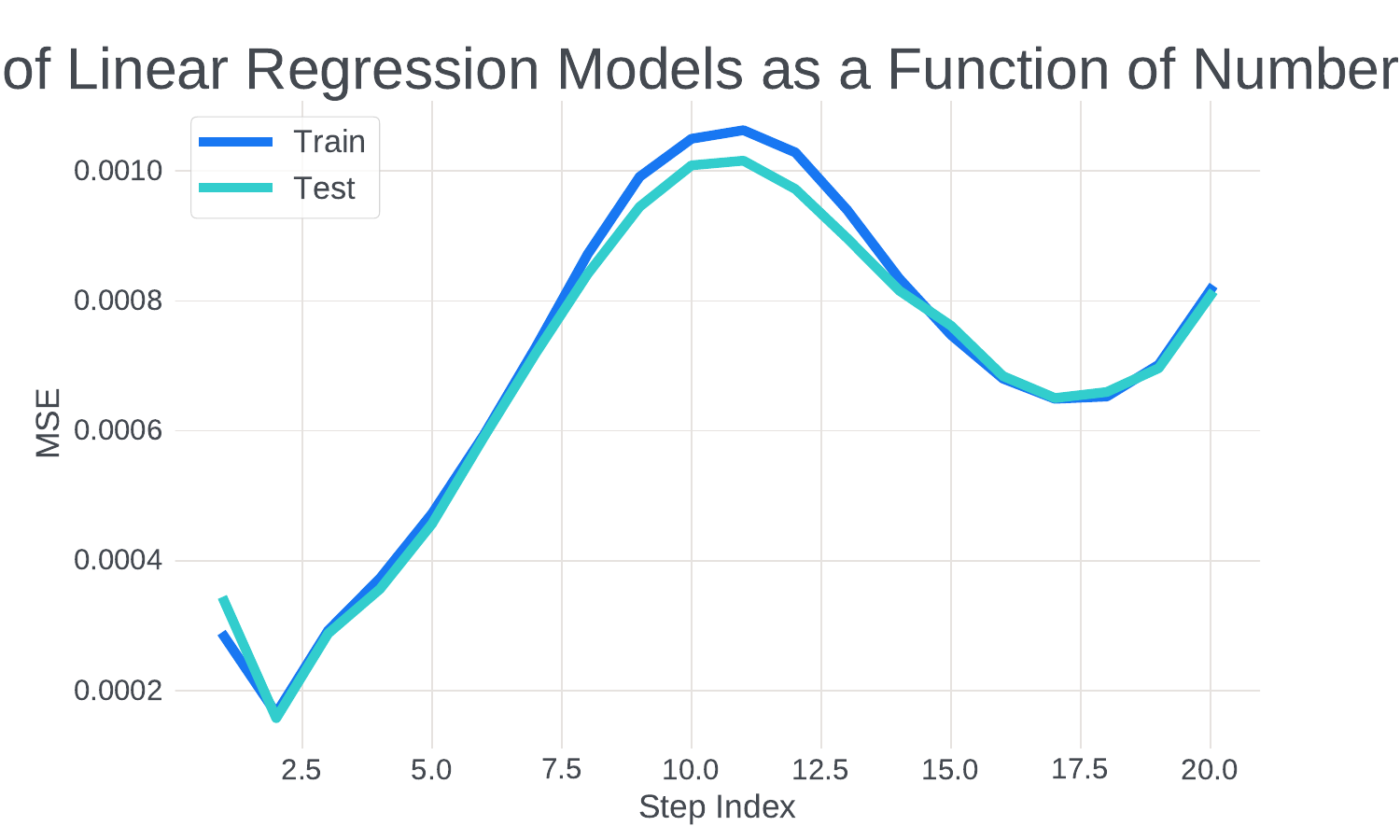}
    \vspace{-0.1cm}
    \caption{\textbf{Per-step OLS errors.} MSE of $\hat{\epsilon}(\mathbf{x}_t,\emptyset)$ and $\epsilon_{\theta}(\mathbf{x}_t,\emptyset)$ after learning the regression coefficients of Eq.~\ref{eq:ols}. The results depict 200 training and 100 test samples.}      
    \label{fig:ols_errors_apx}
\end{figure}
In section \ref{sec:ols}, we have shown that entire neural network calls can be replaced by a simple linear regression in the past. This is particularly relevant for the first half of the diffusion steps, where we found guidance most important. Towards this end, we generated as little as 200 paths from a 20-step \cfg model and trained 20 linear regression models, one for each timestep, always taking the past unconditional as well as the past- and current conditional steps as regressors and the current unconditional as target. Importantly, we learned a single (scalar) regression coefficient for each high-dimensional regressor.\footnote{Simple extensions like doing one OLS per channel did not show any significant improvement} The per-step errors of the learned LR models are depicted in Fig.~\ref{fig:ols_errors_apx}.

After training the LR models, one can replace the unconditional network call $\mathbf{\epsilon}_\theta(\mathbf{x}_t, \emptyset)$ in \cfg with the simple linear combination $\hat{\epsilon}(\mathbf{x}_t,\emptyset)$ from Eq.~\ref{eq:ols}, giving rise to
\begin{equation}\label{eq:cfg-score-appx}
    \hat{\mathbf{\epsilon}}_{\text{cfg}}(\mathbf{x}_t,\mathbf{c},s) =\hat{\epsilon}(\mathbf{x}_t,\emptyset) + s \cdot (\mathbf{\epsilon}_\theta(\mathbf{x}_t, \mathbf{c}) - \hat{\epsilon}(\mathbf{x}_t,\emptyset)).
\end{equation}

Importantly, this $\hat{\mathbf{\epsilon}}_{\text{cfg}}$ update now only costs 1 NFE compared to the 2 NFEs for $\mathbf{\epsilon}_{\text{cfg}}$ from Eq.~\ref{eq:cfg-score}.

We found that the LR estimators $\hat{\epsilon}(\mathbf{x}_t,\emptyset)$ can replace all unconditional network calls $\epsilon(\mathbf{x}_t,\emptyset)$ when given hypothetical ground truth past information. Of course, such information is no longer available once an upstream \cfg step has been replaced with  $\hat{\mathbf{\epsilon}}_{\text{cfg}}$. Having observed that the regression weights $\beta_i$ are highest for the most recent past, we found the best policy to be one that alternates between true \cfg steps and LR-based \cfg steps. For example, for the twenty-step baseline we are using the following policy

\begin{equation}\label{eq:zeta_ols}
\begin{aligned}
\zeta_{\ourlinearmethod}=&[\epsilon_{\text{cfg},T},\hat{\epsilon}_{\text{cfg},T-1},\epsilon_{\text{cfg},T-2},\hat{\epsilon}_{\text{cfg},T-3} \text{...}\epsilon_{\text{cfg},T/2},\\
&\hat{\epsilon}_{\text{cfg},T/2-1} ,\hat{\epsilon}_{\text{cfg},T/2-2}, \text{...},\hat{\epsilon}_{\text{cfg},0} ].
\end{aligned}
\end{equation}

\begin{figure}[h!]
\setlength{\tabcolsep}{0.05em} % width separation
    \centering
    \begin{tabular}{c c c}
      \multicolumn{3}{c}{\footnotesize ``A painting of a gondola in the canals of 16th century Venice''} \\ % Add the title for the first row here
    \begin{subfigure}{0.33\columnwidth} 
        \includegraphics[width=\linewidth, trim=1 1 1 1, clip]{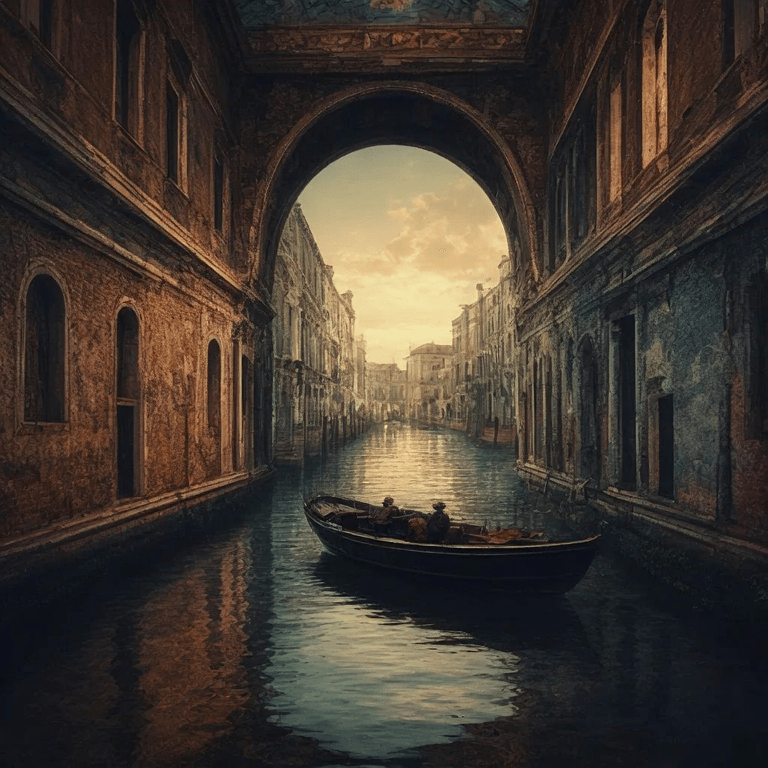}
    \end{subfigure}    &
    \begin{subfigure}{0.33\columnwidth}
        \includegraphics[width=\linewidth, trim=1 1 1 1, clip]{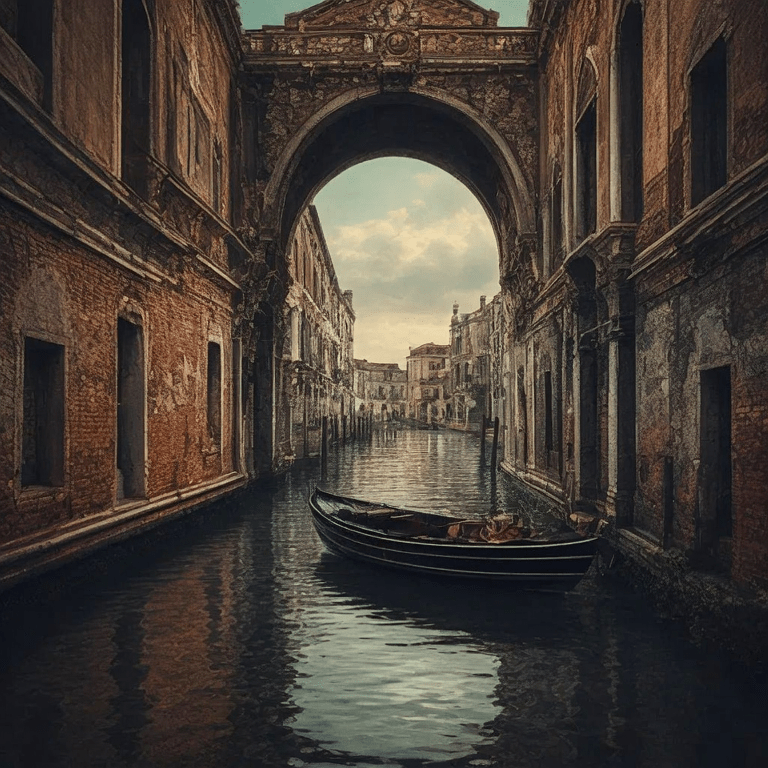}
    \end{subfigure} &
     \begin{subfigure}{0.33\columnwidth}
        \includegraphics[width=\linewidth, trim=1 1 1 1, clip]{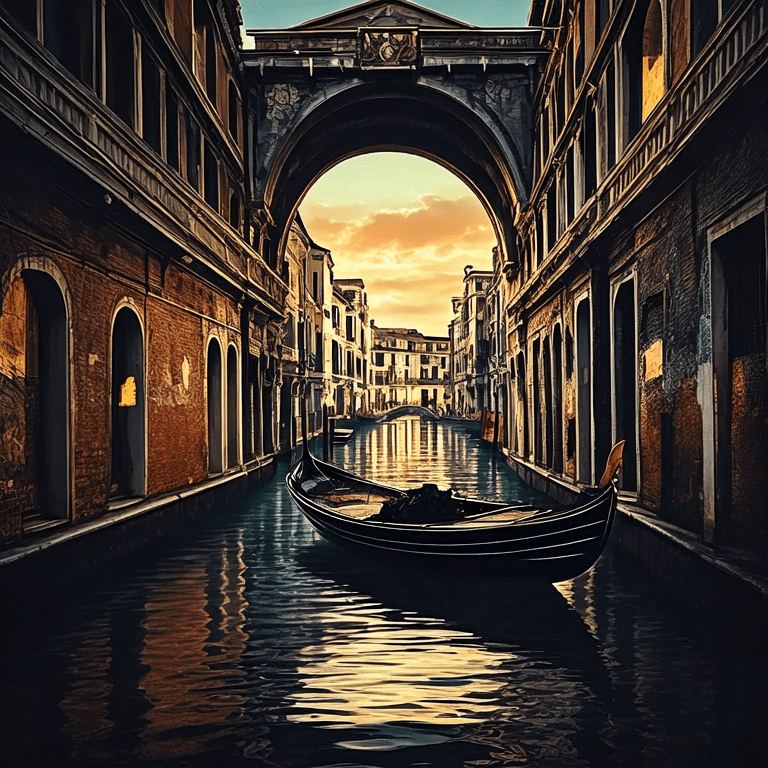}
    \end{subfigure}   \\
     \multicolumn{3}{c}{\footnotesize ``A group of porcelain tucans painted in Inka style''} \\ % Add the title for the first row here
    \begin{subfigure}{0.33\columnwidth} 
        \includegraphics[width=\linewidth, trim=1 1 1 1, clip]{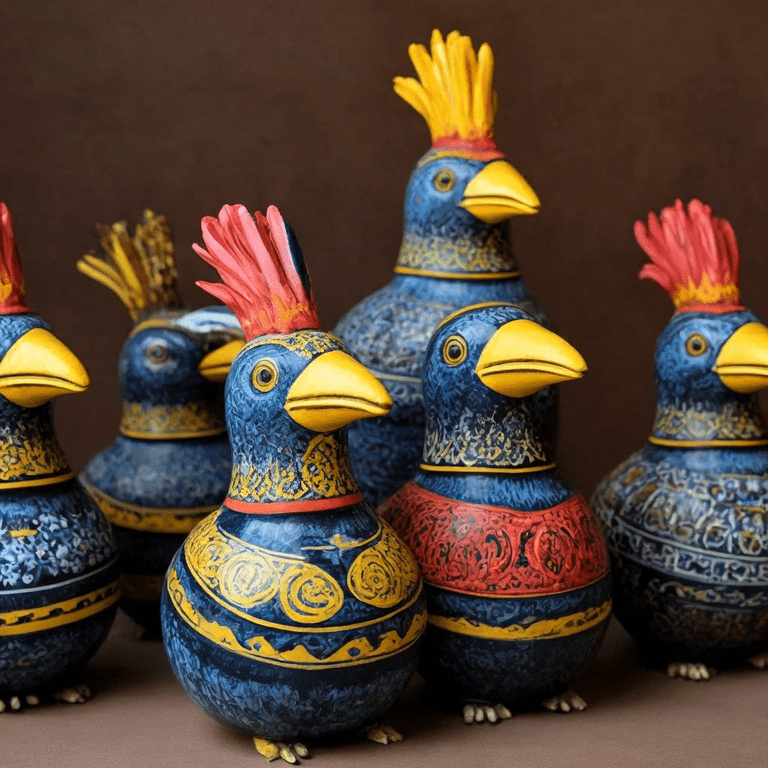}
    \end{subfigure}    &
    \begin{subfigure}{0.33\columnwidth}
        \includegraphics[width=\linewidth, trim=1 1 1 1, clip]{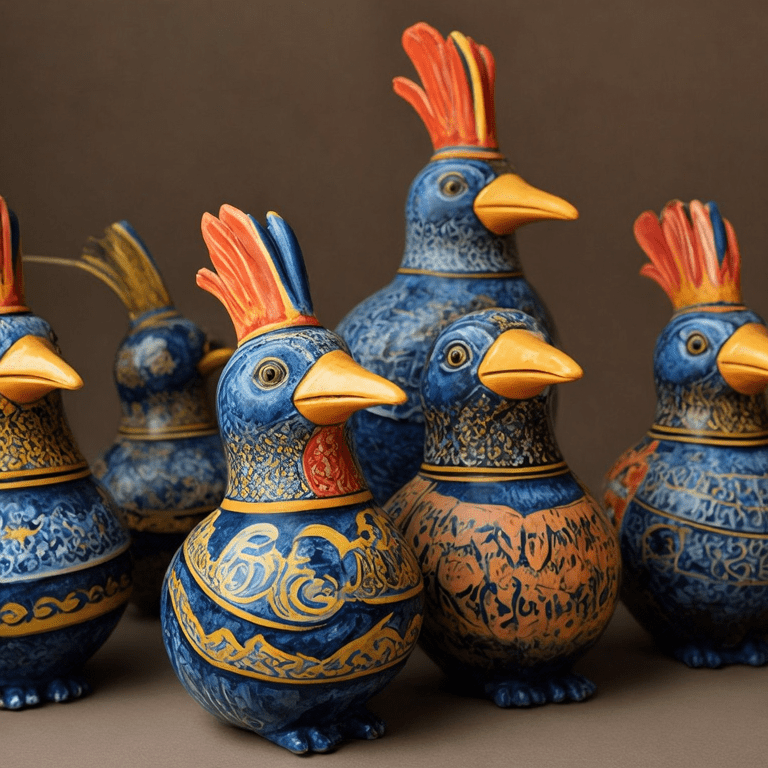}
    \end{subfigure} &
     \begin{subfigure}{0.33\columnwidth}
        \includegraphics[width=\linewidth, trim=1 1 1 1, clip]{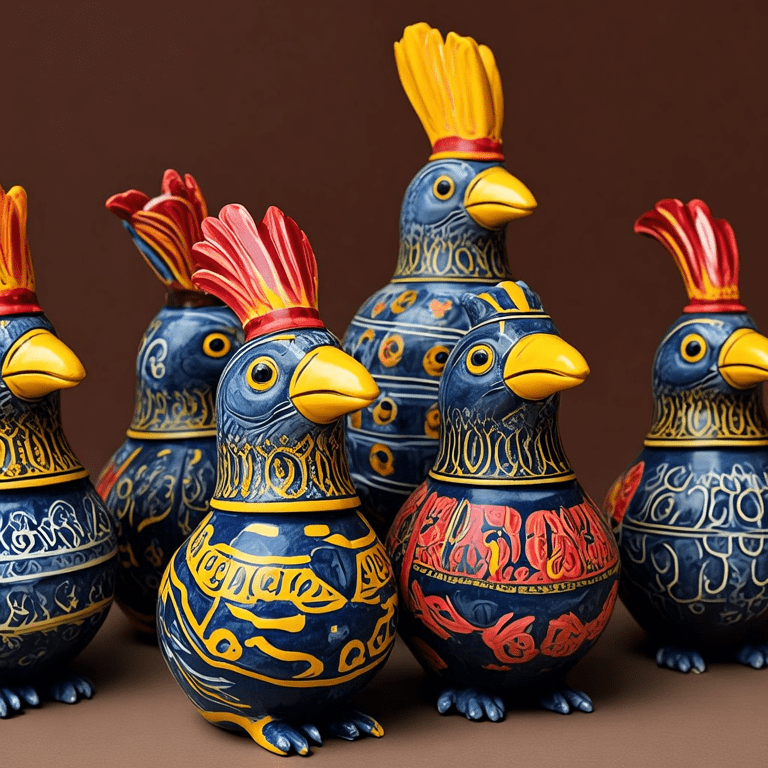}
    \end{subfigure}   \\
    \multicolumn{3}{c}{\footnotesize ``An ancient castle on a cliff overlooking a vast, mist-covered valley''} \\ % Add the title for the first row here
    \begin{subfigure}{0.33\columnwidth} 
        \includegraphics[width=\linewidth, trim=1 1 1 1, clip]{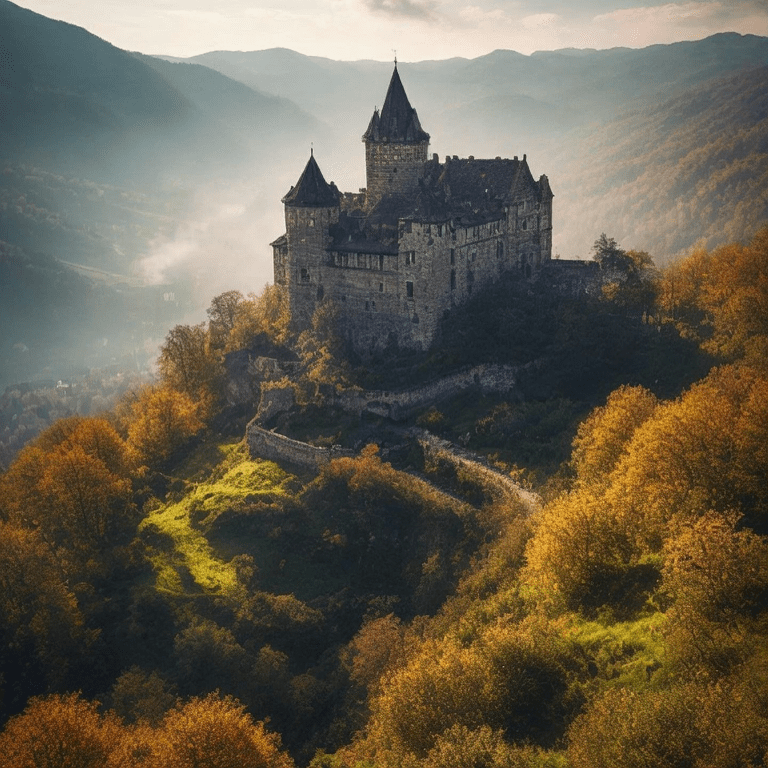}
    \end{subfigure}    &
    \begin{subfigure}{0.33\columnwidth}
        \includegraphics[width=\linewidth, trim=1 1 1 1, clip]{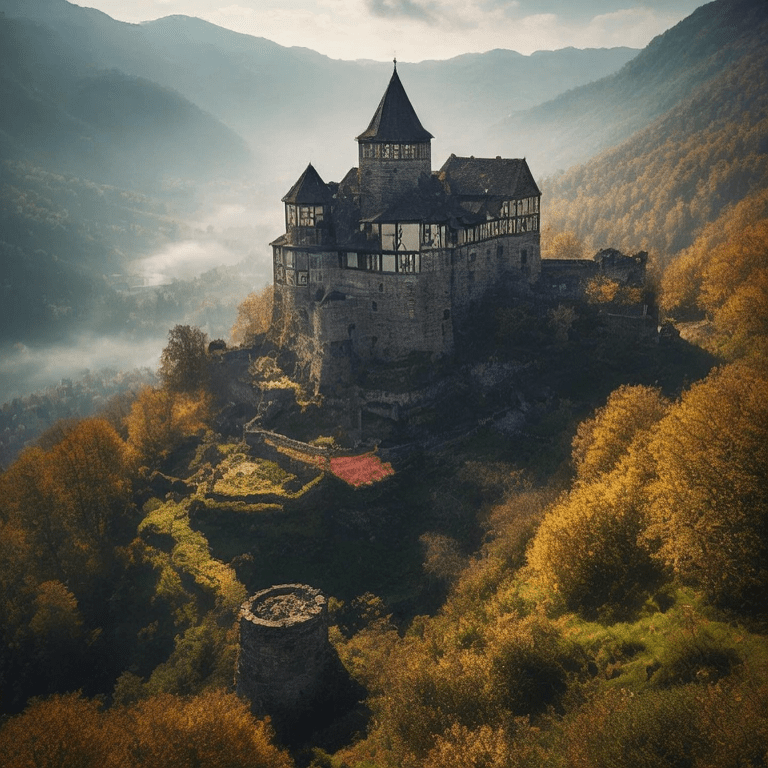}
    \end{subfigure} &
     \begin{subfigure}{0.33\columnwidth}
        \includegraphics[width=\linewidth, trim=1 1 1 1, clip]{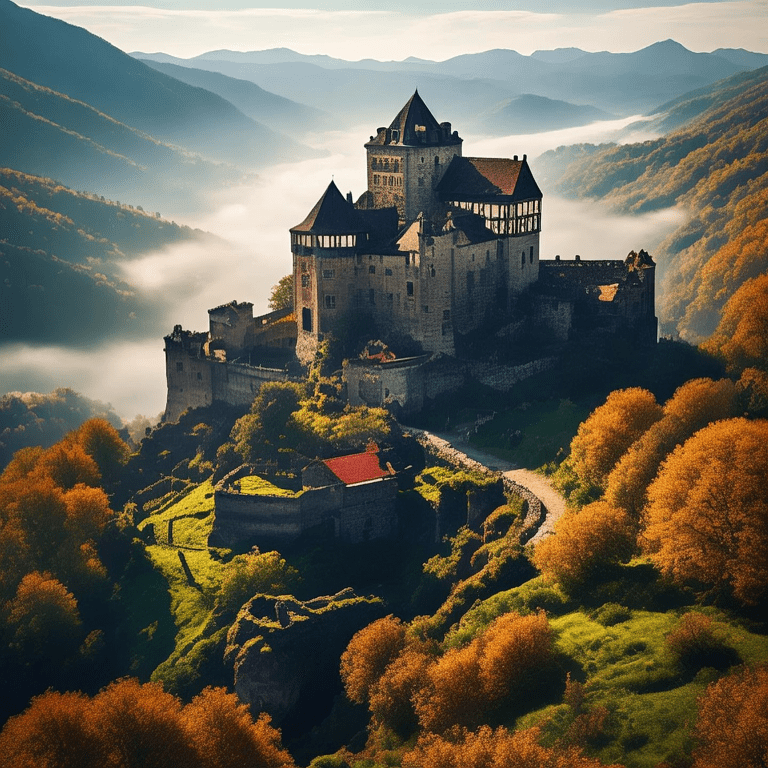}
    \end{subfigure}   \\
    \multicolumn{3}{c}{\footnotesize ``A giraffe eating a green plant''} \\ % Add the title for the second row here
    \begin{subfigure}{0.33\columnwidth} 
        \includegraphics[width=\linewidth, trim=1 1 1 1, clip]{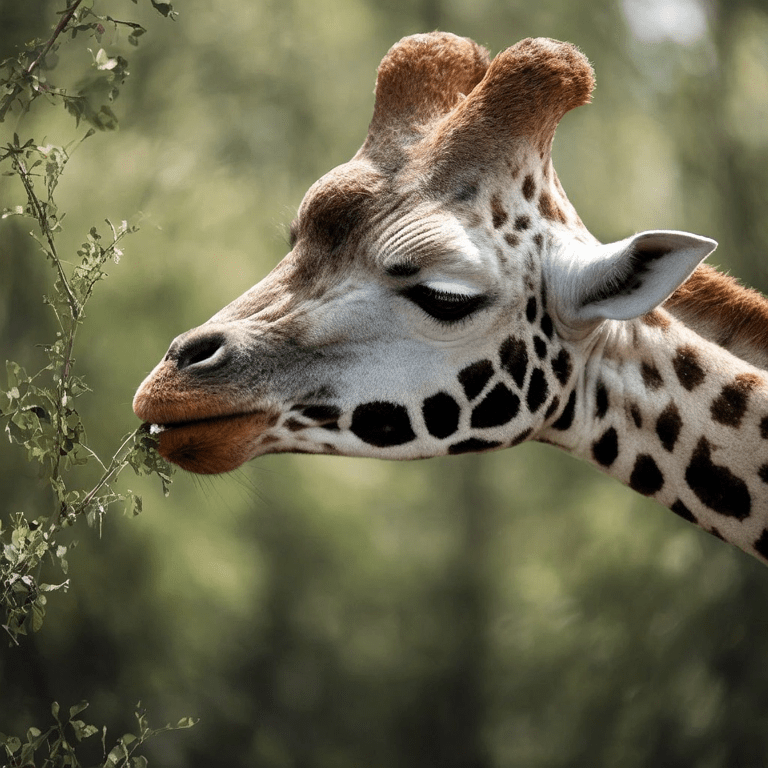}
        \caption{\centering \ourmethod $\Bar{\gamma}=0.975$ (25NFEs)}
    \end{subfigure}    &
    \begin{subfigure}{0.33\columnwidth}
        \includegraphics[width=\linewidth, trim=1 1 1 1, clip]{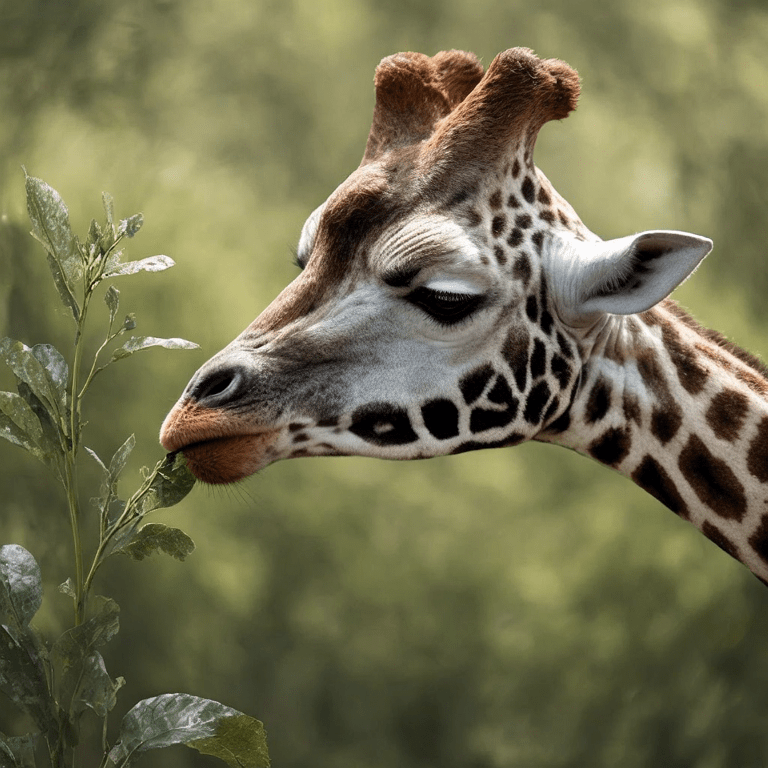}
        \caption{\centering na\"{i}ve interleaving \cfg (25NFEs)}
    \end{subfigure} &
     \begin{subfigure}{0.33\columnwidth}
        \includegraphics[width=\linewidth, trim=1 1 1 1, clip]{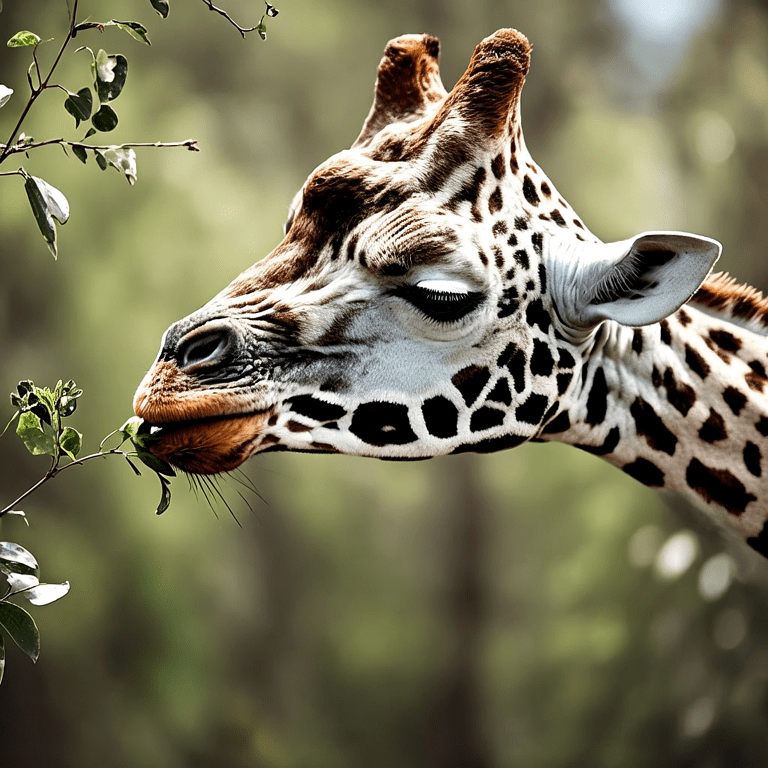}
        \caption{\centering \textcolor{white}{...}  \ourlinearmethod \textcolor{white}{...} (25NFEs)}
    \end{subfigure}   
    \end{tabular}
    \vspace{-0.4cm}
    \caption{\textbf{Replacing \cfg in the first half of diffusion steps.} \footnotesize Three different approaches to reduce the number of NFEs in the first $50\%$ of diffusion steps. As can be seen, \ourlinearmethod samples show increased sharpness, dynamic lightning with higher contrast, and more vivid colors. (Best viewed in zoom.)}
    \label{fig:ols_images_apx}
\end{figure}

All \ourlinearmethod results depicted in Fig.~\ref{fig:ols}, \ref{fig:ols_images_apx} and \ref{fig:negative_apx} used $\zeta_{\ourlinearmethod}$.

\begin{figure*}[htbp]
    \centering
    \begin{subfigure}{\textwidth}
        \centering
        \includegraphics[height=1.2cm]{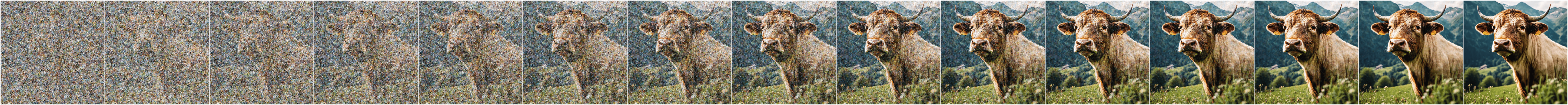} % replace with your first image file name
        \caption{Individual denoising iterates.}
        \label{fig:sub1}
    \end{subfigure}
    \hfill
    \begin{subfigure}{\textwidth}
        \centering
        \includegraphics[height=1.2cm]{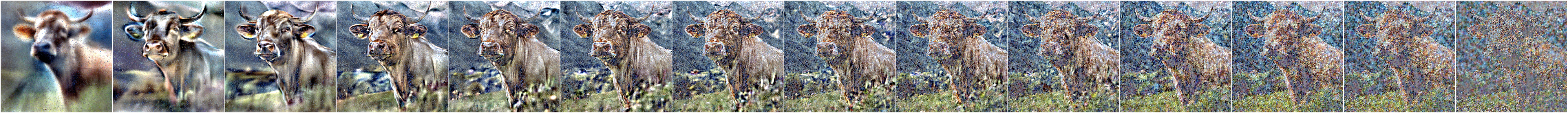} % replace with your second image file name
        \caption{Point-wise difference between iterates.}
        \label{fig:sub2}
    \end{subfigure}
    \caption{
    \textbf{Denoising process displays scene organization even in early iterations.}
    The (post-decoder) individual denoising iterates (top) suggest little information is known in the early iterations of the process.
    However, when computing the point-wise \textit{differences} between the decoded iterates (bottom) shows that even the earliest iterations of the denoising process already display scene organization.
    }
    \label{fig:semantics}
\end{figure*}

\end{document}